%% file: main.tex
\documentclass{article}

\usepackage[top=1in, bottom=1.25in, left=1in, right=1in]{geometry}
\usepackage{amssymb,amsfonts,amsmath, amsthm}
\RequirePackage[authoryear,round]{natbib}
\usepackage[colorlinks,linkcolor=blue,citecolor=blue]{hyperref}
\usepackage[labelfont=bf]{caption}
\usepackage{eufrak, euscript}
\usepackage{dsfont}
\usepackage{pdflscape}
\usepackage{setspace} 
\usepackage[ruled,linesnumbered]{algorithm2e}
\usepackage{graphicx}
\usepackage{float}
\usepackage{titlesec}
\usepackage{geometry}
\usepackage[normalem]{ulem}
\usepackage{authblk}

\makeatletter
\newcommand\footnoteref[1]{\protected@xdef\@thefnmark{\ref{#1}}\@footnotemark}
\makeatother

\author[a]{Sumanta Basu\footnote{SB and KK contributed equally to this work}}
\author[b]{Karl Kumbier$^*$}
\author[c,d,b,e]{James B. Brown\footnote{JB and BY contributed equally to this work}}
\author[b,f]{Bin Yu$^\dagger$}

\affil[a]{\footnotesize{Department of Biological Statistics and Computational Biology, Cornell University}}
\affil[b]{Statistics Department, University of California, Berkeley}
\affil[c]{Centre for Computational Biology, School of Biosciences, University of Birmingham}
\affil[d]{Molecular Ecosystems Biology Department, Lawrence Berkeley National Laboratory}
\affil[e]{Preminon, LLC}
\affil[f]{Department of Electrical Engineering and Computer Sciences, University of California, Berkeley}

\title{iterative Random Forests to discover predictive and stable high-order interactions}
\date{}
\begin{document}
\maketitle
\begin{abstract}
Genomics has revolutionized biology, enabling the interrogation of whole transcriptomes, genome-wide binding sites for proteins, and many other molecular processes. However, individual genomic assays measure elements that interact \textit{in vivo} as components of larger molecular machines. Understanding how these high-order interactions drive gene expression presents a substantial statistical challenge. Building on Random Forests (RF), Random Intersection Trees (RITs), and through extensive, biologically inspired simulations, we developed the iterative Random Forest algorithm (iRF). iRF trains a feature-weighted ensemble of decision trees to detect stable, high-order interactions with same order of computational cost as RF.  We demonstrate the utility of iRF for high-order interaction discovery in two prediction problems: enhancer activity in the early \textit{Drosophila} embryo and alternative splicing of primary transcripts in human derived cell lines. In \textit{Drosophila}, among the 20 pairwise transcription factor interactions iRF identifies as stable (returned in more than half of bootstrap replicates), 80\% have been previously reported as physical interactions. Moreover, novel third-order interactions, e.g. between Zelda (Zld), Giant (Gt), and Twist (Twi), suggest high-order relationships that are candidates for follow-up experiments. In human-derived cells, iRF re-discovered a central role of H3K36me3 in chromatin-mediated splicing regulation, and identified novel 5th and 6th order interactions, indicative of multi-valent nucleosomes with specific roles in splicing regulation. By decoupling the order of interactions from the computational cost of identification, iRF opens new avenues of inquiry into the molecular mechanisms underlying genome biology.  
\end{abstract}

\input{sections-pnas/01-intro-background}

\input{sections-pnas/02-method}
\input{sections-pnas/03-simulation_summary}
\input{sections-pnas/04-a-drosophila_impsamp}
\input{sections-pnas/04-b-splicing_impsamp}
\input{sections-pnas/05-discussion}

\section*{Acknowledgments}

This research was supported in part by grants NHGRI U01HG007031, ARO W911NF1710005, ONR N00014-16-1-2664, DOE DE-AC02-05CH11231, NHGRI R00 HG006698, DOE (SBIR/STTR) Award DE-SC0017069, DOE DE-AC02-05CH11231, and NSF DMS-1613002. We thank the Center for Science of Information (CSoI), a US NSF Science and Technology Center, under grant agreement CCF-0939370. Research reported in this publication was supported by the National Library Of Medicine of the NIH under Award Number T32LM012417. The content is solely the responsibility of the authors and does not necessarily represent the official views of the NIH. BY acknowledges support from the Miller Institute for her Miller Professorship in 2016-2017. SB acknowledges the support of UC Berkeley and LBNL, where he conducted most of his work on this paper as a postdoc. We thank P. Bickel and S. Shrotriya for helpful discussions and comments, T. Arbel for preparing $Drosophila$ dataset, and S. Celniker for help vetting the $Drosophila$ data and for consultation on TF interactions.

\newpage
\setcounter{figure}{0}
\setcounter{section}{0}
\renewcommand\thefigure{S\arabic{figure}}
\renewcommand\thesection{S\arabic{section}}  
\renewcommand\thetable{S\arabic{table}}  
\begin{titlepage}
	 \vspace*{\fill}
	 \begin{center}
	 {\Large iterative Random Forests to discover \\predictive and stable high-order interactions}\par
	 \vspace{0.75in}
 	 {\LARGE \textbf{Supporting Information Appendix} \par}
	 \vspace{3.5in}
	 Sumanta Basu, Karl Kumbier, James B. Brown, and Bin Yu
	 \end{center}
	 \vspace*{\fill}
\end{titlepage}

\input{sections-pnas/supp-irf}

\input{sections-pnas/supp-processing}
\input{sections-pnas/supp-simulation_impsample}

\input{sections-pnas/supp-runtime}
\input{sections-pnas/supp-data}
\input{sections-pnas/supp-figures}

\clearpage

\bibliographystyle{abbrvnat}
\bibliography{sections-pnas/biblio-supp}
\end{document}

%% file: sections-pnas/01-intro-background.tex
\section{Introduction}\label{sec:intro}

High throughput, genome-wide measurements of protein-DNA and protein-RNA interactions are driving new insights into the principles of functional regulation. For instance, databases generated by the Berkeley \textit{Drosophila} Transcriptional Network Project (\href{http://bdtnp.lbl.gov/Fly-Net/}{BDTNP}) and \href{http://encodeproject.org}{ENCODE} consortium provide maps of transcription factor (TF) binding events and chromatin marks for substantial fractions of the regulatory factors active in the model organism \textit{Drosophila melanogaster} and human-derived cell lines respectively \citep{fisher2012dna, thomas2011dynamic, li2008transcription, breeze2016eforge, hoffman2012integrative, encode2012integrated}. A central challenge with these data lies in the fact that ChIP-seq, the principal tool used to measure DNA-protein interactions, assays a single protein target at a time. In well studied systems, regulatory factors such as TFs act in concert with other chromatin-associated and RNA-associated proteins, often through stereospecific interactions \citep{hoffman2012integrative, dong2012modeling}, and for a review see \citep{hota2016atp}. While several methods have been developed to identify interactions in large genomics datasets, for example \citep{zhou2014global, lundberg2016chromnet, yoshida2017sparse}, these approaches either focus on pairwise relationships or require explicit enumeration of higher-order interactions, which becomes computationally infeasible for even moderate-sized datasets. In this paper, we present a computationally efficient tool for directly identifying high-order interactions in a supervised learning framework. We note that the interactions we identify do not necessarily correspond to biomolecular complexes or physical interactions. However, among the pairwise $Drosophila$ TF interactions identified as stable, $80\%$ have been previously reported (SI S4). The empirical success of our approach, combined with its computational efficiency, stability, and interpretability, make it uniquely positioned to guide inquiry into the high-order mechanisms underlying functional regulation.

Popular statistical and machine learning methods for detecting interactions
among features include decision trees and their ensembles: CART
\citep{breiman1984classification}, Random Forests (RFs) \citep{breiman2001random}, Node Harvest \citep{meinshausen2010node}, Forest Garotte \citep{meinshausen2009forest}, and Rulefit3 \citep{friedman2008predictive}, as well as methods more specific to gene-gene interactions with categorical features: logic regression \citep{ruczinski2001sequence}, Multifactor Dimensionality Reduction \citep{ritchie2001multifactor}, and Bayesian Epistasis mapping \citep{zhang2007bayesian}. With the exception of RFs, the above tree-based procedures grow shallow trees to prevent overfitting, excluding the possibility of detecting high-order interactions without affecting predictive accuracy. RFs are an attractive alternative, leveraging high-order interactions to obtain state-of-the-art prediction accuracy. However, interpreting interactions in the resulting tree ensemble remains a challenge.

We take a step towards overcoming these issues by proposing a fast algorithm built on RFs that searches for stable, high-order interactions. Our method, the iterative Random Forest algorithm (iRF), sequentially grows feature-weighted RFs to perform soft dimension reduction of the feature space and stabilize  decision paths. We decode the fitted RFs using a generalization of the Random Intersection Trees algorithm (RIT) \citep{shah2014random}. This procedure identifies high-order feature combinations that are prevalent on the RF decision paths. In addition to the high predictive accuracy of RFs, the decision tree base learner captures the underlying biology of local, combinatorial interactions \citep{li2012extensive},  an important feature for biological data, where a single molecule often performs many roles in various cellular contexts. Moreover, invariance of decision trees to monotone transformations \citep{breiman1984classification} to a large extent mitigates normalization issues that are a major concern in the analysis of genomics data, where signal-to-noise ratios vary widely even between biological replicates \citep{landt2012chip, li2011measuring}. Using empirical and numerical examples, we show that iRF is competitive with RF in terms of predictive accuracy, and extracts both known and compelling, novel interactions in two motivating biological problems in epigenomics and transcriptomics.  An open source \texttt{R} implementation of iRF is available through  \href{https://cran.r-project.org/web/packages/iRF/index.html}{CRAN} \citep{basu2017iRF}.

%% file: sections-pnas/02-method.tex
\section{Our method: iterative Random Forests}\label{sec:method}
The iRF algorithm searches for high-order feature interactions in three steps. First, iterative feature re-weighting adaptively regularizes RF fitting. Second, decision rules extracted from a feature-weighted RF map from continuous or categorical to binary features.  This mapping allows us to identify prevalent interactions in RF through a generalization of RIT, a computationally efficient algorithm that searches for high-order interactions in binary data \citep{shah2014random}. Finally, a bagging step assesses the stability of recovered interactions with respect to the bootstrap-perturbation of the data. We briefly review feature-weighted RF and RIT before presenting iRF.

\subsection{Preliminaries: Feature-weighted RF and RIT}
To reduce the dimensionality of the feature space without removing marginally unimportant features that may participate in high-order interactions, we use a feature-weighted version of RF. Specifically, for a set of non-negative weights $w = (w_1, \ldots, w_p)$, where $p$ is the number of features, let $RF(w)$ denote a feature-weighted RF constructed with $w$. In $RF(w)$, instead of taking a uniform random sample of features at each split, one chooses the $j^{th}$ feature with probability proportional to $w_j$. Weighted tree ensembles have been proposed in \citep{amaratunga2008enriched} under the name ``enriched random forests'' and used for feature selection in genomic data analysis. Note that with this notation, Breiman's original RF amounts to $RF(1/p, \ldots, 1/p)$.

iRF build upon a generalization of RIT, an algorithm that performs a randomized search for high-order interactions among binary features in a deterministic setting. More precisely, RIT searches for co-occurring collections of $s$ binary features, or order-$s$ interactions, that appear with greater frequency in a given class. The algorithm recovers such interactions with high probability (relative to the randomness it introduces) at a substantially lower computational cost than $O(p^s)$, provided the interaction pattern is sufficiently prevalent in the data and individual features are sparse.  We briefly present the basic RIT algorithm and refer readers to the original paper \citep{shah2014random} for a complete description. 

Consider a binary classification problem with $n$ observations and $p$ binary features.
Suppose we are given data in the form $(\mathcal{I}_i, Z_i)$, $i=1, \ldots, n$.
Here, each $Z_i \in \{0,1\}$ is a binary label and $\mathcal{I}_i\subseteq \{1, 2, \ldots, p\}$ is a feature-index subset indicating the indices of ``active'' features associated with observation $i$. In the context of gene transcription,  $\mathcal{I}_i$ can be thought of as a
collection of TFs and histone modifications with abnormally high or low
enrichments near the $i^{th}$ gene's promoter region, and $Z_i$ can indicate
whether gene $i$ is transcribed or not. With these notations, prevalence of an
interaction $S \subseteq \{1, \ldots, p\}$ in the class $C \in \{0, 1\}$ is
defined as 
\begin{equation*}
\mathbb{P}_n (S | Z=C) := \frac{\sum_{i=1}^n \mathds{1}(S \subseteq \mathcal{I}_i)}{\sum_{i=1}^n \mathds{1}(Z_i = C)},
\end{equation*}
where $\mathbb{P}_n$ denotes the empirical probability distribution and $\mathds{1}(\cdot)$ the indicator function. For given thresholds $0 \le \theta_0 < \theta_1 \le 1$, RIT performs a randomized search for interactions $S$ satisfying 
\begin{equation}\label{eqn:rit-defn}
\mathbb{P}_n (S | Z = 1) \ge \theta_1, ~~~~ \mathbb{P}_n (S| Z=0) \le \theta_0.
\end{equation}

For each class $C\in \{0, 1\}$ and a pre-specified integer $D$, let
$j_1, ..., j_D$ be randomly chosen indices from the set of observations
$\{i:Z_{i}=C\}$. To search for interactions $S$ satisfying condition
(\ref{eqn:rit-defn}), RIT takes $D$-fold intersections $\mathcal{I}_{j_1} \cap
\mathcal{I}_{j_2} \cap \ldots \cap  \mathcal{I}_{j_D}$ from the randomly
selected observations in class $C$. To reduce computational complexity, these
interactions are performed in a tree-like fashion (SI S1 Algorithm 1), where each non-leaf node has $n_{child}$ children. 
This process is repeated $M$ times for a given class $C$, resulting in a collection of survived interactions $\mathcal{S}=\bigcup_{m=1}^M\mathcal{S}_m$, where each $\mathcal{S}_m$ is the set of interactions that remains following the $D$-fold intersection process in tree $m=1,\ldots,M$. The prevalences of interactions across different classes are subsequently compared using condition (\ref{eqn:rit-defn}). The main intuition is that if an interaction $S$ is highly prevalent in a particular class, it will survive the $D$-fold intersection with high probability.

\subsection{iterative Random Forests}
The iRF algorithm places interaction discovery in a supervised learning framework to identify class-specific, active index sets required for RIT. This framing allows us to recover high-order interactions that are associated with accurate prediction in feature-weighted RFs.

We consider the binary classification setting with training data $\mathcal{D}$ in the form $\{(\mathbf{x}_i, y_i)\}_{i=1}^n$, with continuous or categorical features $\mathbf{x} = (x_1,\dots,x_p)$, and a binary label $y \in \{0, 1\}$. Our goal is to find subsets $S \subseteq \{1, \ldots, p\}$ of features, or interactions, that are both highly prevalent within a class $C \in \{0,1\}$, and that provide good differentiation between the two classes. To encourage generalizability of our results, we search for interactions in ensembles of decision trees fitted on bootstrap samples of $\mathcal{D}$. This allows us to identify interactions that are robust to small perturbations in the data. Before describing iRF, we present a generalized RIT that uses any RF, weighted or not, to generate active index sets from continuous or categorical features. Our generalized RIT is independent of the other iRF components in the sense that other approaches could be used to generate the input for RIT. We remark on our particular choices in SI S2.

\textbf{Generalized RIT (through an RF):} For each tree $t=1,\dots,T$ in the
output tree ensemble of an RF, we collect all leaf nodes and index them by
$j_t=1,..., J(t)$. Each feature-response pair $(\mathbf{x}_i, y_i)$ is
represented with respect to a tree $t$ by $(\mathcal{I}_{i_t}, Z_{i_t})$, where 
$\mathcal{I}_{i_t}$ is the set of unique feature indices falling on the path of the leaf node containing $(\mathbf{x}_i, y_i)$ in the $t^{th}$ tree.
Hence, each $(\mathbf{x}_i, y_i)$ produces $T$  such index set and label pairs, corresponding to the $T$ trees. We aggregate these pairs across observations and trees as
\begin{equation}
\label{eqn:ruleset}
\mathcal{R} = \{(\mathcal{I}_{i_t}, Z_{i_t}): \mathbf{x}_i 
\text{ falls in leaf node } i_t \text{ of tree } t\}
\end{equation}
and apply RIT on this transformed dataset $\mathcal{R}$ to obtain a set of interactions.

\begin{figure}[h]
 \centering
 \includegraphics[scale=0.3]{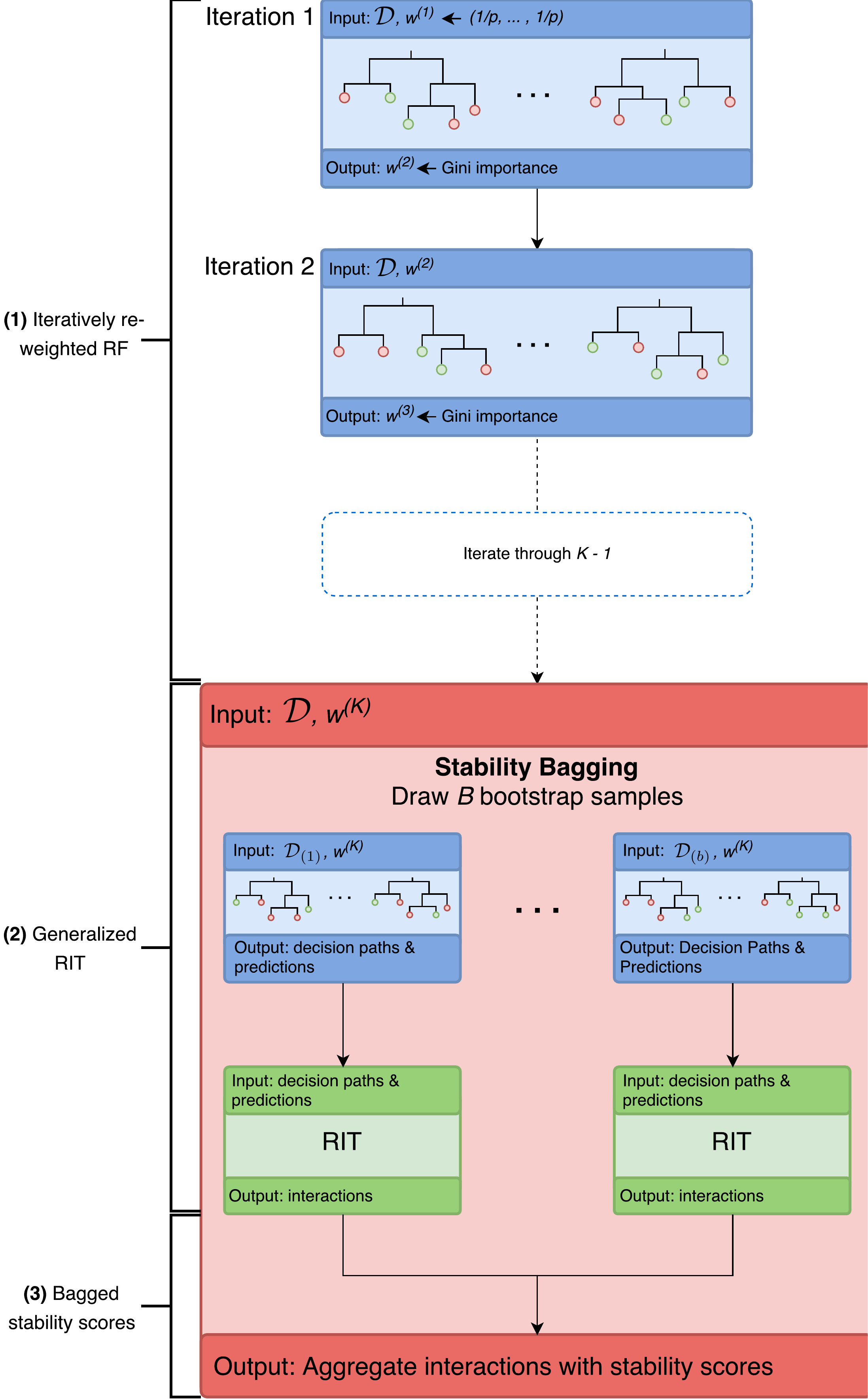}
 \caption{iRF workflow. Iteratively re-weighted RF (blue boxes)
 are trained on full data $\mathcal{D}$ and pass Gini importance as weights to
 the next iteration. In iteration $K$ (red box), feature-weighted RF are grown using
 $w^{(K)}$ on $B$ bootstrap samples of the full data
 $\mathcal{D}_{(1)},\dots,\mathcal{D}_{(B)}$. Decision paths and predicted
leaf node labels are passed to RIT (green box) which computes prevalent
 interactions in the RF ensemble. Recovered interactions are scored for stability across (outer-layer) bootstrap samples.\label{workflow}}
 \end{figure}

We now describe the three components of iRF. A depiction is shown in Fig. \ref{workflow} and the complete workflow is presented in SI S1 Algorithm 2. We remark on the algorithm further in SI S2.

\textbf{1. Iteratively re-weighted RF:} Given an iteration number $K$, iRF
iteratively grows $K$ feature-weighted RFs $RF(w^{(k)})$, $k=1,\dots,K$, on
the data $\mathcal{D}$. The first iteration of iRF ($k=1$) starts with $w^{(1)}
:= (1/p, \ldots, 1/p)$, and stores the importance (mean decrease in Gini
impurity) of the $p$ features as $v^{(1)} = (v^{(1)}_1, \ldots, v^{(1)}_p)$. For
iterations $k=2,\dots,K$, we set $w^{(k)} = v ^ {(k-1)}$ and grow a
weighted RF with weights set equal to the RF feature importance from the
previous iteration. Iterative approaches for fitting RFs have been previously proposed in \citep{anaissi2013balanced} and combined with hard thresholding to select features in microarray data.

\textbf{2. Generalized RIT (through $RF(w^{(K)})$):} We apply generalized
RIT to the last feature-weighted RF grown in iteration $K$. That is, decision rules generated in the process of fitting
$RF(w^{(K)})$ provide the mapping from continuous or categorical to binary features required
for RIT. This process produces a collection of interactions $\mathcal{S}$.

\textbf{3. Bagged stability scores:} In addition to bootstrap sampling
in the weighted RF, we use an ``outer layer'' of bootstrapping to assess the
stability of recovered interactions. We generate bootstrap samples of the data
$\mathcal{D}_{(b)}, b=1,\dots,B$, fit $RF(w^{(K)})$ on
each bootstrap sample $\mathcal{D}_{(b)}$, and use generalized RIT to
identify interactions $\mathcal{S}_{(b)}$ across each bootstrap sample. We
define the \textit{stability score} of an interaction  $S \in \cup_{b=1}^{B}
\mathcal{S}_{(b)}$ as 
\begin{equation*}
\small{sta(S) = \frac{1}{B} \cdot \sum_{b=1}^B \mathds{1} \{S \in
\mathcal{S}_{(b)}\}},
\end{equation*}
representing the proportion of times (out of $B$ bootstrap
samples) an interaction appears as an output of RIT. This averaging step is
exactly the Bagging idea of Breimain \citep{breiman1996bagging}.

\subsection{iRF tuning parameters} 
The iRF algorithm inherits tuning parameters from its two base algorithms, RF and RIT. The predictive performance of RF is known to be highly resistant to choice of parameters \citep{breiman2001random}, so we use the default parameters in the \texttt{R} \texttt{randomForest} package. Specifically, we set the number of trees $\texttt{ntree}=500$, the number of variables sampled at each node $\texttt{mtry}=\sqrt{p}$, and grow trees to purity. For the RIT algorithm, we use the basic version or Algorithm 1 of \citep{shah2014random}, and grow $M = 500$ intersection trees of depth $D = 5$ with $n_{child} = 2$, which empirically leads to a good balance between computation time and quality of recovered interactions. We find that both prediction accuracy and interaction recovery of iRF are fairly robust to these parameter choices (SI S2.5).

In addition to the tuning parameters of RF and RIT, the iRF workflow introduces two additional tuning parameters: (i) number of bootstrap samples $B$ (ii) number of iterations $K$. Larger values of $B$ provide a more precise description of the uncertainty associated with each interaction at the expense of increased computation cost. In our simulations and case studies we set $B \in(10, 30)$ and find that results are qualitatively similar in this range. The number of iterations controls the degree of regularization on the fitted RF. We find that the quality of recovered interactions can improve dramatically for $K>1$ (SI S5). In sections \ref{sec:drosophila} and \ref{sec:splicing}, we report interactions with $K$ selected by $5-$fold cross validation. 

%% file: sections-pnas/03-simulation_summary.tex
\section{Simulation experiments}
We developed and tested iRF through extensive simulation studies based on biologically inspired generative models using both synthetic and real data (SI S5). In particular, we generated responses using Boolean rules intended to reflect the stereospecific nature of interactions among biomolecules \citep{nelson2008lehninger}. In total, we considered 7 generative models built from AND, OR, and XOR rules, with number of observations and features ranging from $100$ to $5000$ and $50$ to $2500$ respectively. We introduced noise into our models both by randomly swapping response labels for up to $30\%$ of observations and through RF-derived rules learned on held-out data.

We find that the predictive performance of iRF ($K > 1$) is generally comparable with RF ($K=1$). However, iRF recovers the full data generating rule, up to an order-8 interaction in our simulations, as the most stable interaction in many settings where RF rarely recovers interactions of order $>2$. The computational complexity of recovering these interactions is substantially lower than competing methods that search for interactions incrementally  (SI S6; SI Fig. S18).

Our experiments suggest that iterative re-weighting encourages iRF to use a stable set of features on decision paths (SI Fig. S9). Specifically, features that are identified as important in early iterations tend to be selected among the first several splits in later iterations (SI Fig. S10). This allows iRF to generate partitions of the feature space where marginally unimportant, active features become conditionally important, and thus more likely to be selected on decision paths. For a full description of simulations and results, see SI S5.

%% file: sections-pnas/04-a-drosophila_impsamp.tex
\section{Case study I: enhancer elements in \textit{Drosophila}}\label{sec:drosophila}

\noindent Development and function in multicellular organisms rely on precisely regulated spatio-temporal gene expression. Enhancers play a critical role in this process by coordinating combinatorial TF binding, whose integrated activity leads to patterned gene expression during embryogenesis \citep{levine2010transcriptional}. In the early \textit{Drosophila} embryo, a small cohort of $\sim$40 TFs drive patterning, for a review see \citep{rivera1996gradients}, providing a well-studied, simplified model system in which to investigate the relationship between TF binding and enhancer activities. Extensive work has resulted in genome-wide, quantitative maps of DNA occupancy for 23 TFs \citep{macarthur2009developmental} and 13 histone modifications \citep{encode2012integrated}, as well as labels of enhancer status for $7809$ genomic sequences in blastoderm (stage 5) \textit{Drosophila} embryos \cite{fisher2012dna, berman2002exploiting}. See SI S3 for descriptions of data collection and preprocessing.

\begin{figure}[H]
\textbf{A} \hspace{0.375\linewidth} \textbf{B}\\
\includegraphics[width=0.375\linewidth]{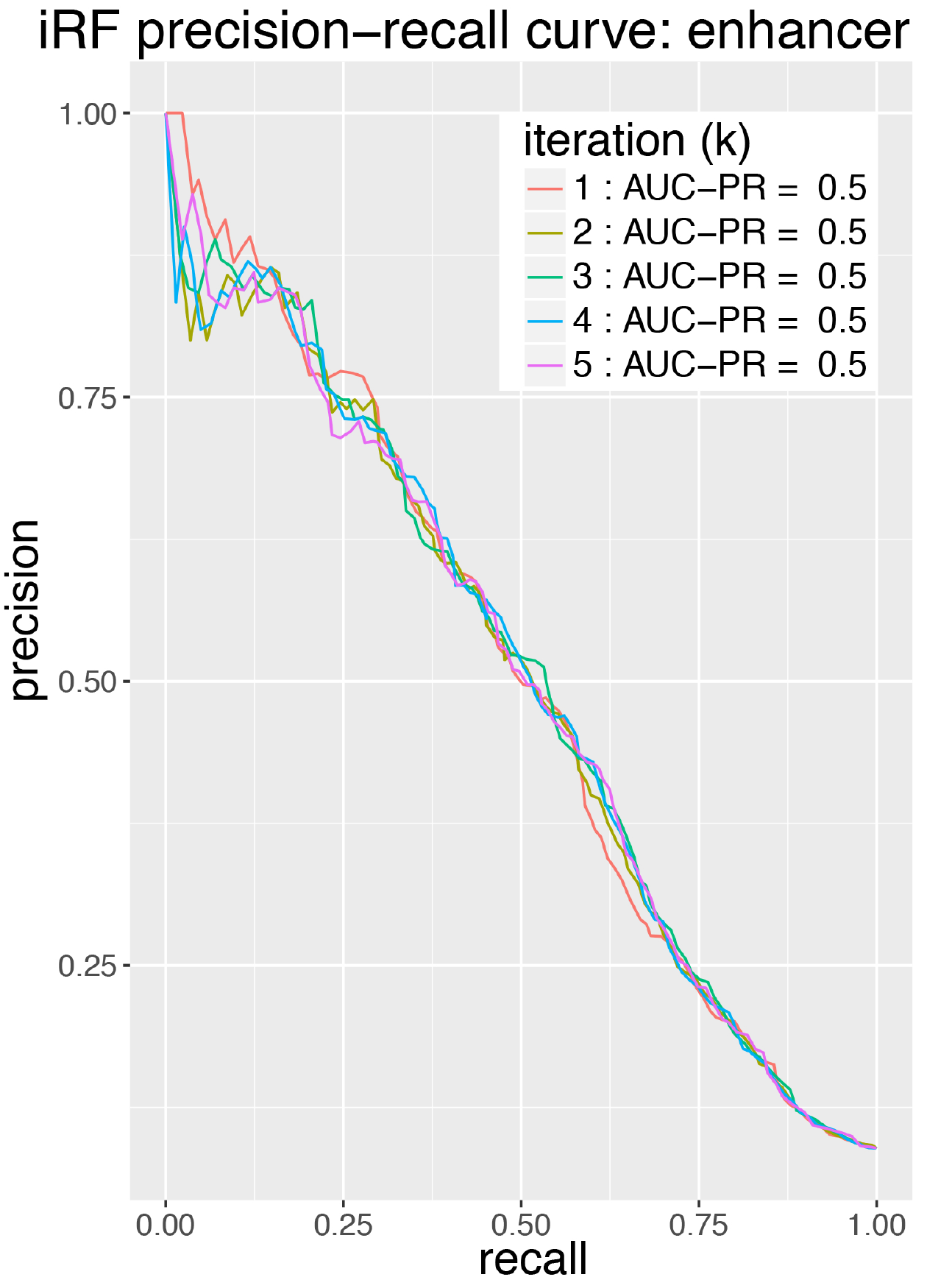}
\includegraphics[width=0.6\linewidth]{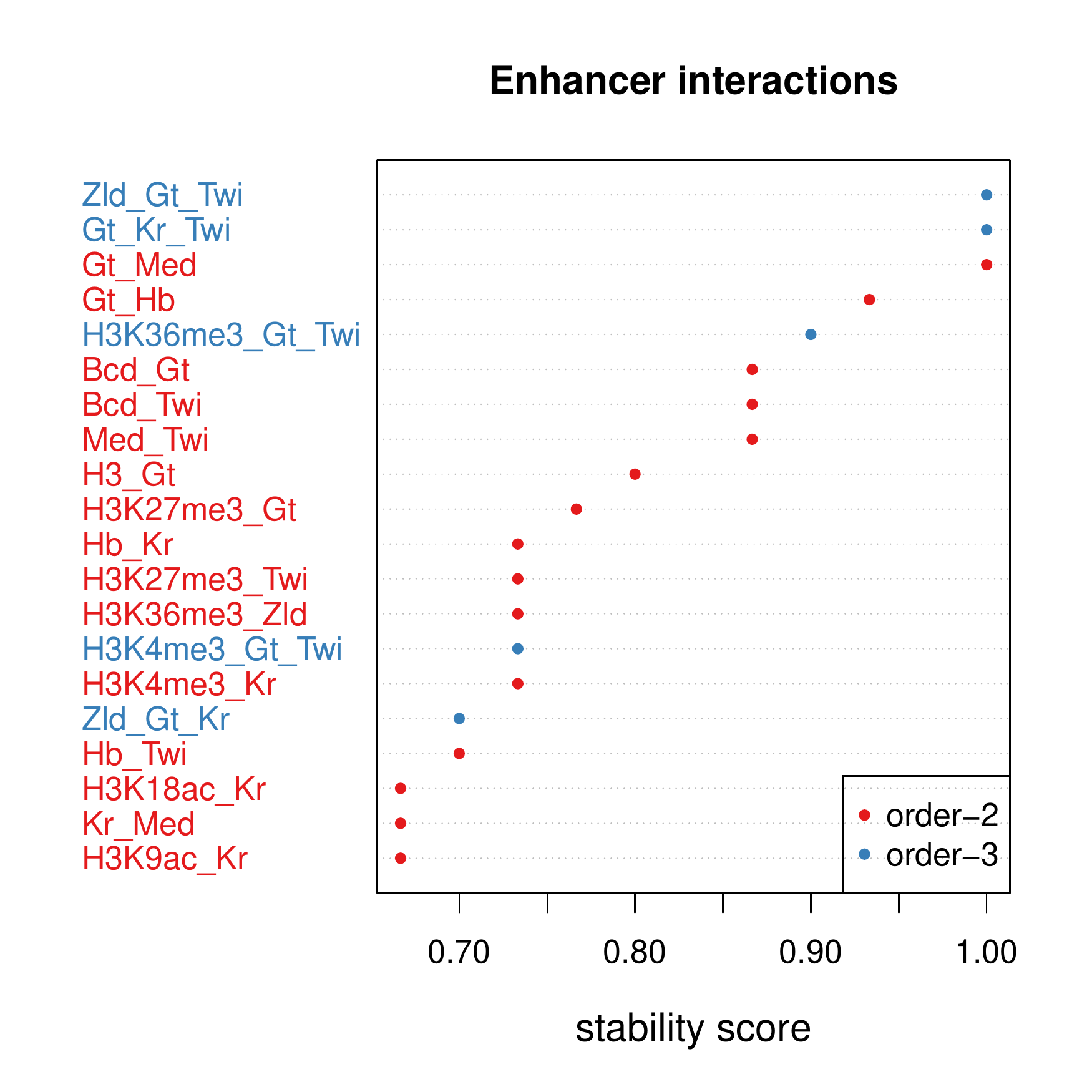} \\
\textbf{C} \hspace{0.475\linewidth} \textbf{D} \\
\includegraphics[width=0.5\linewidth]{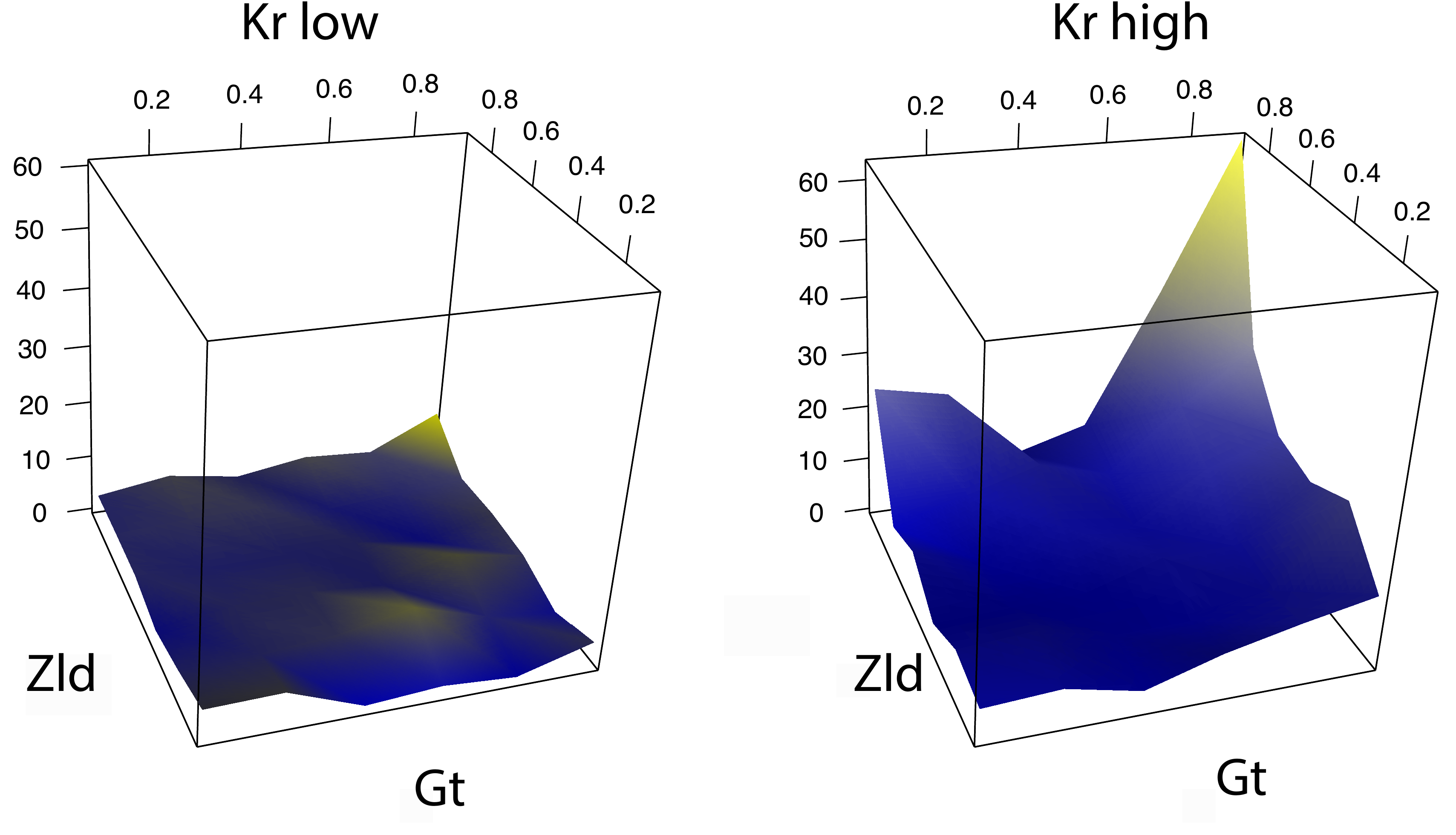}
\includegraphics[width=0.5\linewidth]{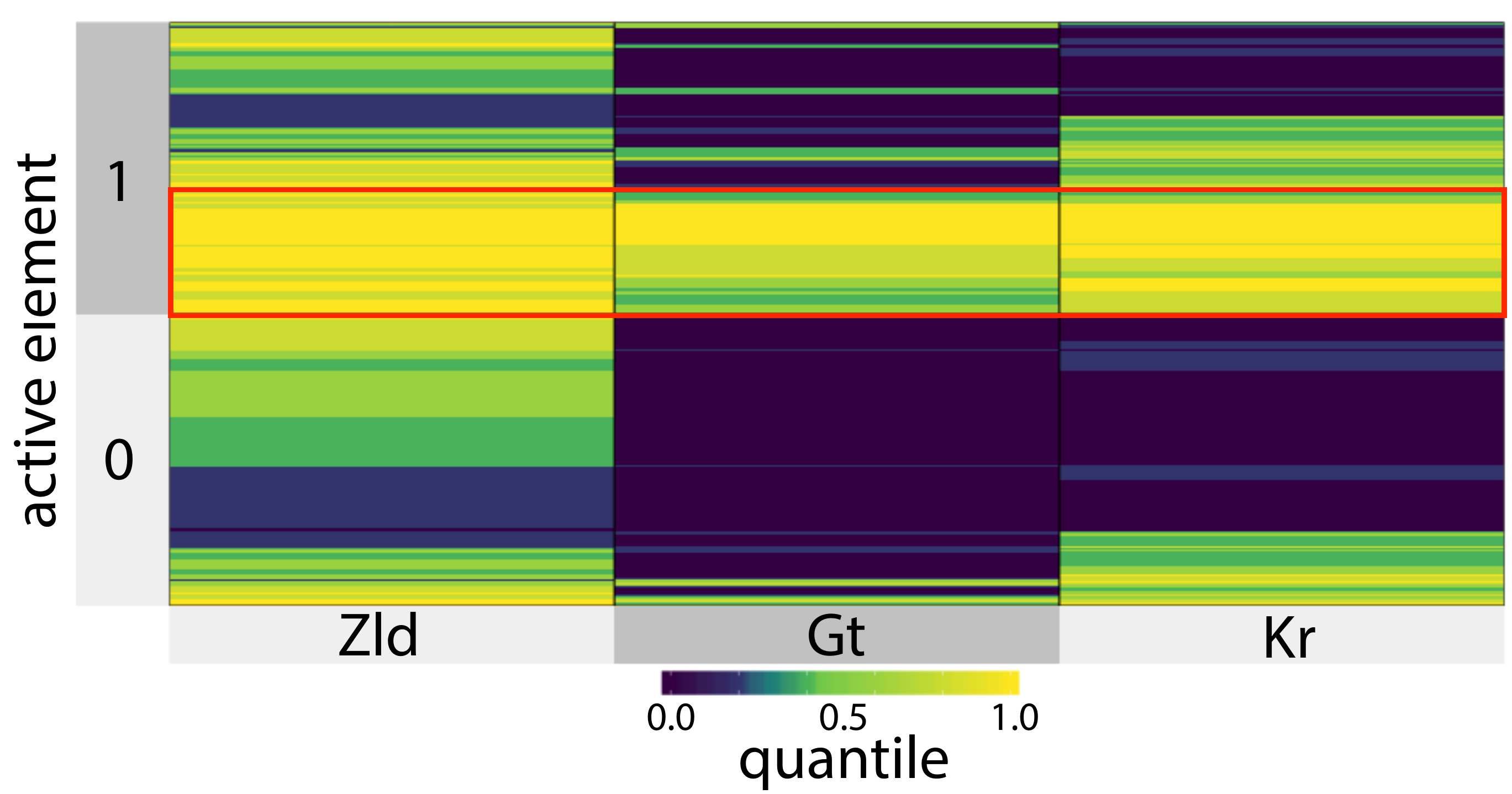}
\caption{\textbf{[A]}: Accuracy of iRF (AUC-PR) in predicting active elements from TF binding
and histone modification data. \textbf{[B]}: 20 most stable interactions recovered by iRF after $5$ iterations. Interactions that are a strict subset of another interaction with stability score $\ge 0.5$ have been removed for cleaner visualization. iRF recovers known
interactions among Gt, Kr and Hb and interacting roles of master regulator Zld. \textbf{[C]}: Surface maps demonstrating the proportion of active enhancers by quantiles of Zld, Gt, and Kr binding (held-out test data). On the subset of data where Kr binding is lower than the median Kr level, proportion of active enhancers does not change with Gt and Zld. On the subset of data with Kr binding above the median level, structure of the response surface reflects an order-3 AND interaction: increased levels of Zld, Gt, and Kr binding are indicative of enhancer status for a subset of observations. 
\textbf{[D]}: Quantiles of Zld, Gt, and Kr binding grouped by enhancer status (balanced sample of held-out test data). The block of active elements highlighted in red represents the subset of observations for which the AND interaction is active. \label{irf-taly-tf-histone}}
\end{figure}

To investigate the relationship between enhancers, TF binding, and chromatin state, we used iRF to predict enhancer status for each of the genomic sequences ($3912$ training, $3897$ test). We achieved an area under the precision-recall curve (AUC-PR) on the held-out test data of 0.5 for $K=5$ (Fig. \ref{irf-taly-tf-histone}A). This corresponds to a Matthews correlation coefficient (MCC) of $0.43$ (positive predictive value (PPV) of $0.71$) when predicted probabilities are thresholded to maximize MCC in the training data.

Fig. \ref{irf-taly-tf-histone}B reports stability scores of recovered interactions for $K=5$. We note that the data analyzed are whole-embryo and interactions found by iRF do not necessarily represent physical complexes. However, for the well-studied case of pairwise TF interactions, 80\% of our findings with stability score $> 0.5$ have been previously reported as physical (Table S1). For instance, interactions among gap proteins Giant (Gt), Kr{\"u}ppel (Kr), and Hunchback (Hb),  some of the most well characterized interactions in the early \textit{Drosophila} embryo \citep{nusslein1980mutations}, are all highly stable ($sta(\text{Gt}-\text{Kr})=1.0$, $sta(\text{Gt}-\text{Hb})=0.93$, $sta(\text{Hb}-\text{Kr})=0.73$). Physical evidence supporting high-order mechanisms is a frontier of experimental research and hence limited, but our excellent pairwise results give us hope that high-order interactions we identify as stable have a good chance of being confirmed by follow-up work.

iRF also identified several high-order interactions surrounding the early regulatory factor Zelda (Zld) ($sta(\text{Zld}-\text{Gt}-\text{Twi})=1.0$, $sta(\text{Zld}-\text{Gt}-\text{Kr})= 0.7$). Zld has been previously shown to play an essential role during the maternal-zygotic transition \citep{liang2008zinc, harrison2011zelda}, and there is evidence to suggest that Zld facilitates binding to regulatory elements \citep{pmid24909324}. We find that Zld binding in isolation rarely drives enhancer activity, but in the presence of other TFs, particularly the anterior-posterior (AP) patterning factors Gt and Kr, it is highly likely to induce transcription. This generalizes the dependence of Bicoid$-$induced transcription on Zld binding to several of the AP factors \citep{pmid24637116}, and is broadly consistent with the idea that Zld is potentiating, rather than an activating factor \citep{pmid24909324}.

More broadly, response surfaces associated with stable high-order interactions indicate AND-like rules (Fig. \ref{irf-taly-tf-histone}C). In other words, the proportion of active enhancers is substantially higher for sequences where all TFs are sufficiently bound, compared to sequences where only some of the TFs exhibit high levels of occupancy. Fig. \ref{irf-taly-tf-histone}C demonstrates a putative third order interaction found by iRF ($sta(\text{Kr}-\text{Gt}-\text{Zld})=0.7$). To the left, the Gt-Zld response surface is plotted using only sequences for which Kr occupancy is lower than the median Kr level, and the proportion of active enhancers is uniformly low (< 10\%). The response surface to the right, is plotted using only sequences where Kr occupancy is higher than median Kr level and shows that the proportion of active elements is as high as 60\% when both Zld and Gt are sufficiently bound. This points to an order-3 AND rule, where all three proteins are required for enhancer activation in a subset of sequences. In Fig. \ref{irf-taly-tf-histone}D, we show the subset of sequences that correspond to this AND rule (highlighted in red) using a superheat map \citep{barter2015superheat}, which juxtaposes two separately clustered heatmaps corresponding to active and inactive elements. Note that the response surfaces are drawn using held-out test data to illustrate the generalizability of interactions detected by iRF. While overlapping patterns of TF binding have been previously reported \citep{macarthur2009developmental}, to the best of our knowledge this is the first report of an AND-like response surface for enhancer activation. Third-order interactions have been studied in only a handful of enhancer elements, most notably eve stripe 2, for a review see \citep{levine2013computing}, and our results indicate that they are broadly important for the establishment of early zygotic transcription, and therefore body patterning.

%% file: sections-pnas/04-b-splicing_impsamp.tex
\section{Case study II: alternative splicing in a human-derived cell line}\label{sec:splicing}

\noindent In eukaryotes, alternative splicing of primary mRNA transcripts is a highly
regulated process in which multiple distinct mRNAs are produced by the same
gene. In the case of messenger RNAs (mRNAs), the result of this process is the
diversification of the proteome, and hence the library of functional molecules
in cells. The activity of the spliceosome, the ribonucleoprotein responsible for most splicing 
in eukaryotic genomes, is driven by complex, cell-type specific interactions with cohorts of 
RNA binding proteins (RBP)  \citep{so2016u1, stoiber2015extensive}, suggesting that 
high-order interactions play an important role in the regulation of alternative splicing. 
However, our understanding of this system derives from decades of study in genetics, biochemistry, 
and structural biology. Learning interactions directly from genomics 
data has the potential to accelerate our pace of discovery in the study of co- and post-transcriptional gene regulation.

Studies, initially in model organisms, have revealed that the chromatin mark H3K36me3, the DNA binding protein CTCF, and a few other factors all  play splice-enhancing roles \citep{kolasinska2009differential, sims2009processing, kornblihtt2012ctcf}. However, the extent to which chromatin state and DNA binding factors interact $en$ $masse$ to modulate co-transcriptional splicing remains unknown \citep{allemand2016broad}.  To identify interactions that form the basis of chromatin mediated splicing, we used iRF to predict thresholded splicing rates for $23823$ exons (RNA-seq Percent-spliced-in (PSI) values \citep{pervouchine2016ipsa}; $11911$ train, $11912$ test), from ChIP-seq assays measuring enrichment of chromatin marks and TF binding events (253 ChIP assays on 107 unique transcription factors and 11 histone modifications). Preprocessing methods are described in the SI S3.

\begin{figure}[H]
\textbf{A} \hspace{0.3\linewidth} \textbf{B}\\
\includegraphics[width=0.3\linewidth]{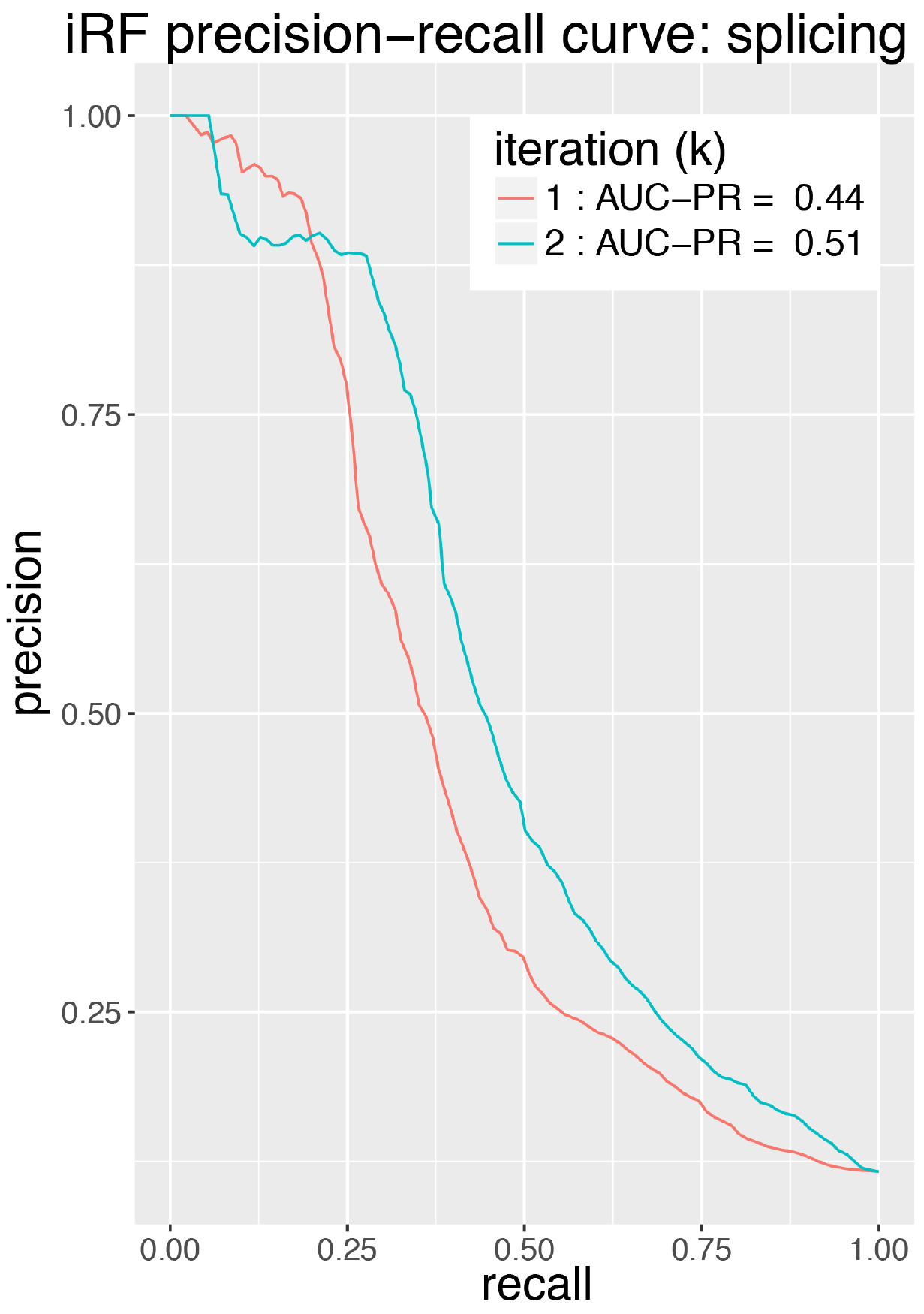}
\includegraphics[width=0.69\linewidth]{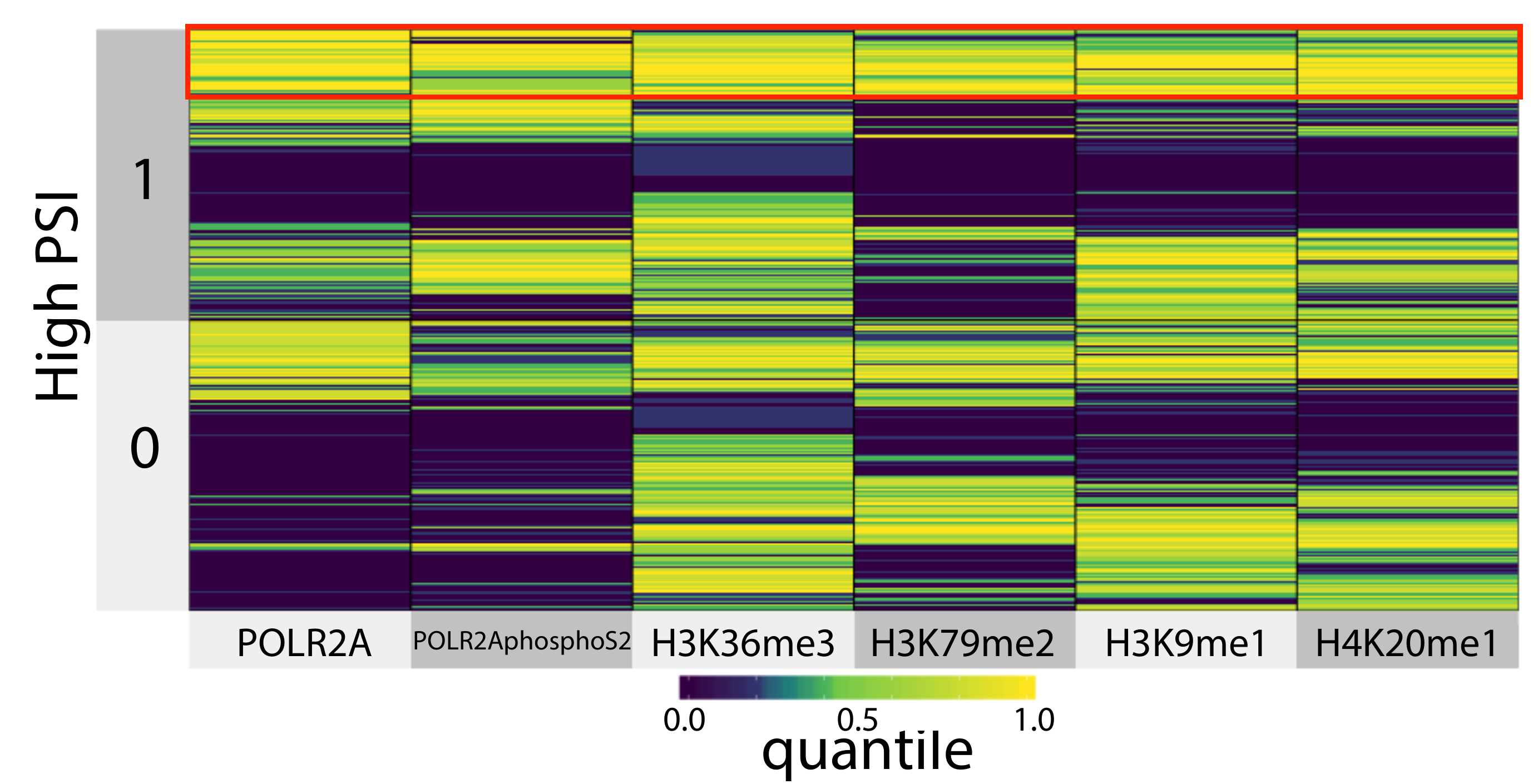}\\
\textbf{C}\\
\vspace{-0.1in}
\begin{center}
\vspace{-0.2in}
\includegraphics[width=0.975\linewidth]{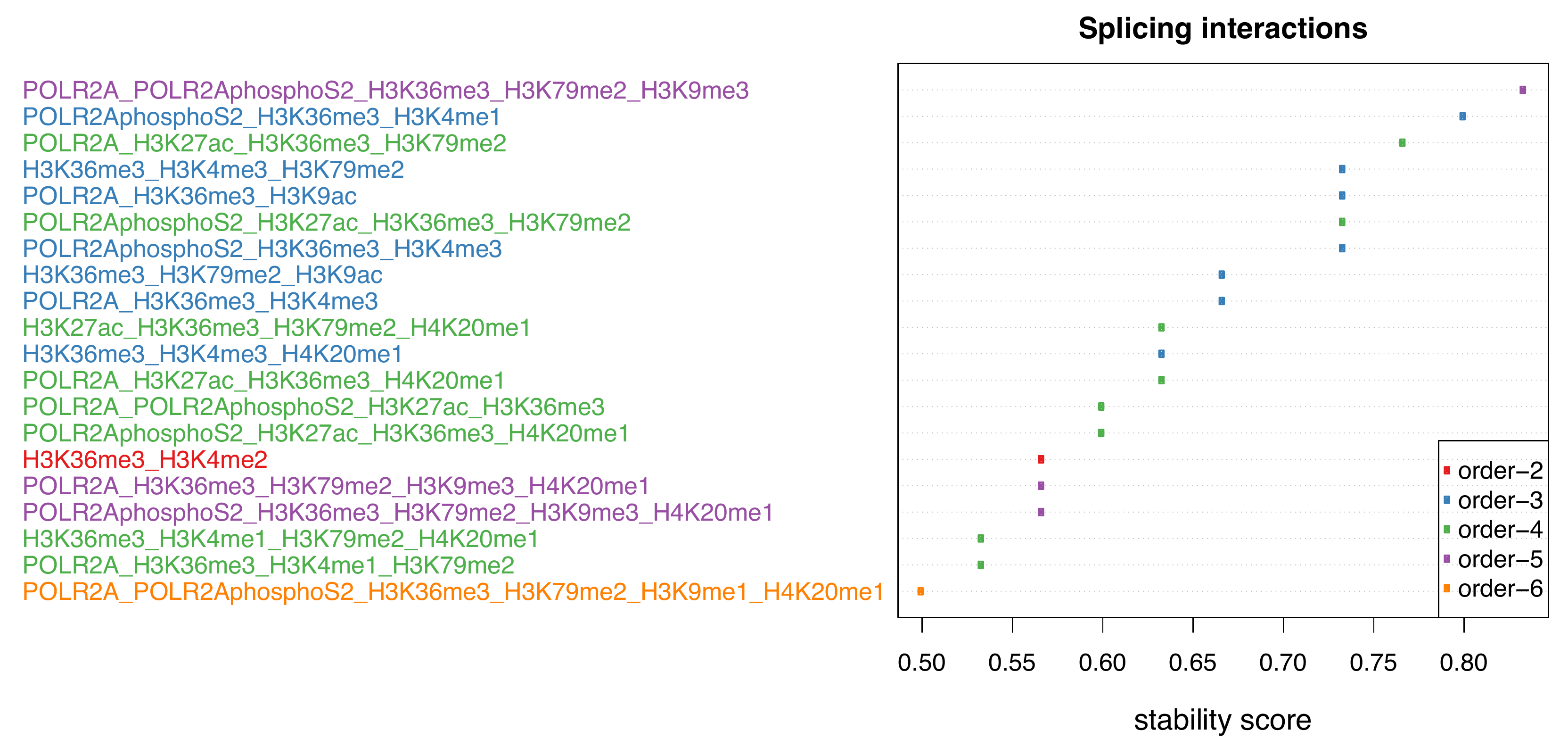} 
\end{center}
\vspace{-0.2in}
\caption{\textbf{[A]}: Accuracy of iRF (AUC-PR) in classifying included exons from
  excluded exons in held-out test data. iRF shows $7\%$ increase in AUC-PR over
  RF.  \textbf{[B]}: An order-6 interaction recovered by iRF (stability score 0.5) displayed on a superheat map which juxtaposes two separately clustered heatmaps of exons with high and low splicing rates. Co-enrichment of all the 6 plotted features reflects an AND-type rule indicative of high splicing rates for the exons highlighted in red (held-out test data). The subset of Pol II, S2 phospho-Pol II, H3K36me3, H3K79me2, and H4K20me1 was recovered as an order-5 interaction in all bootstrap samples (stability score $1.0$). \textbf{[C]}: 20 most stable interactions recovered in the second iteration of iRF. Interactions that are a strict subset of another interaction with stability score $\ge 0.5$ have been removed for cleaner visualization.\label{irf-splicing-tf-histone}}
\end{figure}

In this prediction problem, we achieved an AUC-PR on the held-out test data of 0.51 for $K=2$ (Fig. \ref{irf-splicing-tf-histone}A). This corresponds to a MCC of $0.47$ (PPV $0.72$) on held-out test data when predicted probabilities are thresholded to  maximize MCC in the training data. Fig. \ref{irf-splicing-tf-histone}B reports stability scores of recovered interactions for $K=2$. We find interactions involving H3K36me3, a number of novel interactions involving other chromatin marks, and post-translationally modified states of RNA Pol II. In particular, we find that the impact of serine 2 phosphorylation of Pol II appears highly dependent on local chromatin state. Remarkably, iRF identified an order-$6$ interaction surrounding H3K36me3 and S2 phospho-Pol II (stability score $0.5$, Fig.  \ref{irf-splicing-tf-histone}B,C) along with two highly stable order $5$ subsets of this interaction (stability scores $1.0$). A subset of highly spliced exons highlighted in red is enriched for all 6 of these elements, indicating a potential AND-type rule related to splicing events (Fig. \ref{irf-splicing-tf-histone}C). This observation is consistent with, and offers a quantitative model for the previously reported predominance of co-transcriptional splicing in this cell line \citep{tilgner2012deep}. We note that the search space of order-$6$ interactions is $>$ $10^{11}$, and that this interaction is discovered with an order-zero increase over the computational cost of finding important features using RF. Recovering such interactions without exponential speed penalties represents a substantial advantage over previous methods and positions our approach uniquely for the discovery of complex, nonlinear interactions.

%% file: sections-pnas/05-discussion.tex
\section{Discussion}\label{sec:discussion}
Systems governed by nonlinear interactions are ubiquitous in biology. We developed a predictive and stable method, iRF,  for learning such feature interactions. iRF identified known and novel interactions in early zygotic enhancer activation in the $Drosophila$ embryo, and posit new high-order interactions in splicing regulation for a human-derived system.

Validation and assessment of complex interactions in biological systems is necessary and challenging, but new wet-lab tools are becoming available for targeted genome and epigenome engineering. For instance, the CRISPR system has been adjusted for targeted manipulation of post-translational modifications to histones \citep{hilton2015epigenome}. This may allow for tests to determine if modifications to distinct residues at multivalent nucleosomes function in a non-additive fashion in splicing regulation. Several of the histone marks that appear in the interactions we report, including H3K36me3 and H4K20me1, have been previously identified \citep{pmid9684600} as essential for establishing splicing patterns in the early embryo. Our findings point to direct interactions between these two distinct marks. This observation generates interesting questions: What proteins, if any, mediate these dependencies? What is the role of Phospho-S2 Pol II in the interaction? Proteomics on ChIP samples may help reveal the complete set of factors involved in these processes, and new assays such as Co-ChIP may enable the mapping of multiple histone marks at single-nucleosome resolution \citep{weiner2016co}.

We have offered evidence that iRF constitutes a useful tool for generating new hypotheses from the study of high-throughput genomics data, but many challenges await. iRF currently handles data heterogeneity only implicitly, and the order of detectable interaction depends directly on the depth of the tree, which is on the order of $\log_2(n)$.  We are currently investigating local importance measures to explicitly relate discovered interactions to specific observations. This strategy has the potential to further localize feature selection and improve the interpretability of discovered rules. Additionally, iRF does not distinguish between interaction forms, for instance additive versus non-additive. We are exploring tests of rule structure to provide better insights into the precise form of rule-response relationships. 

To date, machine learning has been driven largely by the need for accurate prediction. Leveraging machine learning algorithms for scientific insights into the mechanics that underlie natural and artificial systems will require an understanding of why prediction is possible. The Stability Principle, which asserts that statistical results should at a minimum be reproducible across reasonable data and model perturbations, has been advocated in \citep{yu2013stability} as a second consideration to work towards understanding and interpretability in science.  Iterative and data-adaptive regularization procedures such as iRF are based on prediction and stability and have the potential to be widely adaptable to diverse algorithmic and computational architectures, improving interpretability and informativeness by increasing the stability of learners.

%% file: sections-pnas/supp-irf.tex
\section{Algorithms}
The basic versions of the Random Intersection Trees (RIT)  and iterative Random Forests (iRF) algorithms are presented below. For a complete description of RIT, including analysis of computational complexity and theoretical guarantees, we refer readers to the original paper \citep{shah2014random}. For a full description of iRF, we refer readers to Section 2.

\begin{algorithm}[H]
\SetAlgoLined
\DontPrintSemicolon

\KwIn{$\{(\mathcal{I}_i, Z_i);  \mathcal{I}_i\subseteq \{1, \ldots, p \}, Z_i \in \{0, 1\} \}_{i=1}^{n}, C \in \{0, 1\}$\; \textbf{Tuning Parameters: }
$(D, M, n_{child})$ }

\For{ tree $m \gets 1$ \textbf{to} $M$}{

Let $m$ be a tree of depth $D$, with each node $j$ in levels $0,\dots,D-1$ having $n_{child}$ children, and denote the parent of node $j$ as $pa(j)$. Let $J$ be the total number of nodes in the tree, and index the nodes such that for every parent-child pair, larger indices are assigned to the child than the parent. For each node $j=1,\dots,J$, let $i_j$ be a uniform sample from the set of class $C$ observations $\{i:Z_{i}=C\}$.

Set $S_1=\mathcal{I}_{i_1}$

\For{$j = 2$ \textbf{to} $J$}{
$S_j \gets \mathcal{I}_{i_j} \cap S_{pa(j)}$
}
\Return{$\mathcal{S}_m=\{S_j: depth(j) = D\}$}\;
}
\KwOut{$\mathcal{S}=\cup_{m=1}^M \mathcal{S}_m$}
\caption{Random Intersection Trees \citet{shah2014random}}
\label{algo:rit}
\end{algorithm}
\smallskip

\begin{algorithm}[H]
\KwIn{$\mathcal{D}$, $C \in \{0, 1\}$, $B$, $K$, $w^{(1)} \gets (1/p, \ldots, 1/p)$}
\textbf{(1) } \For{$k \gets 1$ \textbf{to} $K$}{
	 Fit $RF(w^{(k)})$ on $\mathcal{D}$\\
	 $w^{(k+1)} \gets $ Gini importance of $RF(w^{(k)})$\\
	 }
\textbf{(2)} \For{$b \gets 1$ \textbf{to} $B$}{
	Generate bootstrap samples $\mathcal{D}_{(b)}$ of the form $\{\mathbf{x}_{b(i)}, y_{b(i)}\}$ from $\mathcal{D}$\\
	Fit $RF(w^{(K)})$ on $\mathcal{D}_{(b)}$\\
	$\mathcal{R}_{(b)} \gets$ $\{(\mathcal{I}_{i_t}, Z_{i_t}): \mathbf{x}_{b(i)} 
\text{ falls in leaf node } i_t \text{ of tree } t\}$\\
	$\mathcal{S}_{(b)} \gets $ RIT($\mathcal{R}_{(b)}, C$)
  }
\textbf{(3)} \For{$S \in \cup_{b=1}^{B} \mathcal{S}_{(b)}$}{
    $sta(S) = \frac{1}{B} \cdot \sum_{b=1}^B \mathds{1} \left[S \in \mathcal{S}_{(b)}\right]$
}
\KwOut{$\{ S, sta(S) \}_{S \in \cup_{b=1}^B \mathcal{S}_{(b)}}$}
\KwOut{$\{RF(w^{(K)}) \mbox{ on $\mathcal{D}$} \}$}
\caption{iterative Random Forests}
\label{algo:irf}
\end{algorithm}
\newpage

\section{Remarks on iRF}

\subsection{Iterative re-weighting}
Generalized RIT can be used with any Random Forest (RF) method, weighted or not. We find that iterative re-weighting acts as a soft dimension reduction step by encouraging RF to select a stable set of features on decision paths. This leads to improved recovery of high-order interactions in our numerical simulations and in real data settings. For instance, without feature re-weighting ($k = 1$) iRF rarely recovers interactions of order $> 2$ in our simulations. Feature re-weighting ($k > 1$) allows iRF to identify order-$8$ data generating rules as highly stable interactions for comparable parameter settings. In the enhancer case study, iRF ($k=5$) recovers $9$ order-$3$ interactions with stability score $> 0.5$. Without iterative re-weighting, iRF ($k=1$) does not recover any order-$3$ interactions with stability score $> 0.5$. The fourth iteration of iRF also recovers many additional order-3, order-4, and order-5 interactions with lower stability scores that are not recovered in the first iteration. Although it is unclear which of these high-order interactions represent true biological mechanisms without experimental follow-up, our simulation based on the enhancer data suggests that the overall quality of recovered interactions improves with iteration (Figure \ref{fig:sim3}).

Iterative re-weighting can be viewed as a form of regularization on the base RF
learner, since it restricts the form of functions RF is allowed to fit in a
probabilistic manner. In particular, we find that iterative re-weighting reduces the dimensionality of the feature space without removing marginally unimportant features that participate in high-order interactions (Figure \ref{fig:xor-depth}). Moreover, we find that iteratively re-weighted and unweighted RF achieve similar predictive accuracy on held out test data. We note that other forms of regularization such as \citep{deng2012feature} may also lead to improved interaction recovery, though we do not explore them in this paper.

\subsection{Generalized RIT}
The RIT algorithm could be generalized through any approach that selects active features from continuous or categorical data. However, the feature selection procedure affects recovered interactions and is thus an important consideration in generalizing RIT to continuous or categorical features. There are several reasons we use an RF-based approach. First, RFs are empirically successful predictive algorithms that provide a principled, data-driven procedure to select active features specific to each observation. Second, randomness inherent to tree ensembles offers a natural way to generate multiple active index sets for each observation $\mathbf{x}_i$, making the representations more robust to small data perturbations. Finally, our approach allows us to interpret (in a computationally efficient manner given by RIT) complex, high-order relationships that drive the impressive predictive accuracy of RFs, granting new insights into this widely used class of algorithms.

\subsection{Node sampling}
In the generalized RIT step of iRF, we represent each observation $i=1,\dots,n$ by $T$
rule-response pairs, determined by the leaf nodes containing observation $i$ in
each tree $t=1,\dots,T$ of an RF. We accomplish this by replicating each rule-response
pair $(\mathcal{I}_{j_t}, Z_{j_t})$ in tree $t$ based on the number of observations
in the corresponding leaf node. We view this as a natural representation of
the observations in $\mathcal{D}$, made more robust to sampling perturbations through
rules derived from bootstrap samples of $\mathcal{D}$. Our representation is equivalent to sampling rule-response pairs $(\mathcal{I}_{j_t}, Z_{j_t})$ in RIT with probability proportional to the number of observations that fall in the leaf node. However, one could sample or select a subset of leaf nodes based on other properties such as homogeneity and/or predictive accuracy. We are exploring how different sampling strategies impact recovered interactions in our ongoing work.


\subsection{Bagged stability scores}
iRF uses two layers of bootstrap sampling. The ``inner'' layer takes place when
growing weighted RF. By drawing a separate bootstrap sample from the input data
before growing each tree, we can learn multiple binary representations of each observation
$\mathbf{x}_i$ that are more robust to small data perturbations. The ``outer'' layer
of bootstrap sampling is used in the final iteration of iRF. Growing
$RF(w^{(K)})$ on different bootstrap samples allows us to assess the stability,
or uncertainty, associated with the recovered interactions.  

\subsection{Relation to AdaBoost}
In his original paper on RF \citep{breiman2001random}, Breiman conjectured that in the later stages of iteration, AdaBoost \citep{freund1995desicion} emulates RF. iRF inherits this property, and in addition shrinks the feature space towards more informative features. As pointed out by a reviewer, there is an interesting connection between AdaBoost and iRF. Namely, AdaBoost improves on the least reliable part of the data space, while iRF zooms in on the most reliable part of feature space. This is primarily motivated by the goals of the two learners --- AdaBoost's primary goal is prediction, whereas iRF's primary goal is to select features or combinations of features while retaining predictive power. We envision that zooming in on both the data and feature space simultaneously may harness the strengths of both learners. As mentioned in the conclusion, we are exploring this direction through local feature importance.

\subsection{Sensitivity to tuning parameters}
The predictive performance of RF is known to be highly robust to choice of tuning parameters \citep{breiman2001random}. To test iRF's sensitivity to tuning parameters, we investigated the stability of both prediction accuracy (AUC-PR) and interaction recovery across a range of parameter settings. Results are reported for both the enhancer and splicing datasets presented in our case studies.

The prediction accuracy of iRF is controlled through both RF parameters and number of iterations. Figures \ref{fig:enhancer-cv} and \ref{fig:splice-cv} report $5-$fold cross-validation prediction accuracy as a function of number of iterations ($k$), number of trees in the RF ensemble (\texttt{ntree}), and the number of variables considered for each split (\texttt{mtry}). We do not consider tree depth as a tuning parameter since deep decision trees (e.g. grown to purity) are precisely what allows iRF to identify high-order interactions. Aside from iteration $k=1$ in the splicing data, prediction accuracy is highly consistent across parameter choices. For the first iteration in the splicing data, prediction accuracy increases as a function of \texttt{mtry}. We hypothesize that this is the result of many extraneous features that make it less likely for important features to be among the \texttt{mtry} selected features at each split. Our hypothesis is consistent with the improvement in prediction accuracy that we observe for iterations $k>1$, where re-weighting allows iRF to sample important features with higher probability. This finding also suggests a potential relationship between iterative re-weighting and RF tuning parameters. The extent to which RF tuning parameters can be used to stabilize decision paths and allow for the recovery of high-order interactions is an interesting question for further exploration.

The interactions recovered by iRF are controlled through RIT parameters and the number of iterations. Our simulations in Sections \ref{simple-sims}-\ref{enhancer-sims} extensively examine the relationship between the number of iterations and recovered interactions. Figures \ref{fig:enhancer-cv-rit} and \ref{fig:splice-cv-rit} report the stability scores of recovered interactions in the enhancer and splicing data as a function of RIT parameters. In general, the stability scores of recovered interactions are highly correlated between different RIT parameter settings, indicating that our results are robust over the reported range of tuning parameters. The greatest differences in stability scores occur for low values of depth ($D$) and number of children ($n_{child}$). In particular, a subset of interactions that are highly stable for larger values of $n_{child}$ are less stable with $n_{child} = 1$. In contrast, a subset of interactions that are highly stable for $D=3$ are considered less stable for larger values of $D$. We note that the findings in our case studies are qualitatively unchanged as tuning parameters are varied. Interactions we identified as most stable under the default parameter choices remain the most stable under different parameter choices.

\subsection{Regression and multiclass classification}

We presented iRF in the binary classification setting, but our algorithm can be naturally extended to multiclass or continuous responses. The requirement that responses are binary is only used to select a subset of leaf nodes as input to generalized RIT. In particular, for a given class $C\in\{0,1\}$, iRF runs RIT over decision paths whose corresponding leaf node predictions are equal to $C$. In the multiclass setting, we select leaf nodes with predicted class or classes of interest as inputs to RIT. In the regression setting, we consider leaf nodes whose predictions fall within a range of interest as inputs to generalized RIT. This range could be determined in domain-specific manner or by grouping responses through clustering techniques.

\subsection{Grouped features and replicate assays}\label{grouping}

In many classification and regression problems with omics data,  one faces the
problem of drawing conclusion at an aggregated level of the features at hand. The
simplest example is the presence of multiple replicates of a single assay, 
when there is neither a standard protocol to
choose one assay over the other, nor a known strategy to aggregate the
assays after normalizing them individually. Similar situations arise when there
are multiple genes from a single pathway in the feature sets, and one is only
interested in learning interactions among the pathways and not the individual
genes. 

In linear regression based feature selection methods like Lasso, grouping
information among features is usually incorporated by devising suitable grouped
penalties, which requires solving new optimization problems. The invariance
property of RF to monotone transformations of features and the nature of
the intersection operation used by RIT provide iRF a simple and computationally efficient
workaround to this issue. In particular, one uses all the unnormalized assays in
the tree growing procedure, and collapses the grouped features or replicates
into a ``super feature'' before taking random intersections. iRF then provide
interaction information among these super features, which could be used to
achieve further dimension reduction of the interaction search space.

\subsection{Interaction evaluation through prediction}
We view the task of identifying candidate, high-order interactions as a step towards hypothesis generation in complex systems. An important next step will be evaluationg the interactions recovered by iRF to determine whether they represent domain-relevant hypotheses. This is an interesting and challenging problem that will require subject matter knowledge into the anticipated forms of interactions. For instance, biomolecules are believed to interact in stereospecific groups \citep{nelson2008lehninger} that can be represented through Boolean-type rules. Thus, tests of non-additivity may provide insight into which iRF-recovered interactions warrant further examination in biological systems. 

We do not consider domain-specific evaluation in this paper, but instead assess interactions through broadly applicable metrics based on both stability and predictability. We incorporated the Stability Principle \citep{yu2013stability} through both iterative re-weighting, which encourages iRF to use a consistent set of features along decision paths, and through bagged stability scores, which provide a metric to evaluate how consistently decision rules are used throughout an RF. Here we propose two additional validation metrics based on predictive accuracy. 

\textbf{Conditional prediction:} Our first metric evaluates a recovered interaction $S\subseteq\{1,\dots,p\}$ based on the predictive accuracy of an RF that makes predictions using only leaf nodes for which all features in $S$ fall on the decision path. Specifically, for each observation $i=1,\dots,n$ we evaluate its predicted value from each tree $t=1,\dots T$ with respect to an interaction $S$ as

\begin{eqnarray*}
\hat{y}_{i}(t;S) = 
\begin{cases}
Z_{i_t} &\text{if } S \subseteq \mathcal{I}_{i_t}\\
\mathbb{P}_n (y=1) &\text{else}
\end{cases}
\end{eqnarray*}
where $Z_{i_t}$ is the prediction of the leaf node containing observation $i$ in tree $t$, $\mathcal{I}_{i_t}$ is the index set of features falling on the decision path for this leaf node, and $\mathbb{P}_n (y=1)$ is the empirical proportion of class $1$ observations $\{i:y_i =1\}$. We average these predictions across the tree ensemble to obtain the RF-level prediction for observation $i$ with respect to an interaction $S$

\begin{equation}
\hat{y}_{i}(S) = \frac{1}{T} \cdot \sum_{t=1}^T \hat{y}_{i}(t;S).
\label{int-pred}
\end{equation}
Predictions from equation \eqref{int-pred} can be used to evaluate predictive accuracy using any metric of interest. We report AUC-PR using predictions $\hat{y}_{i}(S)$ for each interaction $S\in \mathcal{S}$ recovered by iRF. Intuitively, this metric asks whether the leaf nodes that rely on an interaction $S$ are good predictors when all other leaf nodes make a best-case random guess.

\textbf{Permutation importance:} Our second metric is inspired by Breiman's permutation-based measure of feature importance \citep{breiman2001random}. In the single feature case, Breiman proposed permuting each column of the data matrix individually and evaluating the change in prediction accuracy of an RF. The intuition behind this measure of importance is that if an RF's predictions are heavily influenced by a particular feature, permuting it will lead to a drop in predictive accuracy by destroying the feature/response relationship. The direct analogue in our setting would be to permute all features in a recovered interaction $S$ and evaluate the change in predictive accuracy of iRF. However, this does not capture the notion that we expect features in an interaction to act collectively. By permuting a single feature, we destroy the interaction/response relationship for any interaction that the feature takes part in. If $S$ contains features that are components of distinct interactions, permuting each feature in $S$ would destroy multiple interaction/response relationships. To avoid this issue, we assess prediction accuracy using only information from the features contained in $S$ by permuting all other features.  

Specifically, let $X_{\pi_{S^c}}$ denote the feature matrix with all columns in $S^c$ permuted, where $S^c$ is the compliment of $S$. We evaluate predictions on permuted data $X_{\pi_{S^c}}$, and use these predictions to assess accuracy with respect to a metric of interest, such as the AUC-PR.  Intuitively, this metric captures the idea that if an interaction is important \textit{independently of any other features}, making predictions using only this interaction should lead to improved prediction over random guessing.

\textbf{Evaluating enhancer and splicing interactions:} Figures \ref{fig:importance-metrics-enhancer} and \ref{fig:importance-metrics-splicing} report interactions from both the enhancer and splicing data, evaluated in terms of our predictive metrics. In the enhancer data, interactions between collections of TFs Zld, Gt, Hb, Kr, and Twi are ranked highly, as was the case with stability scores (Figure \ref{fig:importance-metrics-enhancer}). In the splicing data, POL II, S2 phospho-Pol II, H3K36me3, H3K79me2, H3K9me1, and H4K20me1 consistently appear in highly ranked interactions, providing further validation of the order-6 interaction recovered using the stability score metric (Figure \ref{fig:importance-metrics-splicing}).

While the interaction evaluation metrics yield qualitatively similar results, there is a clear difference in how they rank interactions of different orders. Conditional prediction and stability score tend to favor lower-order interactions and permutation importance higher-order interactions. To see why this is the case, consider interactions $S'\subset S \subseteq \{1,\dots,p\}$. As a result of the intersection operation used by RIT, the probability (with respect to the randomness introduced by RIT) that the larger interaction $S$ survives up to depth $D$ will be less than or equal to the probability that $S'$ survives up to depth $D$. Stability scores will reflect the difference by measuring how frequently an intersection survives across bootstrap samples. In the case of conditional prediction, the leaf nodes for which $S$ falls on the decision path form a subset of leaf nodes for which $S'$ falls on the decision path. As a result, the conditional prediction with respect to $S$ uses more information from the forest and thus we would generally expect to see superior predictive accuracy. In contrast, permutation importance uses more information when making predictions with $S$ since fewer variables are permuted. Therefore, we would generally expect to see higher permutation importance scores for larger interactions. We are currently investigating approaches for normalizing these metrics to compare interactions of different orders.

Together with the measure of stability, the two importance measures proposed here capture different qualitative aspects of an interaction. Conceptually, the stability measure attempts to capture the degree of uncertainty associated with an interaction by perturbing the features and responses jointly. In contrast, the importance measures based on conditional prediction and permutation are similar to effect size, i.e., they attempt to quantify the contribution of a given interaction to the overall predictive accuracy of the learner. The conditional prediction metric accomplishes this by perturbing the predicted responses, while permutation importance perturbs the features.

%% file: sections-pnas/supp-processing.tex
\section{Data processing}
\subsection{\textit{Drosophila} enhancers} \label{enhancer-dataset}

In total, $7809$ genomic sequences have been evaluated for their enhancer activity  \citep{berman2002exploiting, fisher2012dna, pmid20087342, pmid24896182} in a
gold-standard, stable-integration transgenic assay. In this setting, a short
genomic sequence (100-3000nt) is placed in a reporter construct and integrated
into a targeted site in the genome. The transgenic fly line 
is amplified, embryos are collected, fixed, hybridized and immunohistochemistry is performed to 
detect the reporter \citep{tautz1989non, weiszmann2009determination}. The resultant stained embryos are  imaged to determine: 
a) whether or not the genomic segment is sufficient to drive transcription of the reporter construct, and 
b) where and when in the embryo expression is driven. 
For our prediction problem, sequences that drive patterned expression in blastoderm (stage 5) embryos were labeled as 
active elements. To form a set of features for predicting enhancer status, we computed the maximum 
value of normalized fold-enrichment \citep{li2008transcription} of ChIP-seq and ChIP-chip assays 
\citep{macarthur2009developmental, encode2012integrated}
for each genomic segment. The processed data are provided in Supporting Data 1.

Our processing led to a binary classification problem with approximately 
10\% of genomic sequences labeled as active elements. It is important to note that
the tested sequences do not represent a random sample from the genome --- rather
they were chosen based on prior biological knowledge and may therefore exhibit a
higher frequency of positive tests than one would expect from genomic sequences
in general. We randomly divided the dataset into training and test sets of $3912$ and $3897$ observations respectively, with approximately equal portions of
positive and negative elements, and applied iRF with $B = 30$, $K = 5$. The
tuning parameters in RF were set to default levels of the \texttt{R}
\texttt{randomForest} package, and 500 Random Intersection Trees of depth 5 with
$n_{child}=2$ were grown to capture candidate interactions.

\subsection{Alternative splicing}

The ENCODE consortium has collected extensive genome-wide data on both chromatin
state and splicing in the human-derived erythroleukemia cell line K562
\citep{encode2012integrated}. To identify critical
interactions that form the basis of chromatin mediated splicing, we used
splicing rates (Percent-spliced-in, PSI values, \citep{pervouchine2012intron, pervouchine2016ipsa})
from ENCODE RNA-seq data, along with ChIP-seq assays measuring enrichment of
chromatin marks and transcription factor binding events (253 ChIP assays on 107
unique transcription factors and 11 histone modifications,
\href{https://www.encodeproject.org/}{https://www.encodeproject.org/}). A complete description of the assays, including accession numbers, is provided in Supporting Data 2.

For each ChIP assay, we computed the maximum value of normalized fold-enrichment over the genomic region corresponding to each exon. This yielded a set of $p = 270$ features for our
analysis. We took our response to be a thresholded function of the PSI values
for each exon. Only internal exons with high read count (at least 100 RPKM) were
used in downstream analysis. Exons with Percent-spliced-in index (PSI) above
70\% were classified as frequently included ($y = 1$) and exons with PSI below
30\% were classified as frequently excluded exons ($y = 0$). This led to a total
of $23823$ exons used in our analysis. The processed data are provided in Supporting Data 3.

Our threshold choice resulted in $\sim 90\%$ of observations belonging to class $1$. To account for this imbalance, we report AUC-PR for the class $0$ observations. 
We randomly divided the dataset into balanced training and test sets of 11911 and 11912 observations respectively, and applied iRF with $B = 30$ and $K = 2$. The tuning parameters in RF were set to default levels of the \texttt{R}
\texttt{randomForest} package, and 500 binary random intersection trees of
depth 5 with $n_{child}=2$ were grown to capture candidate interactions.  

\section{Evaluating \textit{Drosophila} enhancer interactions}
The Drosophila embryo is one of the most well studied systems in developmental biology and provides a valuable test case for evaluating iRF. Decades of prior work have identified physical, pairwise TF interactions that play a critical role in regulating spatial and temporal patterning, for reviews see \citet{rivera1996gradients} and \citet{Jaeger2011}.  We compared our results against these previously reported physical interactions to evaluate interactions found by iRF.  Table \ref{pairwise} indicates the $20$ pairwise TF interactions we identify with stability score $> 0.5$, along with references that have previously reported physical interactions among each TF pair. In total, $16$ ($80\%$) of the $20$ pairwise TF interactions we identify as stable have been previously reported in one of two forms: (i) one member of the pair regulates expression of the other (ii) joint binding of the TF pair has been associated with increased expression levels of other target genes. Interactions for which we could not find evidence supporting one of these forms are indicated as $``-"$ in Table \ref{pairwise}. We note that high-order interactions have only been studied in a small number of select cases, most notably $eve$ stripe 2, for a review see \citep{levine2013computing}. These limited cases are not sufficient to conduct a comprehensive analysis of the high-order interactions we identify using iRF.

\begin{table}[H]
\caption{Previously identified pairwise TF interactions recovered by iRF with stability score $> 0.5$}\label{pairwise}

\begin{tabular}{ccp{0.75\textwidth}}
\hline
interaction $(S)$&	$sta(S)$ &	\multicolumn{1}{c}{references}\\
\hline
\hline
\vspace{0.5ex}
Gt, Zld	&1	&\cite{harrison2011zelda, nien2011temporal} \\
\vspace{0.5ex}
Twi, Zld	&1	&\cite{harrison2011zelda, nien2011temporal}\\
\vspace{0.5ex}
Gt, Hb	&1	&\cite{kraut1991mutually, kraut1991spatial, eldon1991interactions}\\
\vspace{0.5ex}
Gt, Kr	&1	&\cite{kraut1991spatial, struhl1992control, capovilla1992giant, schulz1994autonomous}\\
\vspace{0.5ex}
Gt, Twi	&1	&\cite{li2008transcription}\\
\vspace{0.5ex}
Kr, Twi	&1	&\cite{li2008transcription}\\
\vspace{0.5ex}
Kr, Zld	&0.97	&\cite{harrison2011zelda, nien2011temporal}\\
\vspace{0.5ex}
Gt, Med	&0.97	& $-$\\
\vspace{0.5ex}
Bcd, Gt	&0.93	&\cite{kraut1991spatial, eldon1991interactions}\\
\vspace{0.5ex}
Bcd, Twi	&0.93	&\cite{li2008transcription}\\
\vspace{0.5ex}
Hb, Twi	&0.93	&\cite{zeitlinger2007whole}\\
\vspace{0.5ex}
Med, Twi	&0.93	&\cite{nguyen1998drosophila}\\
\vspace{0.5ex}
Kr, Med	&0.9	& $-$\\
\vspace{0.5ex}
D, Gt 	&0.87& $-$	\\
\vspace{0.5ex}
Med, Zld	&0.83&	\cite{harrison2011zelda}\\
\vspace{0.5ex}
Hb, Zld	&0.80	&\cite{harrison2011zelda, nien2011temporal}\\
\vspace{0.5ex}
Hb, Kr	&0.80	&\cite{nusslein1980mutations, jackle1986cross, hoch1991gene}\\
\vspace{0.5ex}
D, Twi	&0.73&	$-$\\
\vspace{0.5ex}
Bcd, Kr	&0.67&	\cite{hoch1991gene, hoch1990cis}\\
\vspace{0.5ex}
Bcd, Zld	&0.63&	\cite{harrison2011zelda, nien2011temporal}\\
\hline
\end{tabular}
\end{table}

%% file: sections-pnas/supp-simulation_impsample.tex
\section{Simulation experiments}\label{sims}
We developed iRF through extensive simulation studies based on biologically
inspired generative models using both synthetic and real data. In particular, we
generated responses using Boolean rules intended to reflect the stereospecific
nature of interactions among biomolecules \citep{nelson2008lehninger}. In this
section, we examine interaction recovery and predictive accuracy of iRF in a
variety of simulation settings. 

For all simulations in Sections \ref{simple-sims}-\ref{bigp-sims}, we evaluated predictive accuracy in terms of area under the precision-recall curve (AUC-PR) for a held out test set of $500$ observations. To evaluate interaction recovery, we use three metrics that are intended to give a broad sense of the overall quality of interactions $\mathcal{S}$ recovered by iRF. For responses generated from an interaction $S^*\subseteq \{1,\dots,p\}$, we consider interactions of any order between only active features $\{j:j\in S^*\}$ to be true positives and interactions containing any non-active variable $\{j:j\notin S^*\}$ to be false positives. This definition accounts for the fact that subsets of $S^*$ are still informative of the data generating mechanism. However, it conservatively considers interactions that includes any non-active features to be false positives, regardless of how many active features they contain.

\begin{enumerate}
  \item \textbf{Interaction AUC:} We consider the area under the receiver operating characteristic (ROC) curve generated by
    thresholding interactions recovered by iRF at each unique stability score. This metric provides a
    rank-based measurement of the overall quality of iRF interaction stability scores, and takes a value 	 of $1$ whenever the complete data generating mechanism is recovered as the most stable interaction.
 \item \textbf{Recovery rate:} We define an interaction as ``recovered'' if it is returned in any of the $B$ bootstrap samples (i.e. stability score $>$ 0), or if it is a subset of any recovered interaction. This eliminates the need to select thresholds across a wide variety of parameter settings. For a given interaction order $s=2, \ldots, |S|$, we calculate the proportion of the total $|S| \choose s$ true positive order-$s$ interactions recovered by iRF. This metric is used to distinguish between models that recover high-order interactions at different frequencies, particularly in settings where all models recover low-order interactions.
  \item \textbf{False positive weight:} Let $\mathcal{S} = \mathcal{S}_{T} \cup
    \mathcal{S}_{F}$ represent the set of interactions recovered by iRF, where
    $\mathcal{S}_{T}$ and $\mathcal{S}_{F}$ are the sets of recovered true and false
    positive interactions respectively. For a given interaction order 
    $s=2,\ldots, |S|$, we calculate
    \begin{equation*}
      \frac{\sum_{S\in\mathcal{S}_{F}:|S|=s}sta(S)}
      {\sum_{S\in\mathcal{S}:|S|=s} sta(S)}.
    \end{equation*}
This metric measures the aggregate weight of stability scores for false positive order-$s$ interactions, $S\in\mathcal{S}_{F}:|S|=s$, relative to all recovered order-$s$ interactions, $S\in\mathcal{S}:|S|=s$. This metric also includes all recovered interactions (stability score $>$ 0), eliminating the need to select thresholds. It can be thought of as the weighted analogue to false discovery proportion.
\end{enumerate}
    
\subsection{Simulation 1: Boolean rules} \label{simple-sims}

Our first set of simulations demonstrates the benefit of iterative re-weighting
for a variety of Boolean-type rules. We sampled features $\mathbf{x}=(x_{1},
\ldots, x_{50})$ from independent, standard Cauchy distributions to reflect heavy-tailed data, and generated the binary responses from three rule settings (OR, AND, and XOR) as 

\begin{eqnarray}
  y ^ {(OR)}&=& \mathds{1} \left[x_1 > t_{OR} \,|\, x_2 > t_{OR} \,|\, x_3 > t_{OR} \,|\, x_4 > t_{OR} \right],
\label{eq:or}\\
y ^{(AND)} &=& \prod_{i=1}^4 \mathds{1} \left[ x_i > t_{AND}\right],
\label{eq:and}\\
y ^{(XOR)} &=& \mathds{1} \left[\sum_{i=1} ^ 4 \mathds{1}(x_i > t_{XOR}) \equiv 1 \pmod{2} \right].
\label{eq:xor}
\end{eqnarray}
We injected noise into these responses by swapping the labels for $20\%$ of the observations selected at random. From a modeling perspective, the rules in equations \eqref{eq:or}, \eqref{eq:and}, and \eqref{eq:xor} give rise to non-additive main effects that can be represented as an order-$4$ interaction between the active features $x_1, x_2, x_3,$ and  $x_4$. Inactive features $x_{5}, \ldots, x_{50}$ provide an additional form of noise that allowed us to assess the performance of iRF in the presence of extraneous features. For the AND and OR models, we set $t_{OR} = 3.2$, $t_{AND} = -1 $ to ensure reasonable class balance ($\sim 1/3$ class $1$ observations) and trained on samples of size $100, 200, \ldots,500$ observations. We set $t_{XOR} = 1$ both for class balance ($\sim 1/2$ class $1$ observations) and to ensure that some active features were marginally important relative to inactive features. At this threshold, the XOR interaction is more difficult to recover than the others due to the weaker marginal associations between active features and the response. To evaluate the full range of performance for the XOR model, we trained on larger samples of size $200, 400, \ldots,1000$ observations.  We report the prediction accuracy and interaction recovery for iterations $k\in\{1,2, \ldots, 5\}$ of iRF over 20 replicates drawn from the above generative models. The RF tuning parameters were set to default levels for the \texttt{R} \texttt{randomForest} package \citep{liaw2002classification}, $M = 100$ RITs of depth $5$ were grown with $n_{child}=2$, and $B = 20$ bootstrap replicates were taken to determine the stability scores of recovered interactions.

Figure \ref{fig:simple}A shows the prediction accuracy of iRF (AUC-PR), evaluated on held out test data, for each generative model and a selected subset of training sample sizes as a function of iteration number ($k$). iRF achieves comparable or better predictive performance for increasing $k$, with the most dramatic improvement in the XOR model. It is important to note that only $4$ out of the $50$ features are used to generate responses in equations \eqref{eq:or}, \eqref{eq:and}, and \eqref{eq:xor}. Iterative re-weighting restricts the form of functions fitted by RF and may hurt predictive performance when the generative model is not sparse.

Figure \ref{fig:simple}B shows interaction AUC by generative model, iteration number, and training sample size, demonstrating that iRF ($k>1$) tends to rank true interactions higher with respect to stability score than RF ($k=1$). Figure \ref{fig:simple}C breaks down recovery by interaction order, showing the proportion of order-$s$
interactions recovered across any bootstrap sample (stability score $> 0$),
averaged over 20 replicates. For each of the generative models, RF ($k=1$) never recovers
the true order-$4$ interaction while iRF $(k=4,5)$ always identifies it as the most
stable order-$4$ interaction given enough training observations. The improvement in interaction recovery with iteration is accompanied by an increase in the stability scores of false positive interactions (Figure \ref{fig:simple}D). We find that this increase is generally due to many false interactions with low stability scores as opposed to few false interactions with high stability scores. As a result, true positives can be easily distinguished through stability score ranking (Figure \ref{fig:simple}B).

These findings support the idea that iterative re-weighting allows iRF to recover high-order interactions without limiting predictive performance. In particular, improved interaction recovery with iteration indicates that iterative re-weighting stabilizes decision paths, leading to more interpretable models. We note that a principled approach for selecting the total number of iterations $K$ can be formulated in terms of estimation stability with cross validation (ESCV) \citep{lim2015estimation}, which would balance trade-offs between interpretability and predictive accuracy.

\subsection{Simulation 2: marginal importance}

Section \ref{simple-sims} demonstrates that iterative re-weighting
improves the recovery of high-order interactions. The following
simulations develop an intuition for how iRF constructs high-order
interactions, and under what conditions the algorithm fails. In particular, the simulations demonstrate that iterative re-weighting allows iRF to select marginally important
active features earlier on decision paths. This leads to more favorable
partitions of the feature space, where active features that are marginally less
important are more likely to be selected. 

We sampled features $\textbf{x}=(x_1, \ldots, x_{100})$ from independent,
standard Cauchy distributions, and generated the binary response $y$ as

\begin{equation}
 y = \mathds{1} 
\left[ \sum_{i\in S_{XOR}}  \mathds{1}(x_i > t_{XOR}) \equiv 1 \pmod{2} \right],
\label{xor-8}
\end{equation}
$S_{XOR}=\{1,\ldots,8\}$. We set $t_{XOR} = 2$, which resulted in a mix of marginally important and unimportant active features, allowing us to study how iRF constructs interactions. For all simulations described in this section, we generated $n = 5000$ training observations and evaluated the fitted model on a test set of $500$ held out observations. RF parameters were set to their default values with the exception of \texttt{ntree}, which was set to
$200$ for computational purposes. We ran iRF for $k\in\{1,\ldots, 5\}$
iterations with $10$ bootstrap samples and grew $M = 100$ RITs of depth $5$ with $n_{child}=2$. Each simulation was replicated $10$
times to evaluate performance stability.

\subsubsection{Noise level}\label{xor-noise}
In the first simulation, we considered the effect of noise on interaction
recovery to assess the underlying difficulty of the problem. We generated responses
using equation \eqref{xor-8}, and swapped labels for $10\%, 15\%,$ and $20\%$ of randomly selected responses. 

Figure \ref{fig:xorNoise} shows performance in terms of predictive accuracy and
interaction recovery for the $15\%$ and $20\%$ noise levels. At each noise level, increasing $k$ leads to superior performance, though there is a substantial drop in both absolute performance and the rate of improvement over iteration for increased noise levels. 

The dramatic improvement in interaction recovery (Figure \ref{fig:xorNoise}C) 
reinforces the idea that regularization is critical for recovering high-order interactions. Figure \ref{fig:xor-weights} shows the distribution of iRF weights, which reflect the degree of regularization, by iteration. iRF successfully recovers the full XOR interaction in settings where there is clear separation between the distribution of active and inactive variable weights. This separation develops over several iterations, and at a noticeably slower rate for higher noise levels, indicating that further iteration may be necessary in low signal-noise regimes.

\textbf{Marginal importance and variable selection}: iRF's improvement with
iteration suggests that the algorithm leverages informative lower-order
interactions to construct the full data generating rule through adaptive
regularization. That is, by re-weighting towards some active features, iRF are
more likely to produce partitions of the feature space where remaining active
variables are selected. To investigate this idea further, we examined the
relationship between marginal importance and the average depth at which features
are first selected across the forest. We define a variable's marginal
importance as the best case decrease in Gini impurity if it were selected as the
first splitting feature. We note that this definition is different from the
standard measure of RF importance (mean decrease in Gini
impurity), which captures an aggregate measurement of
marginal and conditional importance over an RF. We considered this particular
definition to examine whether iterative re-weighting leads to more ``favorable''
partitions of the feature space, where marginally unimportant features are
selected earlier on decision paths.

Figure \ref{fig:xor-depth} shows the relationship between marginal importance
and feature entry depth. On average over the tree ensemble, active features enter the model earlier with further iteration, particularly in settings where iRF successfully recovers the full XOR interaction. We note that this occurs for active features with both high and low marginal importance,
though more marginally important, active features enter the model earliest. This
behavior supports the idea that iRF constructs high-order interactions by
identifying a core set of active features, and using these, partitions the
feature space in a way that marginally less important variables become
conditionally important, and thus more likely to be selected.

\subsubsection{Mixture model}
Our finding that iRF uses iterative re-weighting to build up interactions around marginally important features, suggests that the algorithm may struggle to recover interactions
in the presence of other marginally important features. To test this
idea, we considered a mixture model of XOR and AND rules. A proportion $\pi \in \{0.5, 0.75, 0.9\}$ of randomly selected observations were generated
using equation \eqref{xor-8}, and the remaining proportion $1-\pi$ of observations were generated as

\begin{equation}
y = \prod_{i\in S_{AND}} \mathds{1} \left[ x_i > t_{AND} \right].
\label{eq:andmix}
\end{equation}
We introduced noise by swapping labels for $10\%$ of
the responses selected at random, a setting where iRF easily recovers the full XOR rule, and set $S_{AND} = \{9, 10, 11, 12\}$, $t_{AND} = -0.5$  to ensure that the XOR and AND interactions were dominant with respect to marginal importance for $\pi=0.9$ and $\pi=0.5$ respectively.

Figure \ref{fig:mixxor-pred-rec} shows performance in terms of predictive
accuracy (A) and interaction recovery of XOR (B) and AND (C) rules
at each level of $\pi$. When one rule is clearly dominant (AND: $\pi=0.5$; XOR:
$\pi=0.9$), iRF fail to recover the the other (Figure \ref{fig:mixxor-pred-rec}
B,C). This is driven by the fact that the algorithm
iteratively updates feature weights using a global measure of importance,
without distinguishing between features that are more important for certain
observations and/or in specific regions of the feature space. One could address
this with local measures of feature importance, though we do not explore the
idea in this paper. 

In the $\pi=0.75$ setting, none of the interactions are clearly more important, and iRF recovers
subsets of both the XOR and AND interactions (Figure \ref{fig:mixxor-pred-rec}).
While iRF may recover a larger proportion of each rule with further iteration,
we note that the algorithm does not explicitly distinguish between rule types, and would
do so only when different decision paths in an RF learn distinct rules.
Characterizing the specific form of interactions recovered by iRF is an
interesting question that we are exploring in our ongoing work.

\subsubsection{Correlated features}
In our next set of simulations, we examined the effect of correlated features
on interaction recovery. Responses were generated using equation
(\ref{xor-8}), with features $\textbf{x}=(x_1, \ldots, x_{100})$ drawn from a Cauchy
distribution with mean $0$ and covariance $\Sigma$, and active set $S_{XOR}$, $|S_{XOR}|=8$ sampled uniformly at random from $\{1,\dots,100\}$. We considered both a decaying covariance structure: $\Sigma_{ij} = \rho ^ {|i - j|}$, and a block covariance structure:

\begin{equation*}
\Sigma_{ij} = 
\begin{cases}
1, & i=j \\
\rho, & i,j \subset G_{l} \text{ and } i\ne j\\
0, & \text{else}
\end{cases}
\end{equation*}
where $G_{l}\subseteq\{1,\dots,p\}$ and $l=1, \dots, L$ partition $\{1,\dots,p\}$ into blocks of features. For the following simulations, we considered both low and high levels of feature correlation $\rho\in\{0.25, 0.75\}$ and blocks of $10$ features.

Prediction accuracy and interaction recovery are fairly consistent for moderate
values of $\rho$ (Figures \ref{fig:xortoeplitz}, \ref{fig:xorblock}), while interaction recovery degrades for larger values of $\rho$, particularly in the block covariance
setting (Figure \ref{fig:xorblock}B,C). For instance when
$\rho=0.75$, iRF only recovers the full order-8 interaction at $k=5$, and
simultaneously recovers many more false positive interactions. The drop in
interaction recovery rate is greater for larger interactions due to the
fact that for increasing $\rho$, inactive features are more frequently selected in place of active features. These findings
suggest both that iRF can recover meaningful interactions in highly correlated
data, but that these interactions may also contain an increasing proportion of
false positive features.

We note that the problem of distinguishing between many highly correlated
features, as in the $\rho=0.75$ block covariance setting, is difficult for any
feature selection method. With $a$ $priori$ knowledge about the relationship between
variables, such as whether variables represent replicate assays or components of
the same pathway, one could group features as described in Section \ref{grouping}.

\subsection{Simulation 3: big $p$}\label{bigp-sims}

Our final set of synthetic data simulations tested the performance of iRF in settings where the number of features is large relative to the number of observations. Specifically, we drew 500 independent, $p-$dimensional standard Cauchy features, with $p\in\{1000, 2500\}$. Responses were generated using the order-$4$ AND interaction from equation $\eqref{eq:and}$, selected to reflect the form of interactions recovered in the splicing and enhancer case studies. We injected noise into the responses by swapping labels for $20\%$ and $30\%$ of randomly selected observations.

Figures \ref{fig:p0.2} and \ref{fig:p0.3} show prediction accuracy and interaction recovery of iRF at each of the different noise levels. Prediction accuracy improves noticeably with iteration and stabilizes at the $20\%$ noise level (Figures \ref{fig:p0.2}A, \ref{fig:p0.3}A). For $k=1$, iRF rarely recovers correct interactions and never recovers interactions of order $> 2$, while later iterations recover many true interactions (Figures \ref{fig:p0.2}C, \ref{fig:p0.3}C). These findings indicate that iterative re-weighting is particularly important in this highly sparse setting and is effectively regularizing RF fitting. Based on the results from our previous simulations, we note that the effectiveness of iterative re-weighting will be related to the form of interactions. In particular, iRF should perform worse in settings where $p>>n$ and interactions have no marginally important features.

\subsection{Simulation 4: enhancer data}\label{enhancer-sims}
To test iRF's ability to recover interactions in real data, we incorporated biologically inspired Boolean rules into the \emph{Drosophila} enhancer dataset analyzed in Section 4 (see also  Section \ref{enhancer-dataset} for a description of the dataset). These simulations were motivated by our desire to assess iRF's ability to recover signals embedded in a noisy, non-smooth and realistic response surface with feature correlation and class imbalance comparable to our case studies. Specifically, we used all TF binding features from the enhancer data and embedded a 5-dimensional AND rule between Kr{\"u}ppel, (Kr), Hunchback (Hb), Dichaete (D), Twist (Twi), and Zelda (Zld):

\begin{equation}
 y  = \mathds{1} [x_{kr} > 1.25 \,\&\, x_{hb} > 1.25 \,\&\, x_{D} > 1.25 \,\&\, x_{twi} > 1.25 \,\&\, x_{zld} > 75].
 \label{eq:enhancer-gen}
\end{equation}
The active TFs and thresholds were selected to ensure that the proportion of positive responses was comparable to the true data ($\sim 10\%$ active elements), and the interaction type was selected to match the form of interactions recovered in both the enhancer and splicing data. 

In this set of simulations, we considered two types of noise. For the first, we
incorporated noise by swapping labels for a randomly selected subset of $20\%$
of active elements and an equivalent number of inactive elements. We note that this resulted in a fairly limited proportion of swapped labels among class $0$ observations due to class imbalance. Our second noise setting was based on an RF/sample splitting procedure. Specifically, we divided the data into two disjoint groups of equal size. For each group, we trained an RF and used it to predict the responses of observations in the held out group. This process resulted in
predicted class probabilities for each observation $i=1,\dots,n$. We repeated this procedure
$20$ times to obtain the average predicted probability that $y_i=1$. With a slight abuse of notation, we denote this predicted probability as $\pi_i$. For each observation we sampled a Bernoulli noising variable $\tilde{y}_i \sim Bernoulli(\pi_i)$ and used these to generate a binary response for each observation

\begin{equation*}
y_i  = \tilde{y}_i \, | \, \mathds{1} [x_{kr} > 1.25 \,\&\, x_{hb} > 1.25 \,\&\, x_{D} > 1.25 \,\&\, x_{twi} > 1.25 \,\&\, x_{zld} > 75]. 
\end{equation*}
That is, the response for observation $i$ was to set $1$ whenever the noising variable $\tilde{y}_i$ or equation \eqref{eq:enhancer-gen} was active. This noising procedure introduced an additional $\sim 5\%$ of class 1 observations beyond the $\sim 10 \%$ of observations that were class 1 as a result of equation \eqref{eq:enhancer-gen}. Intuitively, this model derives its noise from rules learned by an RF. Feature interactions that are useful for classifying observations in the split data are built into the predicted class probabilities $\pi_i$. This results in an underlying noise model that is heterogeneous, composed of many ``bumps'' throughout the feature space.

In each setting, we trained on samples of $200, 400, \ldots, 2000$ observations and tested prediction performance on the same number of observations used to train. We repeated this process 20 times to assess variability in interaction recovery and prediction accuracy. The RF tuning parameters were set to default levels for the \texttt{R} \texttt{randomForest} package, $M = 100$ random intersection trees of depth $5$ were grown with $n_{child}=2$, and $B = 20$ bootstrap replicates were taken to determine the stability scores of recovered interactions.

Figure \ref{fig:sim3}A shows that different iterations of iRF achieve comparable predictive accuracy in both noise settings. When the number of training observations increases beyond $400$, the overall quality of recovered interactions as measured by interaction AUC improves for iterations $k>1$. In some instances, there is a drop in the quality of recovered interactions for the largest values of $k$ after the initial jump at $k=2$ (Figure \ref{fig:sim3}).  All iterations frequently recover true order-$2$ interactions, though the weighted false positive rate for order-$2$ interactions drops for iterations $k > 1$, suggesting that iterative re-weighting helps iRF filter out false positives. Iterations $k > 1$ of iRF recover true high-order interactions at much greater frequency for a fixed sample size, although these iterations also recover many false high-order interactions (Figure \ref{fig:sim3}C,D). We note that true positive interactions are consistently identified as more stable (Figure \ref{fig:enhancerStab}), suggesting that the large proportion of weighted false discoveries in Figure \ref{fig:sim3}D is the result of many false positives with low stability scores.

%% file: sections-pnas/supp-runtime.tex
\section{Computational cost of detecting high-order interaction}\label{runtime}
We used the enhancer data from our case studies to demonstrate the computational advantage of iRF for detecting high-order interactions in high-dimensional data. Rulefit3 serves as a benchmark, which has competitive prediction accuracy to RF and also comes with a flexible framework for detecting nonlinear interactions hierarchically, using the so-called ``H-statistic'' \citep{friedman2008predictive}. For moderate to large dimensional datasets typically encountered in omics studies, the computational complexity of seeking high-order interactions hierarchically (select marginally important features first, then look for pairwise interaction among them, and so on) increases rapidly, while the computation time of iRF grows far more slowly with dimension.

We fit iRF and Rulefit3 on balanced training samples from the enhancer dataset ($7809$ samples, $80$ features) using subsets of $p$ randomly selected features, where $p \in \{10, 20, \ldots, 80\}$. We ran Rulefit3 with default parameters, generating null interaction models with $10$ bootstrap samples and looked for higher order interactions among features whose H-statistics are at least one null standard deviation above their null average (following \citep{friedman2008predictive}). The current implementation of Rulefit3 only allows H-statistic calculation for interactions of up to order $3$, so we do not assess higher order interactions. We ran iRF with $B=10$ bootstrap samples, $K=3$ iterations, and the default RF and RIT tuning parameters. The run time (in minutes) and the AUC for different values of $p$, averaged over $10$ replications of the experiment by randomly permuting the original features in enhancer data, are reported in Figure \ref{fig:pic/runtime-taly-splicing}.

The plot on the left panel shows that the runtime for Rulefit3's interaction detection increases exponentially  as $p$ increases, while the increase is linear for iRF.  The search space of Rulefit3 is restricted to all possible interactions of order $3$, while iRF searches for arbitrarily high-order interactions, leveraging deep decision trees in RF. The linear vs. polynomial growth of computing time is not an optimization issue, it is merely a consequence of the exponentially growing search space of high-order interactions.

In addition to the comparison with Rulefit3, we profiled memory usage of the iRF \texttt{R} package using the splicing dataset described in Section 5 ($n=11911$, $p=270$) with  $B=30$ and $K=3$. The program was run on a server using 24 cores (CPU Model: Intel(R) Xeon(R) CPU E5-2697 v2 @ 2.70GHz, clock speed: 1200 MHz, Operating System: Ubuntu 14.04). The profiling was done using \texttt{R} functions \texttt{Rprof} and \texttt{summaryRprof}. iRF completed in 26 minutes 59 seconds, with a 499910 Mb memory consumption.

%% file: sections-pnas/supp-data.tex
\newpage
\section{List of datasets}
Scripts and data used for the case studies and simulations described in this paper are available on \href{https://doi.org/10.5281/zenodo.885529}{Zenodo}.

\subsubsection*{Scripts}
\begin{enumerate}
\item{\texttt{enhancer.R}}: \texttt{R} script used to run \texttt{iRF} on the enhancer data.
\item{\texttt{splicing.R}}: \texttt{R} script used to run \texttt{iRF} on the splicing data.
\item{\texttt{booleanSimulations.R}}: \texttt{R} script used to run \texttt{iRF} for boolean generative models (Sections \ref{simple-sims}-\ref{bigp-sims}).
\item{\texttt{enhancerSimulations.R}}: \texttt{R} script used to run \texttt{iRF} for enhancer data simulations (Section \ref{enhancer-sims}).
\item{\texttt{runtime/irf.R}}: \texttt{R} script used to run the runtime analysis for iRF (Section \ref{runtime}).
\item{\texttt{runtime/rulefit.R}}: \texttt{R} script used to run the runtime analysis for Rulefit3 (Section \ref{runtime}).
\item{\texttt{runtime/rulefit}}: \texttt{R}  package for running Rulefit3 \citep{friedman2008predictive}. The package we provide is set up for use on linux systems. Other versions are available through \href{http://statweb.stanford.edu/~jhf/R_RuleFit.html}{statweb.stanford.edu}.

\end{enumerate}

\subsubsection*{Datasets}
\begin{enumerate}
\item{\texttt{irfSuppData1.csv}}: Processed data for the enhancer case study (Supporting Data 1).
\item{\texttt{irfSuppData2.csv}}: Description of the splicing assays including ENCODE accession number, assay name, and assay type (Supporting Data 2).
\item{\texttt{irfSuppData3.csv}}: Processed data used for the splicing case study (Supporting Data 3).
\item{\texttt{enhancer.Rdata}}: An \texttt{Rdata} file containing all variables required to run the \texttt{enhancer.R} script:
	\begin{itemize}
    	\item{\texttt{X}}: $7809 \times 80$ feature matrix, rows corresponding to genomic regions and columns corresponding to assays.
  		\item{\texttt{Y}}: length $7809$ response vector, $1$ indicating active element.
  		\item{\texttt{train.id}}: length $3912$ vector giving the indices of training observations.
  		\item{\texttt{test.id}}: length $3897$ vector giving the indices of testing observations.
  		\item{\texttt{varnames.all}}: $80 \times 2$ data frame, the first column giving a unique
    		identifier for each assay and the second column giving collapsed terms used
    		to group replicate assays.
    \end{itemize}

\item{\texttt{splice.Rdata}}: An \texttt{Rdata} file containing all variables required to run the \texttt{splicing.R} script:
	\begin{itemize}
  		\item{\texttt{x}}: $23823 \times 270$ feature matrix, rows corresponding to exons and columns corresponding to assays.
  		\item{\texttt{y}}: length $23823$ response vector, $1$ indicating a highly spliced exon.
  		\item{\texttt{train.id}}: length $11911$ vector giving the indices of training observations.
  		\item{\texttt{test.id}}: length $11912$ vector giving the indices of testing observations.
  		\item{\texttt{varnames.all}}: $270 \times 2$ data frame, the first column giving a unique
    		identifier for each assay and the second column giving collapsed terms used
    		to group replicate assays.
    \end{itemize}

\item{\texttt{rfSampleSplitNoise.Rdata}}: An \texttt{Rdata} file containing RF predicted probabilities used for noising the enhancer simulation:
  	\begin{itemize}
  		\item{\texttt{pred.prob}}: $7809 \times 20$ matrix, giving the predicted probability that 				each genomic element is active. These probabilities were generated using the
    		sample splitting procedure described in Section \ref{enhancer-sims} and used to noise the enhancer
    		simulation.
		\end{itemize}
\end{enumerate}

\newpage

%% file: sections-pnas/supp-figures.tex
\newpage
\begin{figure}

\hspace{0.2in}
\textbf{A}\hspace{0.49\textwidth}\textbf{B}\\
\vspace{-0.2in}
\begin{center}
    \includegraphics[width=0.49\linewidth]{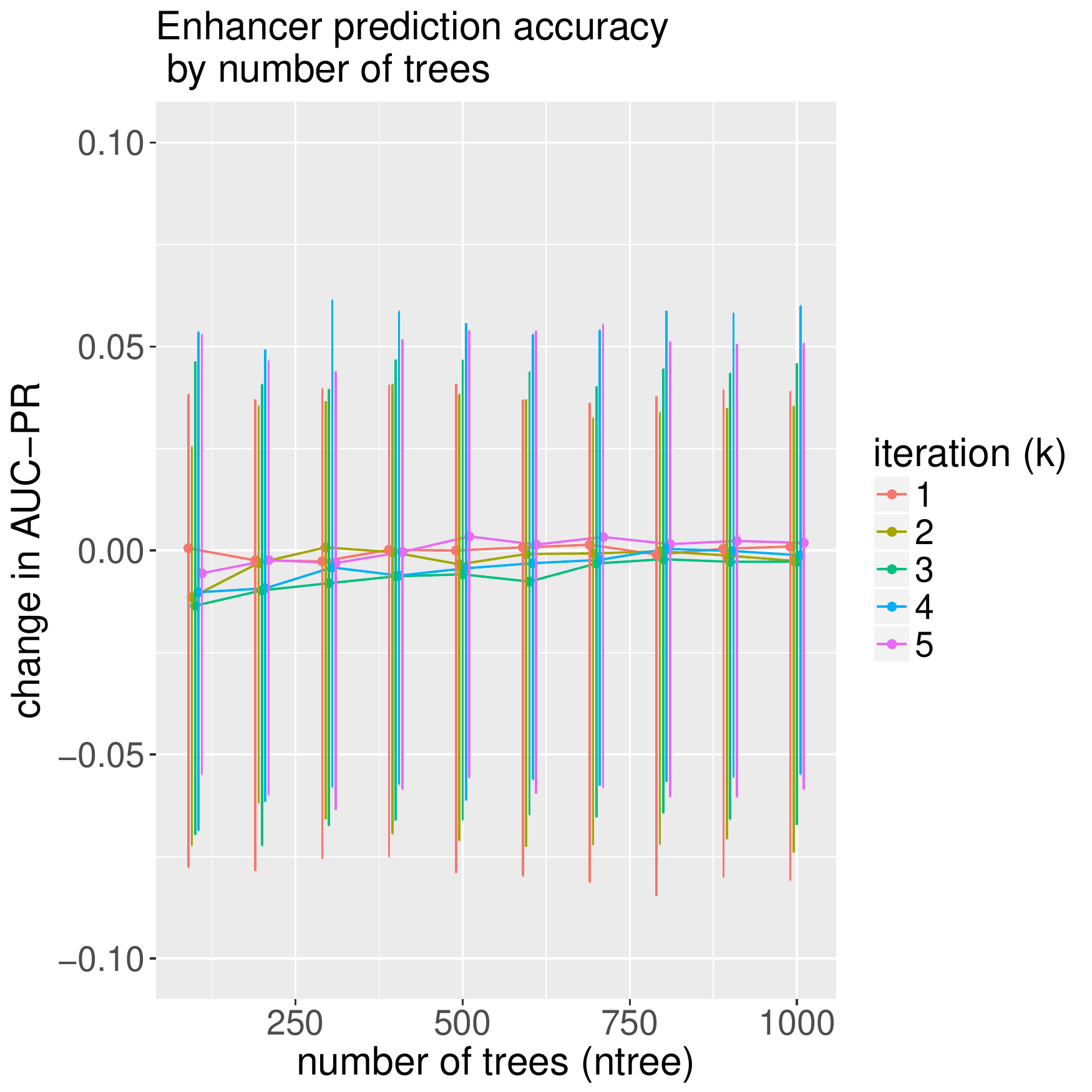}
    \includegraphics[width=0.49\linewidth]{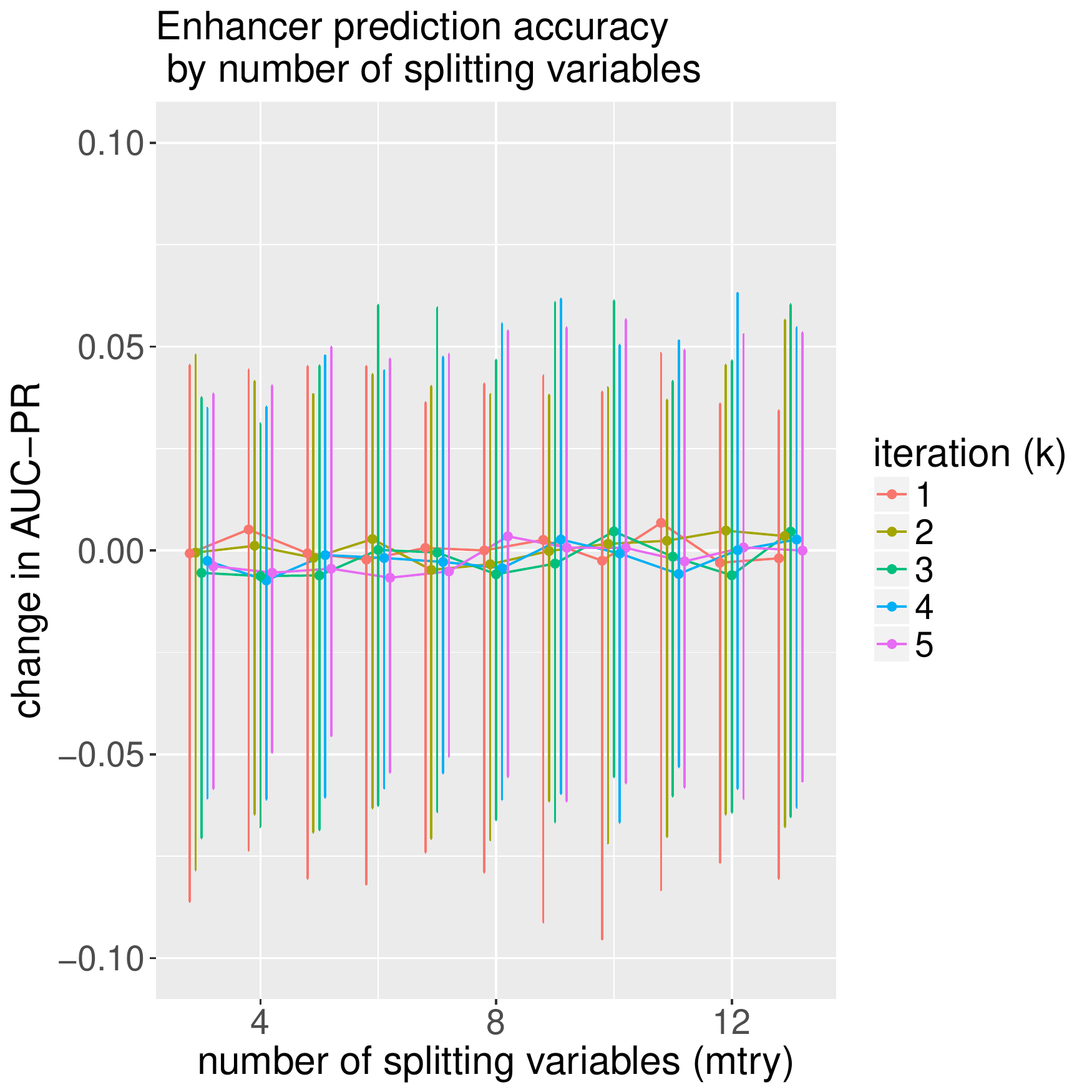}\\
\end{center}
    \caption[Enhancer prediction tuning]{Enhancer data cross-validation AUC-PR change from baseline as a function of RF tuning parameters, evaluated over $5$ folds. Baseline performance is given by Random Forest ($k=1$) with default parameters (\texttt{ntree}$=500$, \texttt{mtry}$=8$). Error bars indicate the minimum and maximum change in AUC-PR across folds.  \textbf{[A]} Prediction accuracy as a function of number of trees (\texttt{ntree}), with number of splitting variables (\texttt{mtry}) set to default ($\lfloor{\sqrt{p}\rfloor}=8$). \textbf{[B]} Prediction accuracy as a function of \texttt{mtry}, with \texttt{ntree} set to default ($500$).\label{fig:enhancer-cv}}
\end{figure}

\begin{figure}

\hspace{0.2in}
\textbf{A}\hspace{0.49\textwidth}\textbf{B}\\
\vspace{-0.2in}
\begin{center}
    \includegraphics[width=0.49\linewidth]{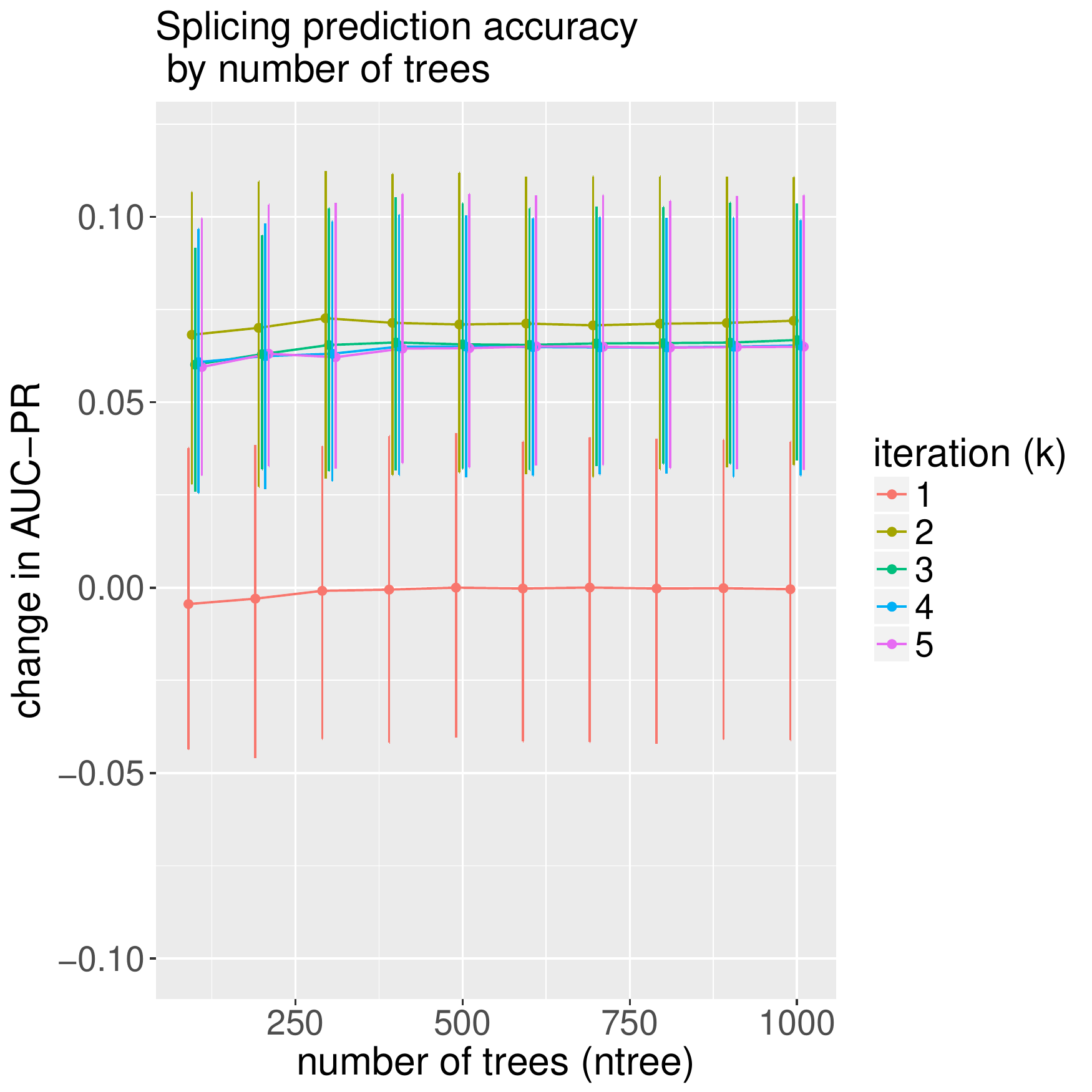}
    \includegraphics[width=0.49\linewidth]{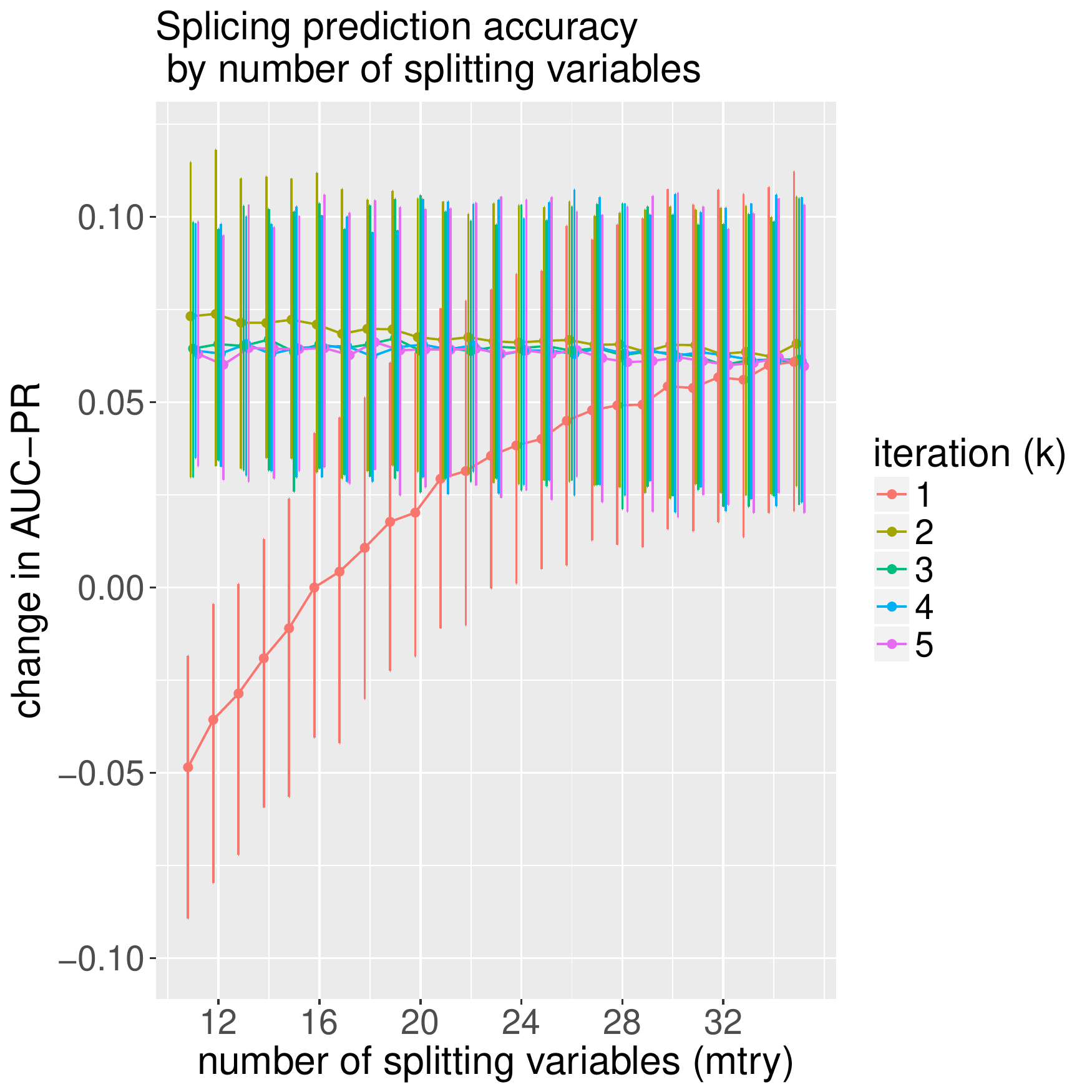}\\
\end{center}
    \caption[Splicing prediction tuning]{Splicing data cross-validation AUC-PR change from baseline as a function of RF tuning parameters, evaluated over $5$-folds. Baseline performance is given by Random Forest ($k=1$) with default parameters (\texttt{ntree}$=500$, \texttt{mtry}$=16$). Error bars indicate the minimum and maximum change in AUC-PR across folds. For iterations $k>1$, performance is robust to choice of tuning parameters. \textbf{[A]} Prediction accuracy as a function of number of trees (\texttt{ntree}), with the number of splitting variables (\texttt{mtry}) set to default ($\lfloor{\sqrt{p}\rfloor}=16$). \textbf{[B]} Prediction accuracy as a function of \texttt{mtry}, with \texttt{ntree} set to default ($500$).\label{fig:splice-cv}}
\end{figure}

\newpage
\begin{figure}

\hspace{0.2in}
\textbf{A}\hspace{0.49\textwidth}\textbf{B}\\
\vspace{-0.2in}
\begin{center}
    \includegraphics[width=0.49\linewidth]{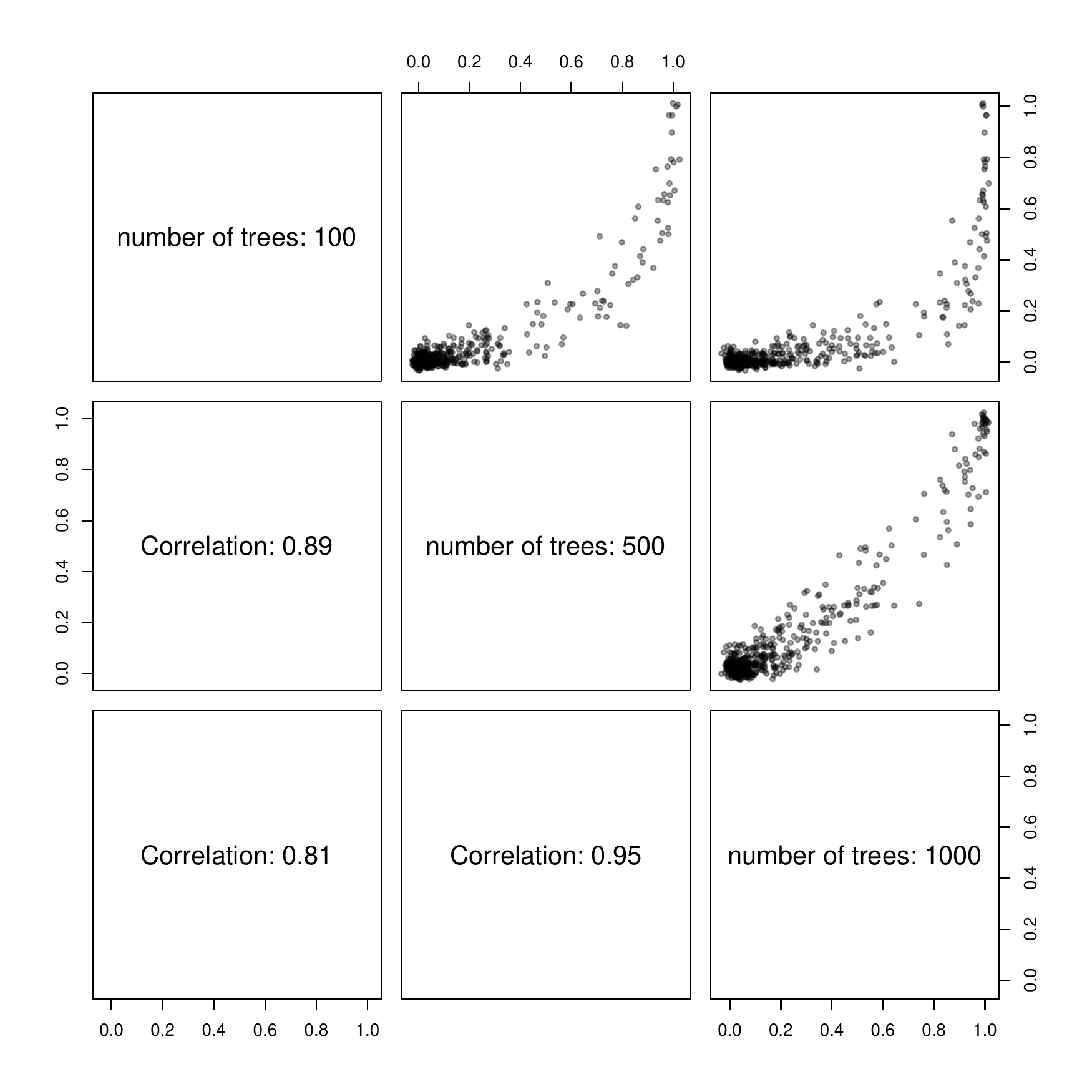}
    \includegraphics[width=0.49\linewidth]{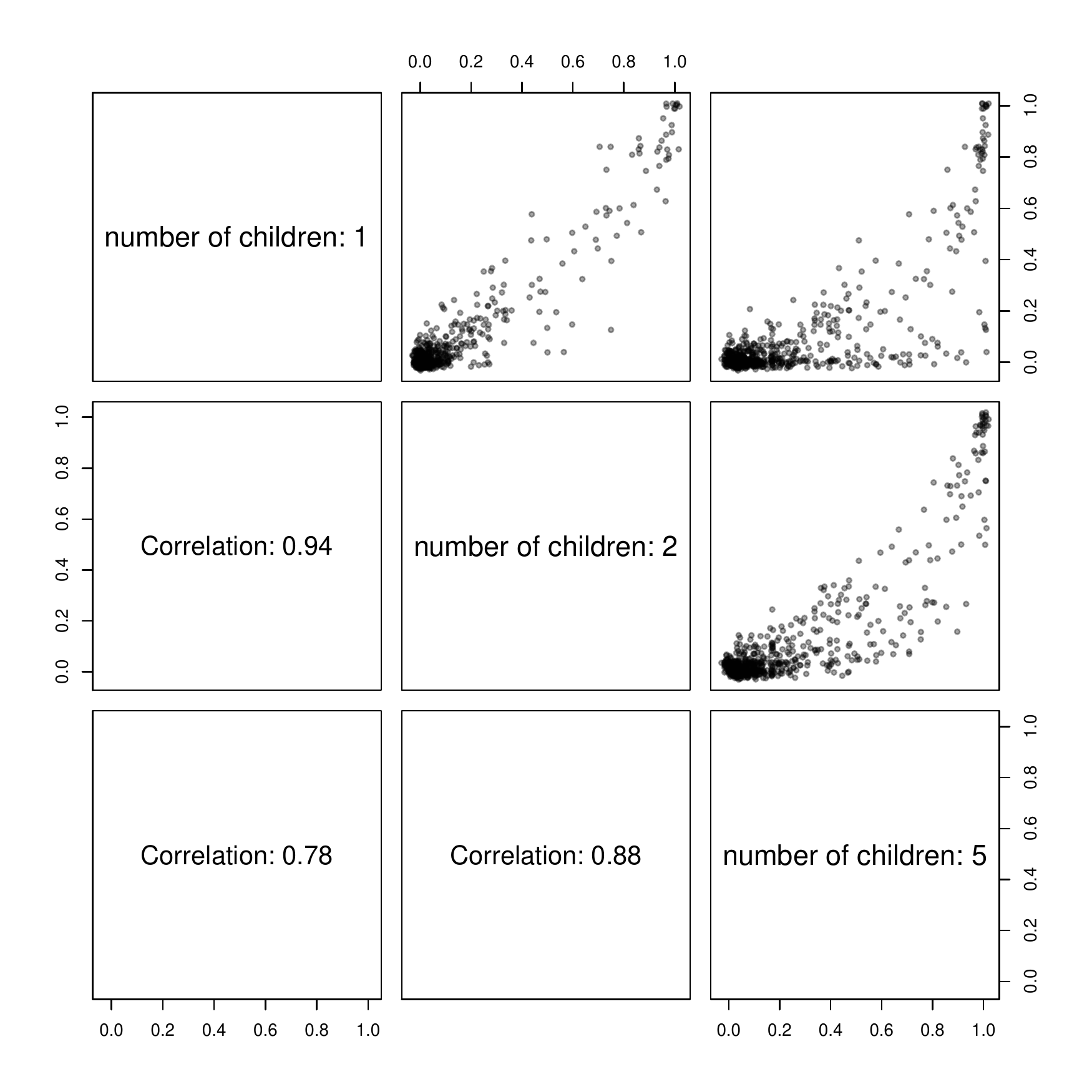}\\
\end{center}

\hspace{1.95in}
\textbf{C}
\vspace{-0.2in}
\begin{center}
    \includegraphics[width=0.49\linewidth]{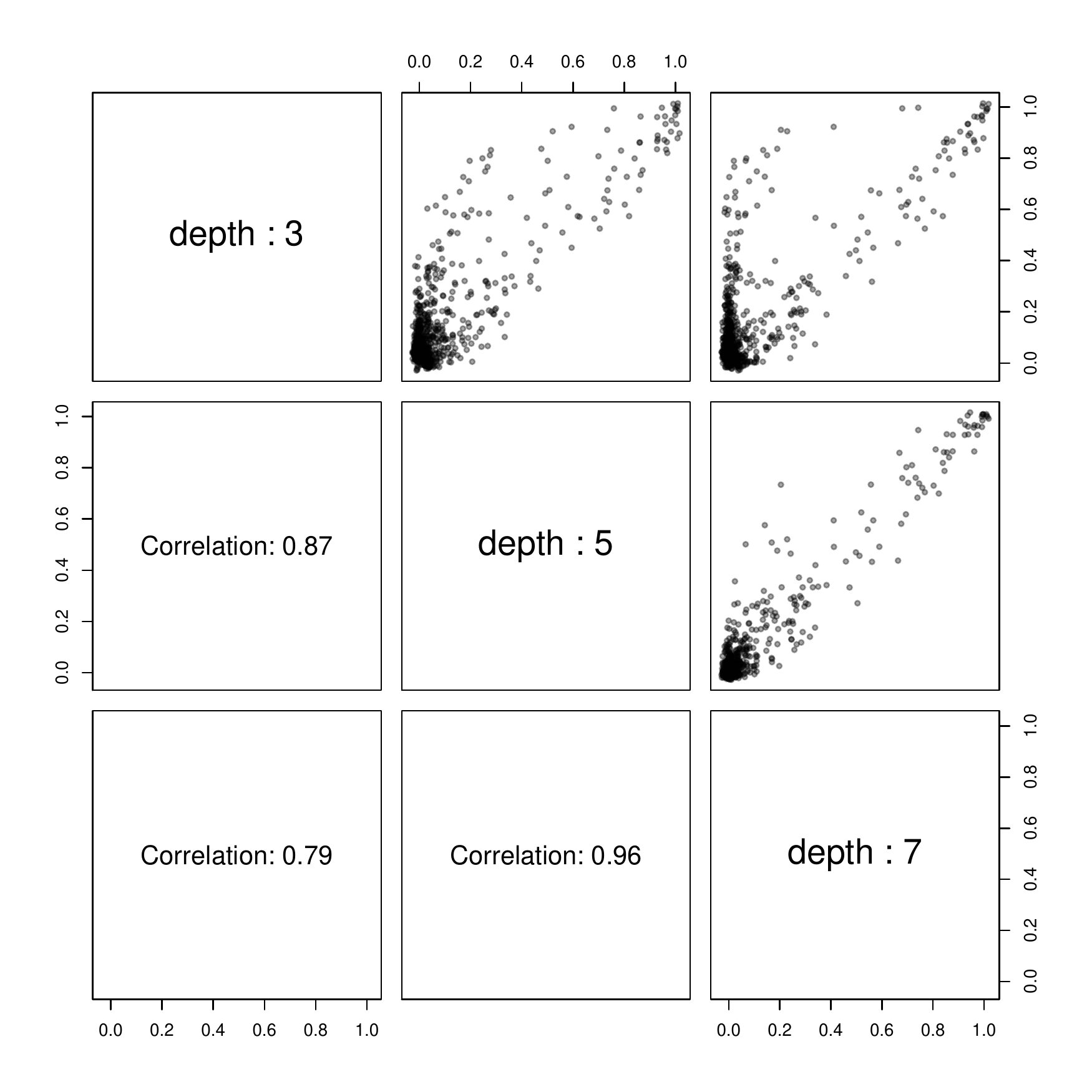}
\end{center}
    \caption[Enhancer interaction recovery tuning]{Enhancer data interaction stability scores as a function of RIT parameters. Each point represents a single interaction, and the point's coordinates indicate its stability score under two parameter settings. Lower panels give Pearson correlation between interaction stability scores across pairs of parameter settings. \textbf{[A]} Interaction stability scores as a function of the number of trees in RIT. Number of children and depth are set to default levels of $2$ and $5$ respectively. \textbf{[B]} Interaction stability scores as a function of number of children in RIT. Number of trees and depth are set to default levels of $500$ and $5$ respectively. \textbf{[C]} Interaction stability scores as a function of depth in RIT. Number of trees and number of children are set to default levels of $500$ and $2$ respectively.\label{fig:enhancer-cv-rit}}
\end{figure}

\begin{figure}

\hspace{0.2in}
\textbf{A}\hspace{0.49\textwidth}\textbf{B}\\
\vspace{-0.2in}
\begin{center}
    \includegraphics[width=0.49\linewidth]{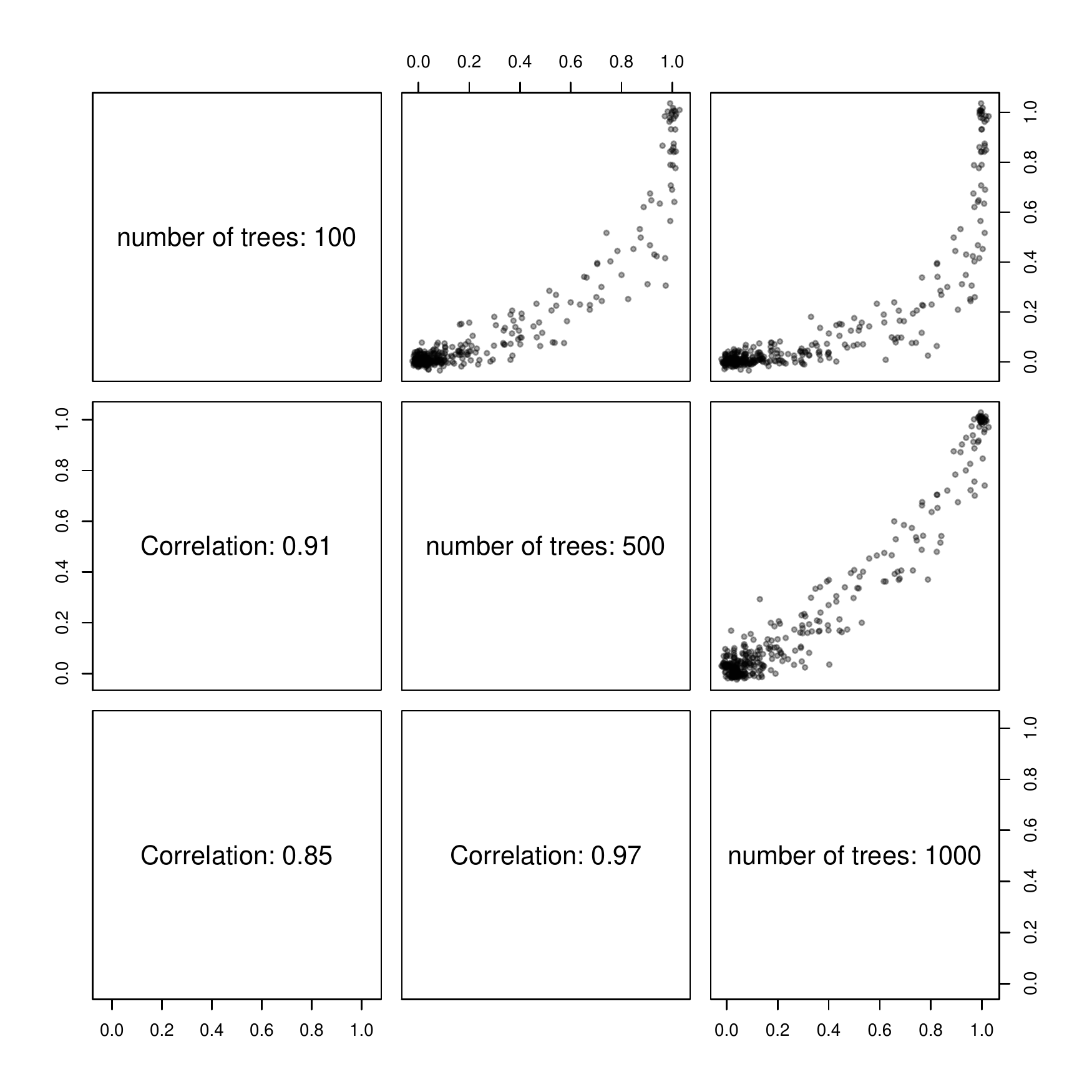}
    \includegraphics[width=0.49\linewidth]{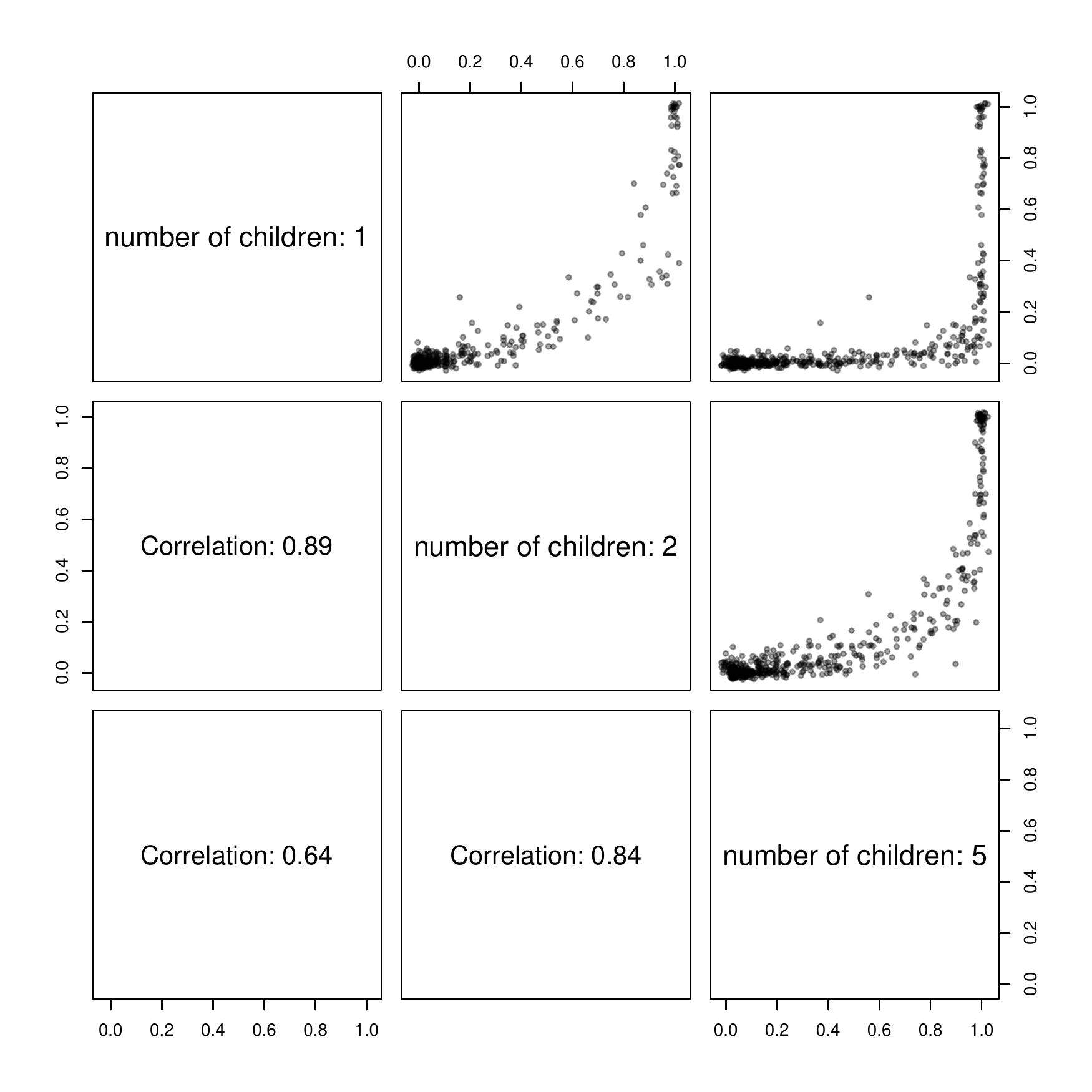}\\
\end{center}

\hspace{1.95in}
\textbf{C}
\vspace{-0.2in}
\begin{center}
    \includegraphics[width=0.49\linewidth]{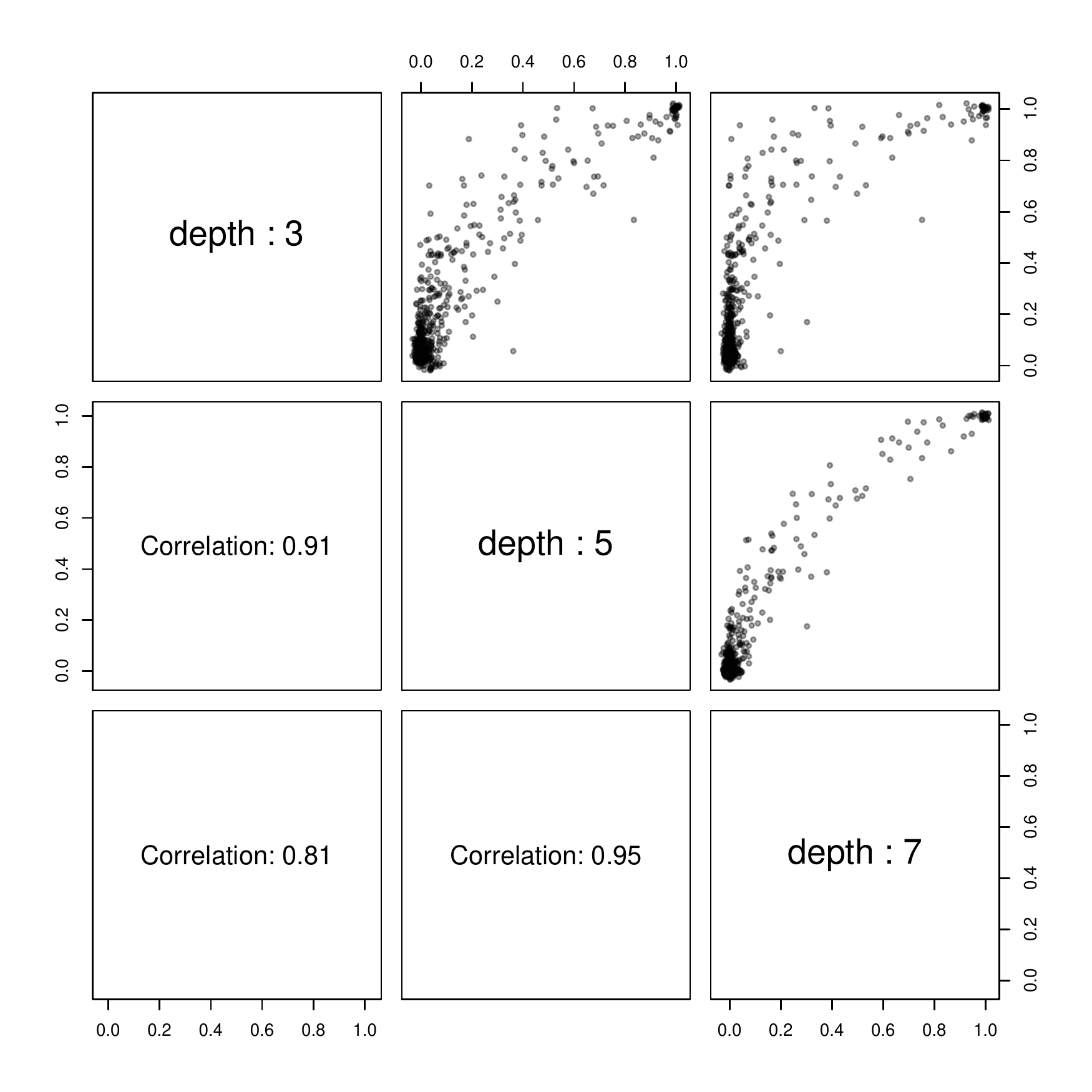}
\end{center}
    \caption[Splicing interaction recovery tuning]{Splicing data interaction stability scores as a function of RIT parameters. Each point represents a single interaction, and the point's coordinates indicate its stability score under two parameter settings. Lower panels give Pearson correlation between interaction stability scores across pairs of parameter settings. \textbf{[A]} Interaction stability scores as a function of the number of trees in RIT. Number of children and depth are set to default levels of $2$ and $5$ respectively. \textbf{[B]} Interaction stability scores as a function of number of children in RIT. Number of trees and depth are set to default levels of $500$ and $5$ respectively. \textbf{[C]} Interaction stability scores as a function of depth in RIT. Number of trees and number of children are set to default levels of $500$ and $2$ respectively.\label{fig:splice-cv-rit}}
\end{figure}

\begin{figure}[h]
	\textbf{A}\\ 
    \vspace{-0.4in}
    \begin{center}
    \includegraphics[width=0.9\linewidth]{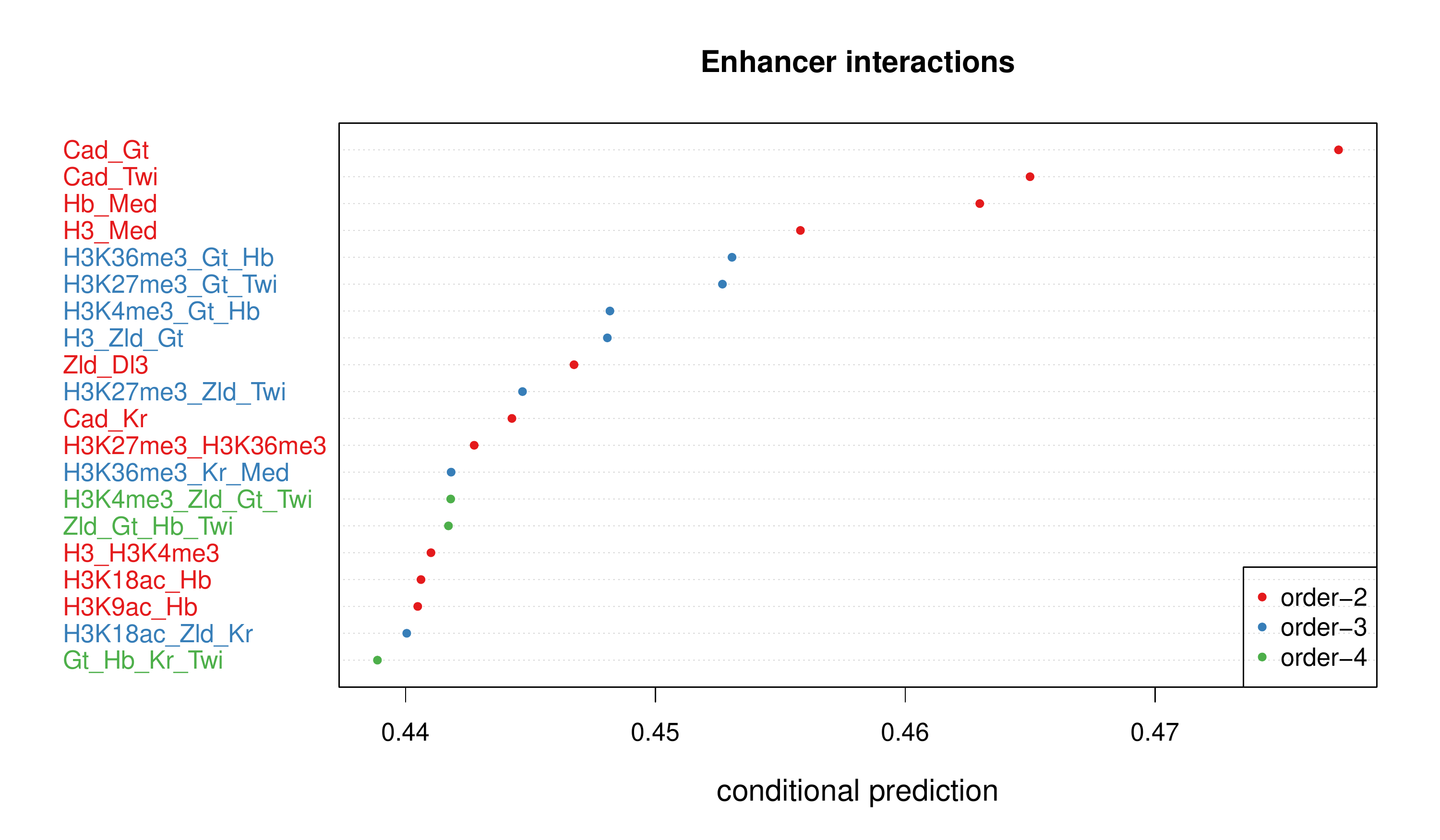}\\
    \end{center}
    \textbf{B}\\
     \vspace{-0.4in}
    \begin{center}
    \includegraphics[width=0.9\linewidth]{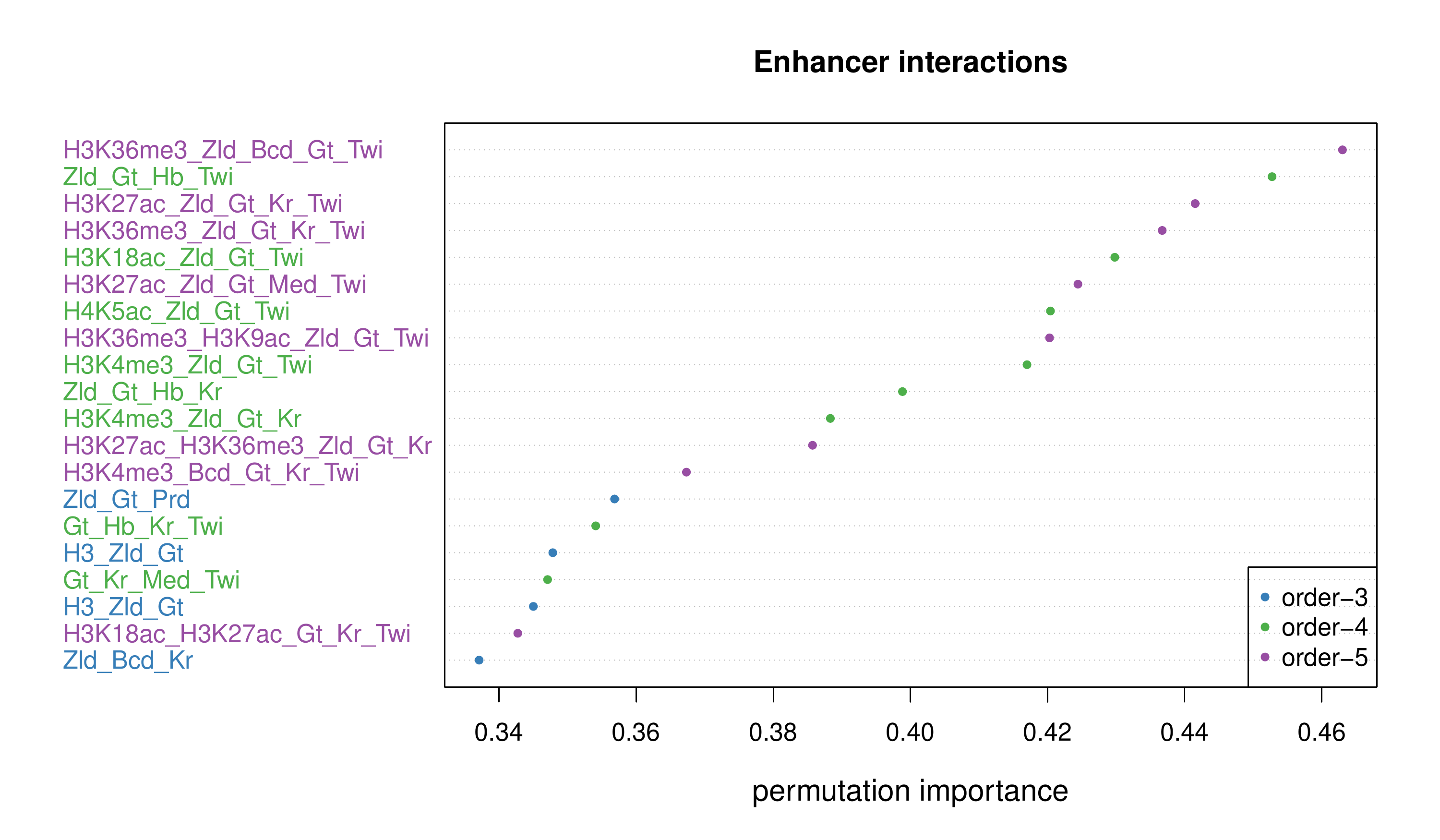}
    \end{center}
    \caption[Prediction-based interaction importance: enhancer data]{Prediction-based validation metrics for enhancer data. Each plot shows the top 20 interactions with respect to prediction based importance metrics. Lower-order interactions that are a strict subset of some higher-order interactions have been removed for clearer visualization. The interactions reported here are qualitatively similar to those with high stability scores. \textbf{[A]} Conditional prediction. \textbf{[B]} Permutation importance. 
    \label{fig:importance-metrics-enhancer}}
\end{figure}

\begin{figure}[p]
	\textbf{A}\\ 
    \vspace{-0.4in}
    \begin{center}
    \includegraphics[width=0.9\linewidth]{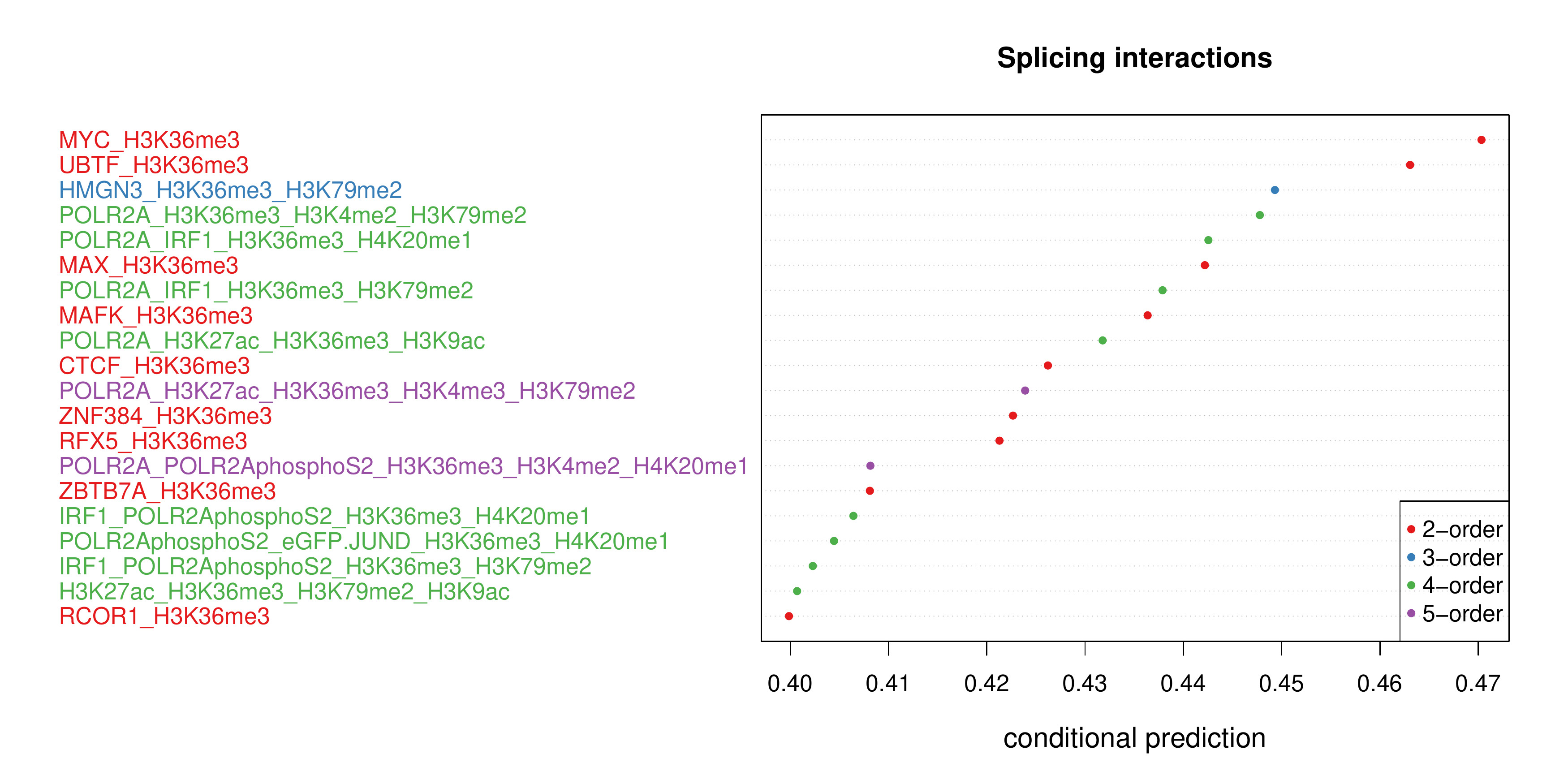}\\
    \end{center}
    \textbf{B}\\ 
    \vspace{-0.4in}
    \begin{center}
    \includegraphics[width=0.9\linewidth]{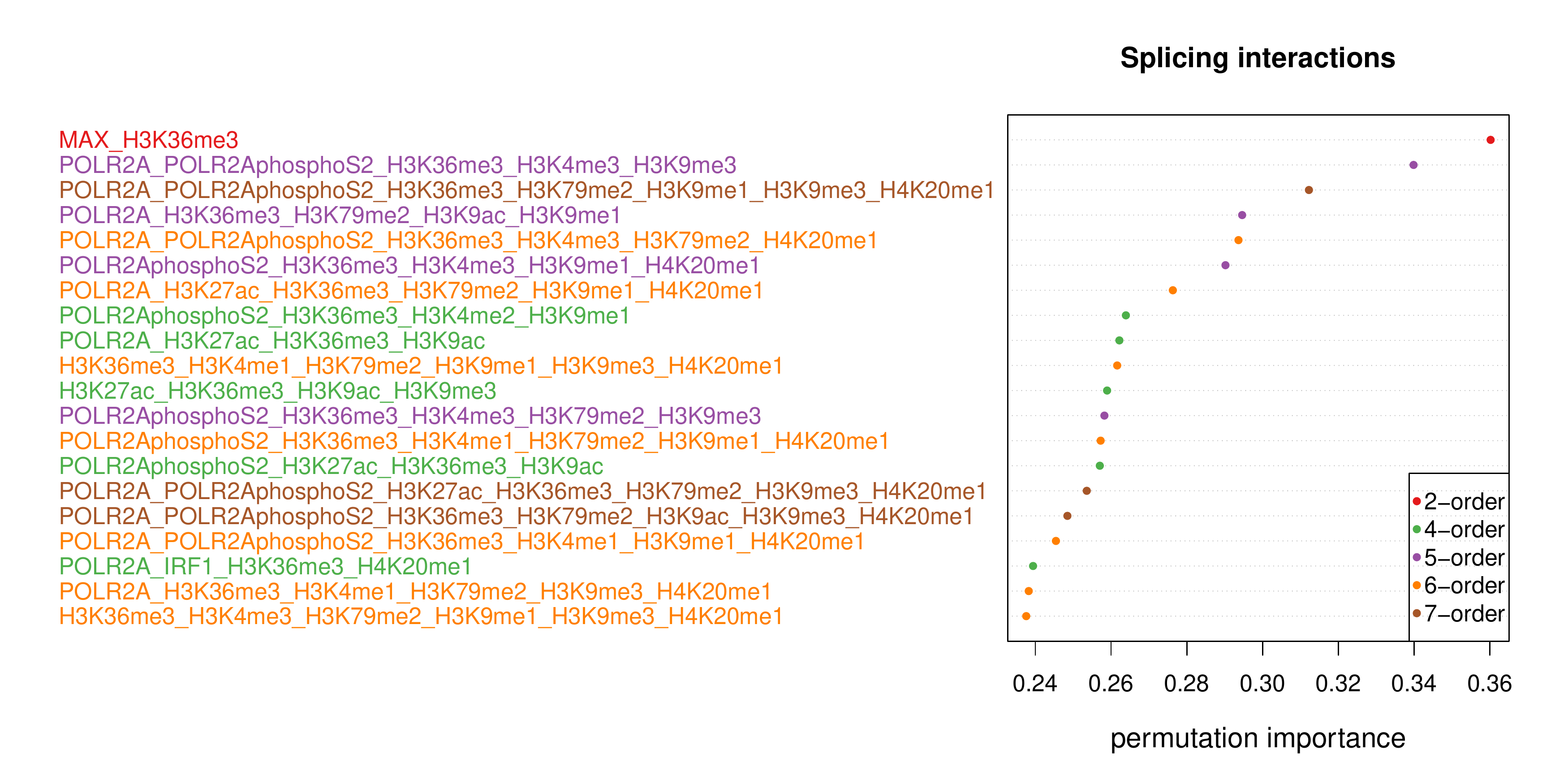}
    \end{center}
    \caption[Prediction-based interaction importance: splicing data]{Prediction-based validation metrics for splicing data. Each plot shows the top 20 interactions with respect to prediction based importance metrics. Lower-order interactions that are a strict subset of recovered  higher-order interactions have been removed for clearer visualization. \textbf{[A]} Conditional prediction. \textbf{[B]} Permutation importance. The interactions reported here are qualitatively similar to those with high stability scores. \label{fig:importance-metrics-splicing}}
\end{figure}

\begin{figure}[p]
    \textbf{A}\hspace{0.49\textwidth}\textbf{B}\\

    \includegraphics[width=0.49\linewidth]{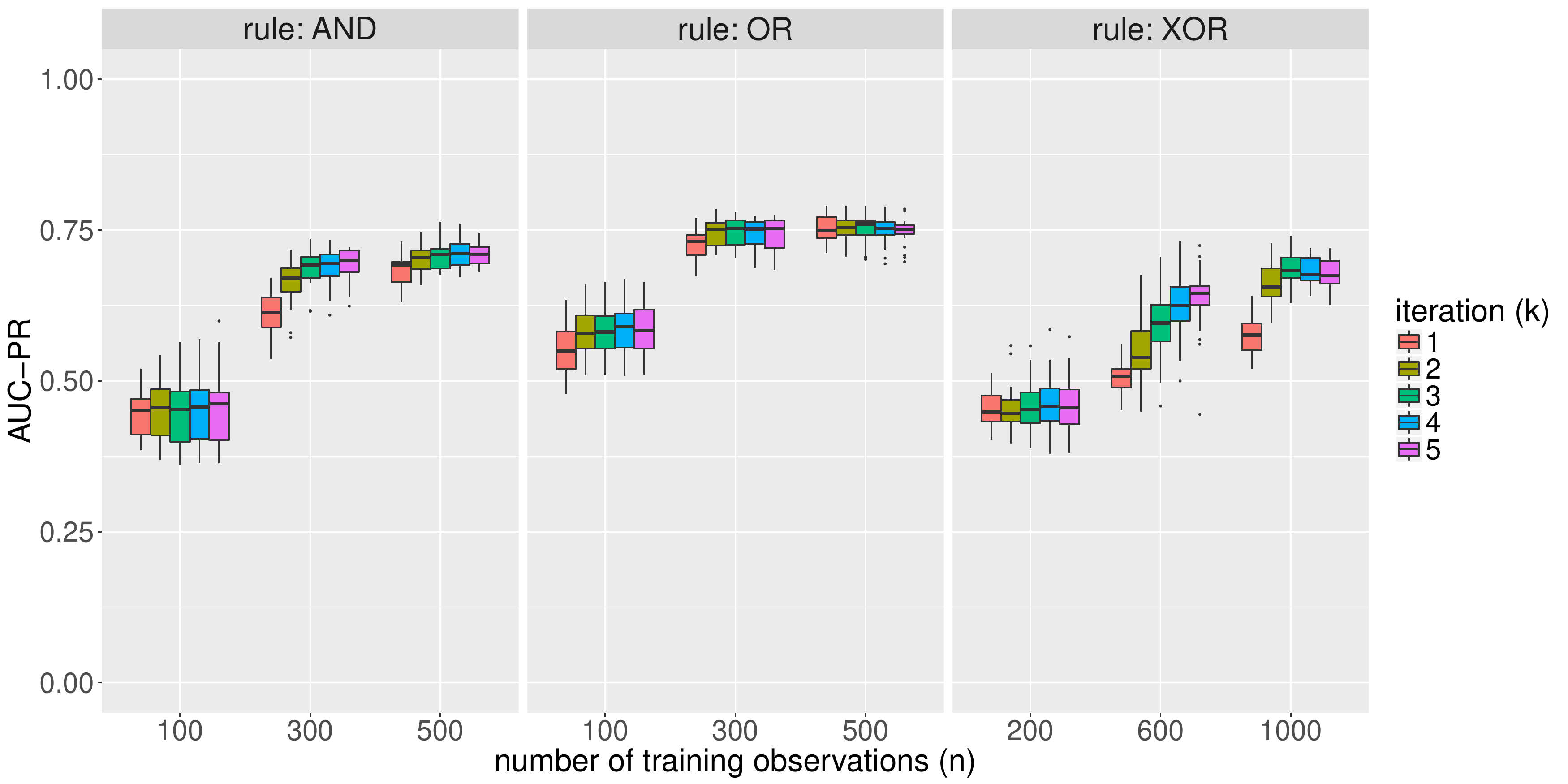}
    \includegraphics[width=0.49\linewidth]{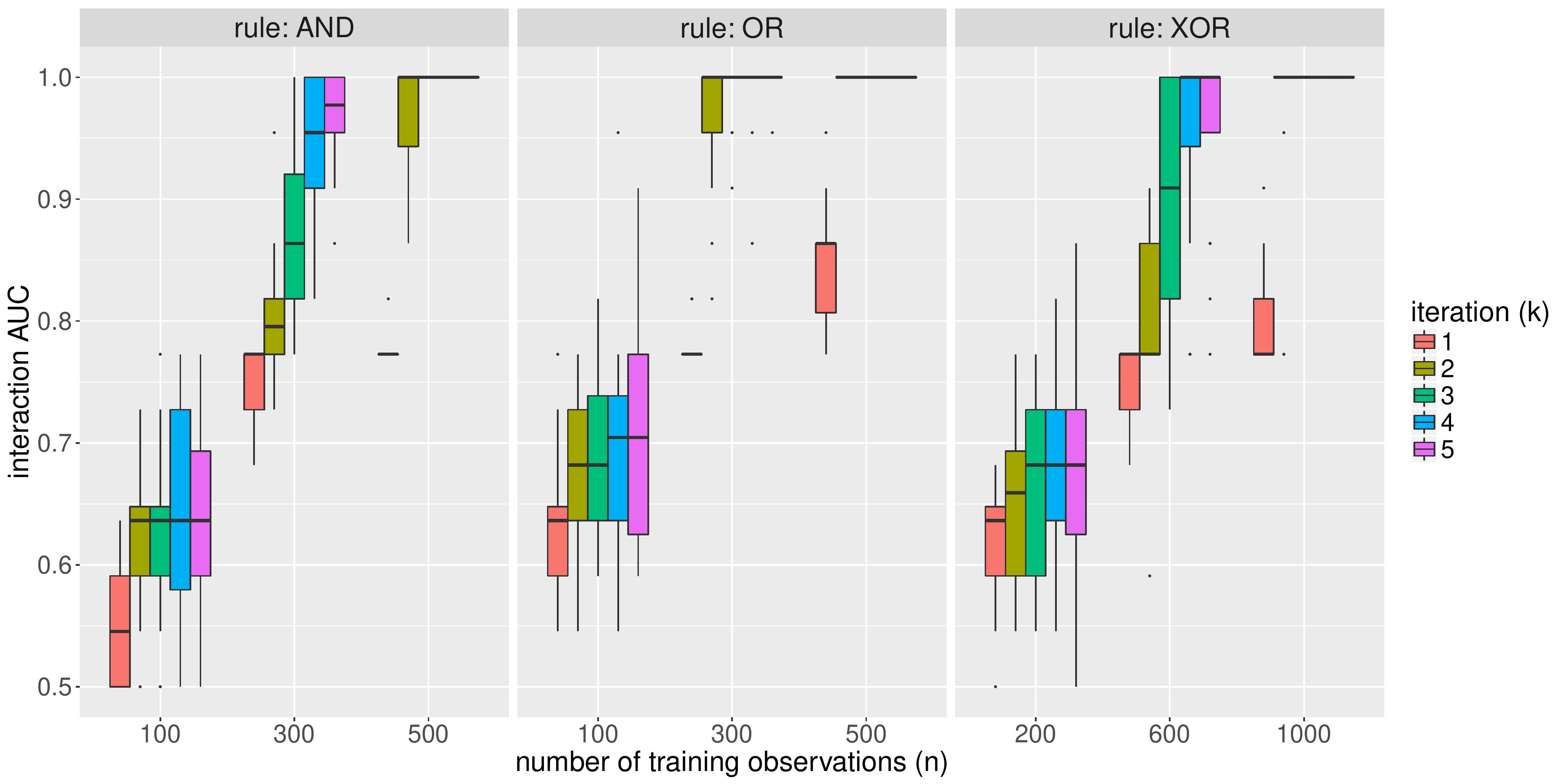}\\
    
    \textbf{C}\hspace{0.49\textwidth}\textbf{D}\\
    
    \includegraphics[width=0.49\linewidth]{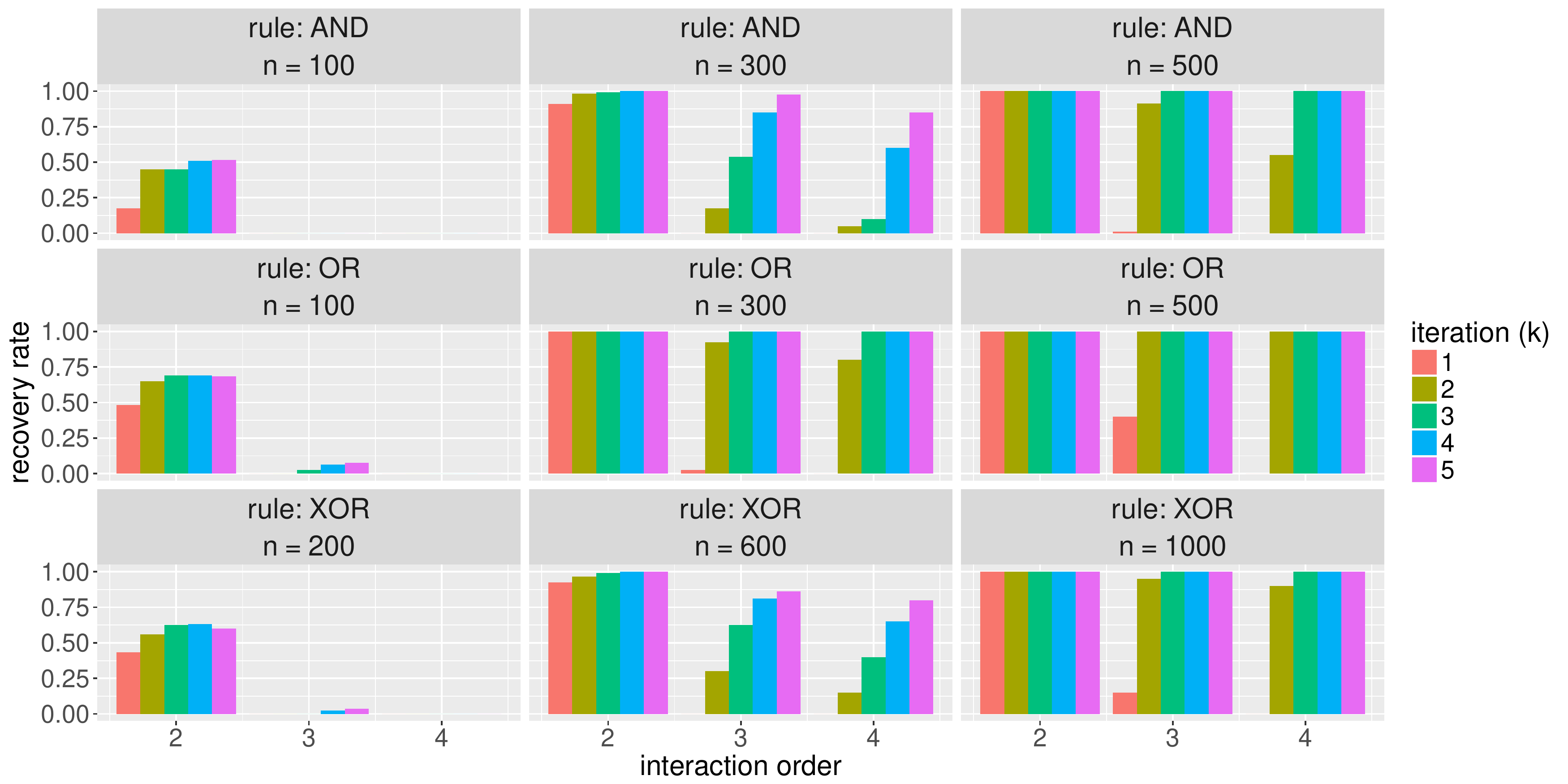}
    \includegraphics[width=0.49\linewidth]{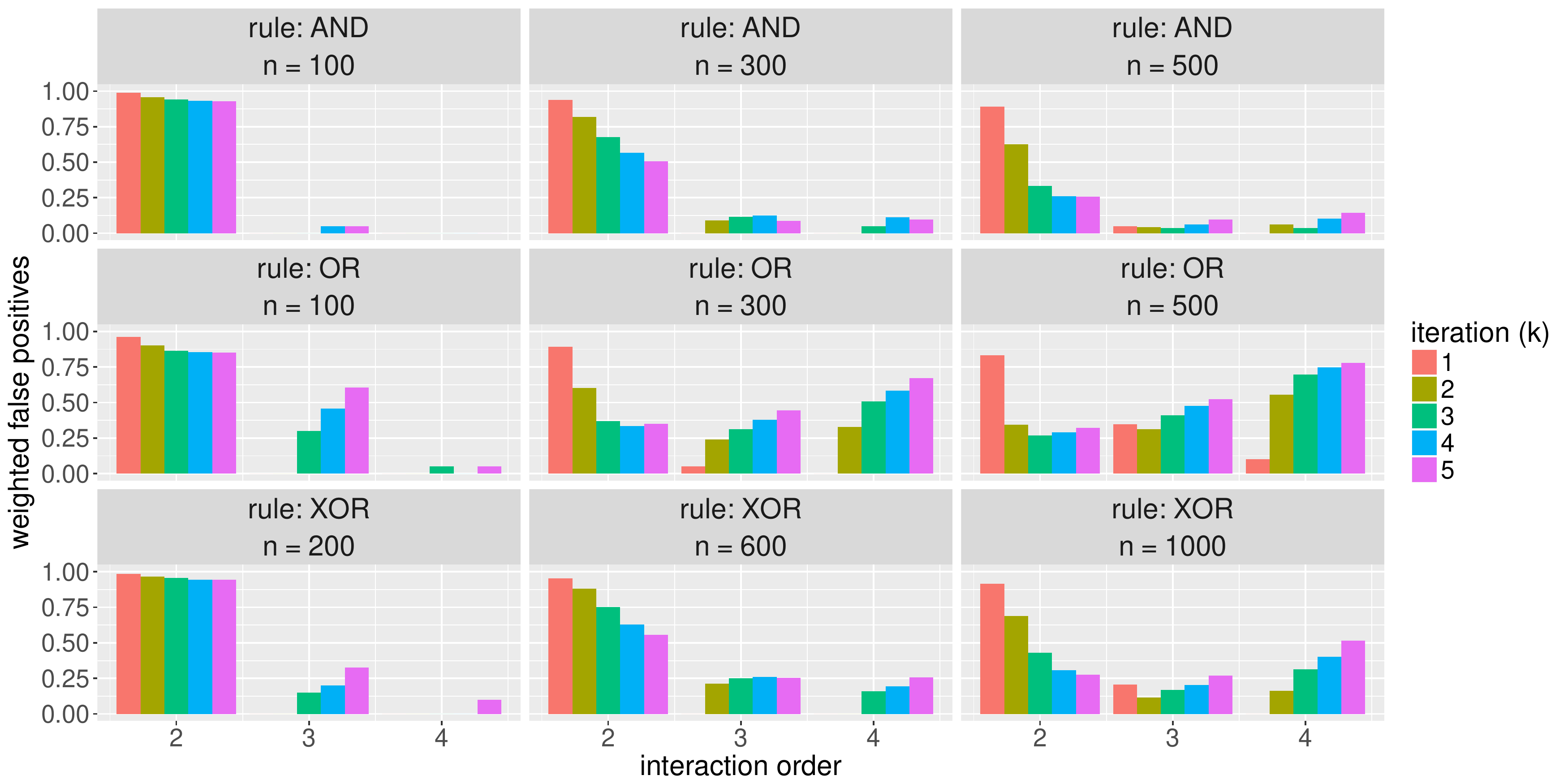} 

\caption[iRF performance for order-$4$ AND, OR, and XOR rules]{iRF performance for order-$4$ AND, OR, and XOR rules over 20 replicates. Results are shown for models trained using $100, 300,$ and $500$ observations in the AND and OR models. Training sample size is increased to $200, 600,$ and $1000$ in the XOR model to account for the low marginal importance of features under this rule. \textbf{[A]} Prediction accuracy (AUC-PR) improves with increased number of training observations and is comparable or improves for increasing $k$. \textbf{[B]} Interaction AUC improves with increasing $k$. For larger values of $k$, iRF always recovers the full data generating rule as the most stable interaction (AUC of $1$) with enough training observations. \textbf{[C]} Recovery rate for interactions of all orders improves with increasing $k$. In particular, $k=1$ fails to recover any order-$4$ interactions. \textbf{[D]} Weighted false positives increase in settings where iRF recovers high-order interactions as a result of many false positives with low stability scores.}

\label{fig:simple}
\end{figure}

\begin{figure}[p]
    \textbf{A}\hspace{0.49\textwidth}\textbf{B}\\

    \includegraphics[width=0.49\linewidth]{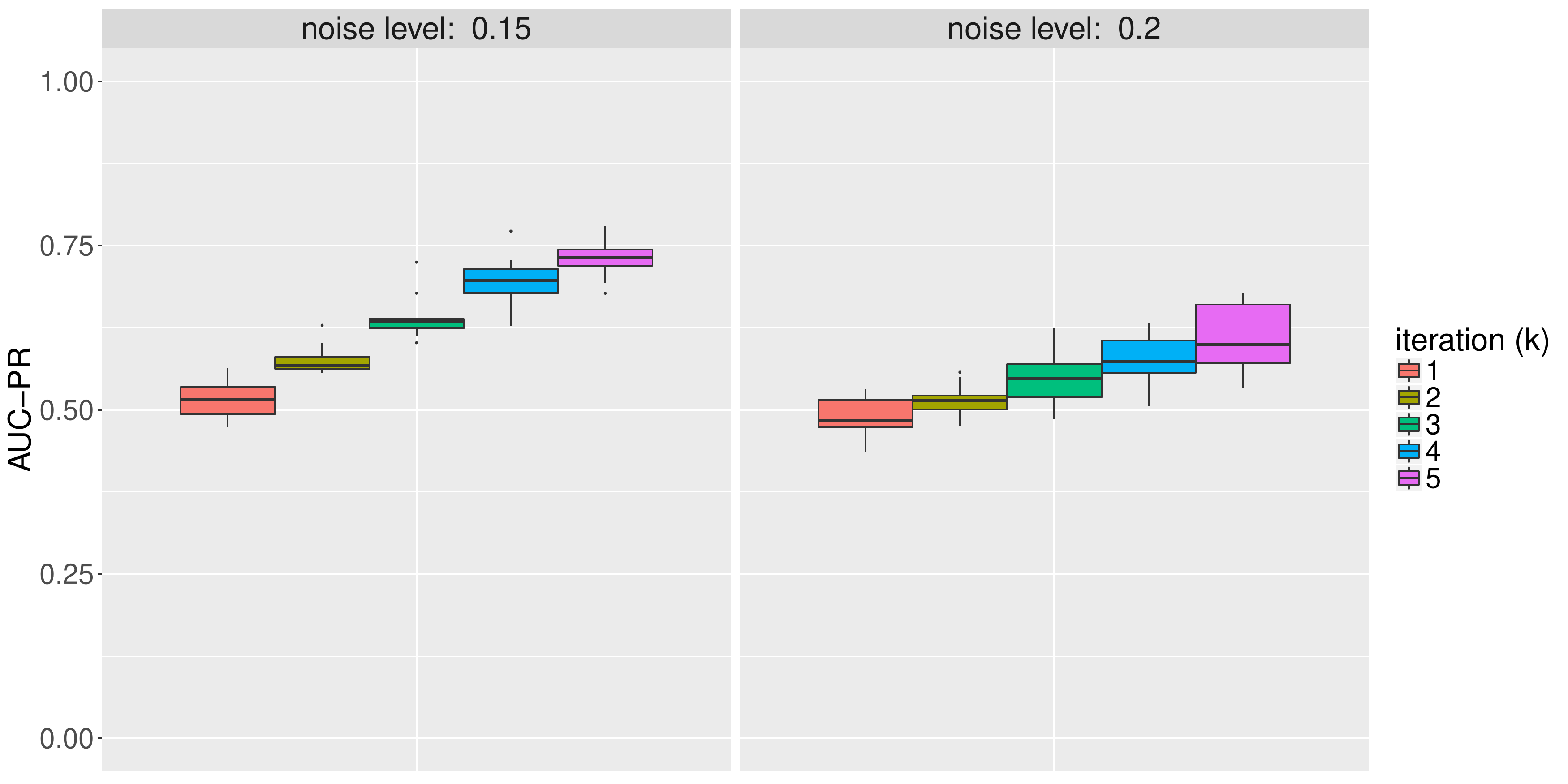}
    \includegraphics[width=0.49\linewidth]{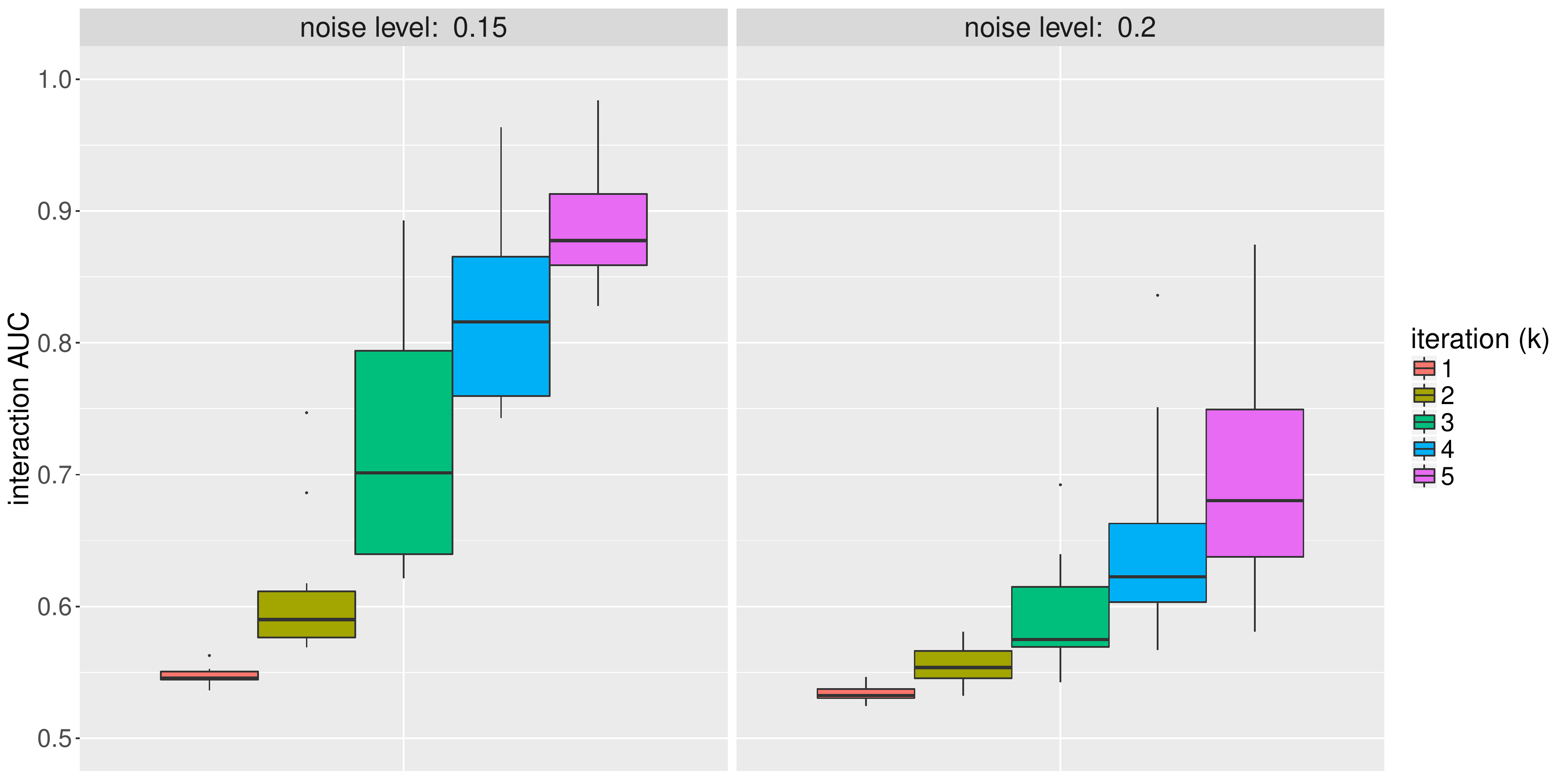}\\
    
    \textbf{C}\hspace{0.49\textwidth}\textbf{D}\\
    
    \includegraphics[width=0.49\linewidth]{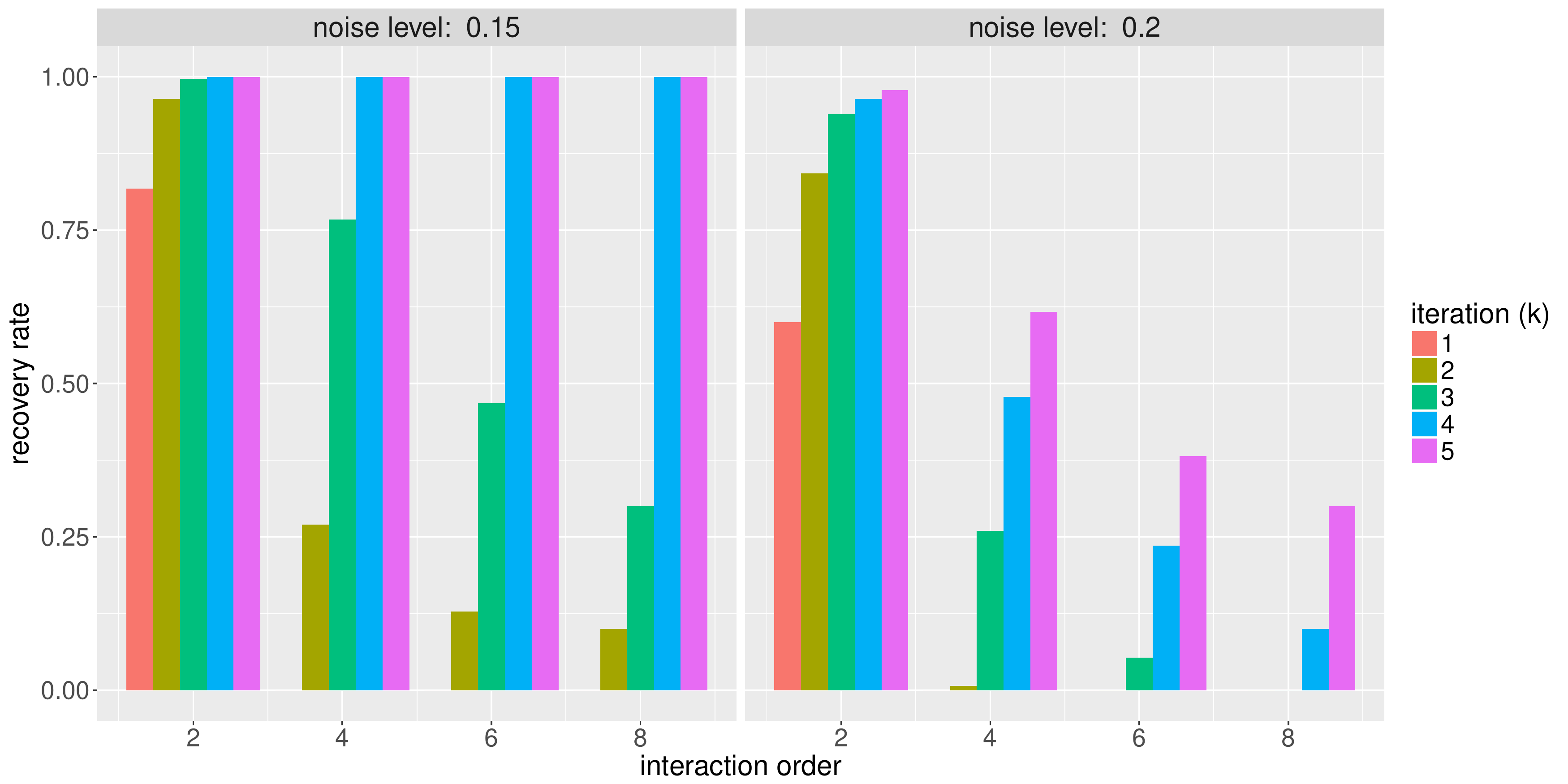}
    \includegraphics[width=0.49\linewidth]{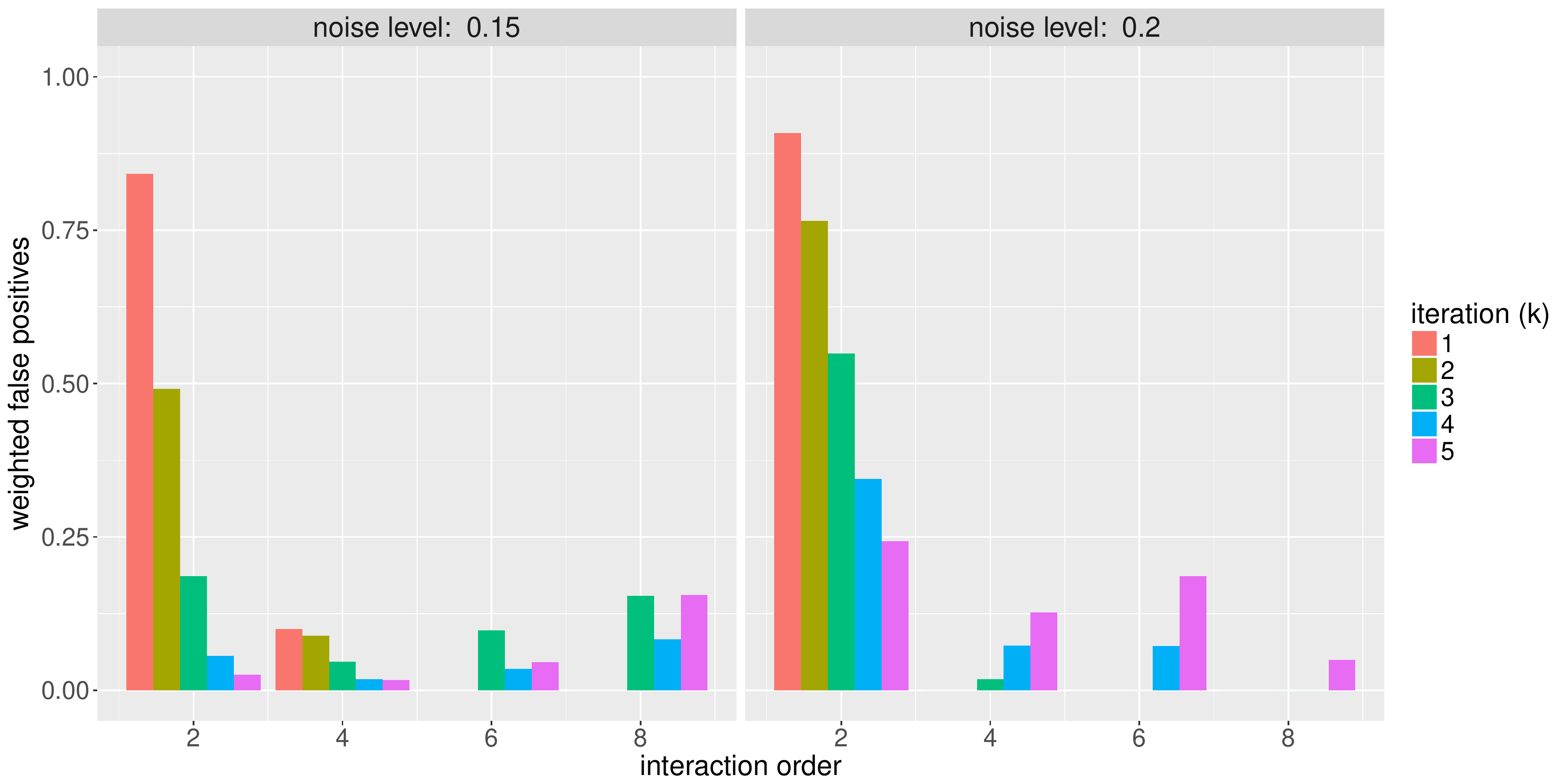} 
\caption[iRF performance for order-$8$ XOR rule]{iRF performance for order-$8$ XOR
rule over 10 replicates as a function of noise level. All models were trained using
$5,000$ observations. \textbf{[A]} Prediction accuracy (AUC-PR) improves for increasing $k$ and at a slower rate for increased noise levels. \textbf{[B]} Interaction AUC improves with increasing $k$. \textbf{[C]} Recovery rate for interactions of all orders improves with increasing $k$. In particular, $k=1$ does not recover any interactions of order $ > 2$ at either noise level. Recovery of higher order interactions drops substantially at higher noise levels. \textbf{[D]} Weighted false positives increase in settings where iRF recovers high-order interactions as a result of many false positives with low stability scores. For order-$2$ interactions, later iterations of iRF filter out many of the false positives identified in earlier iterations.}
\label{fig:xorNoise}
\end{figure}

\begin{figure}[p]
\centering
\includegraphics[width=0.9\textwidth]{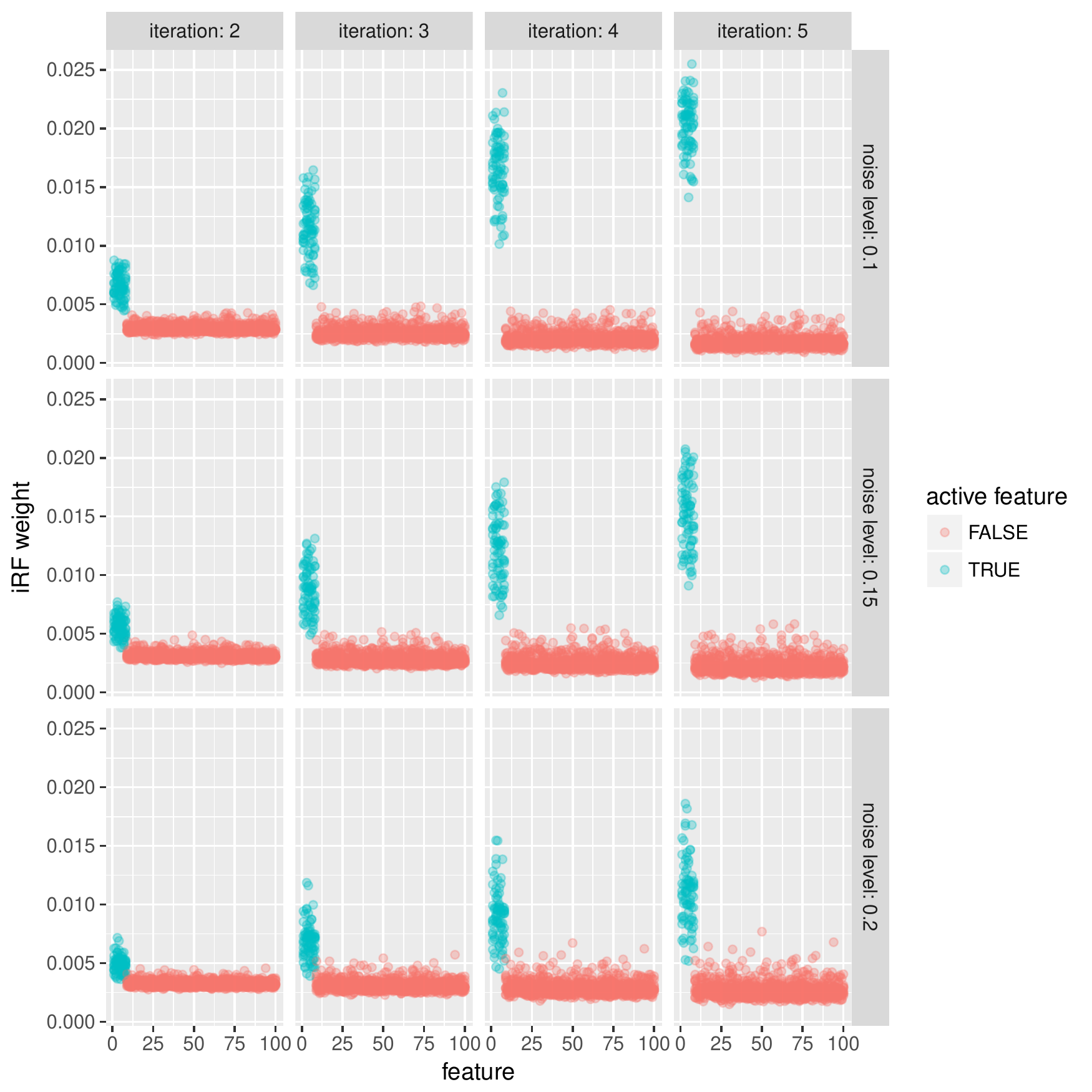}
\caption[Order-$8$ XOR rule weight distribution by iteration]{iRF weights for active (blue) and inactive (red) features as a function of iteration and noise level over 10 replicates. The distribution of weights in later iterations shows a clear separation between active and inactive features, indicating that iRF has identified active features as important and incorporates them into the model with higher probability in later iterations.}
\label{fig:xor-weights}
\end{figure}

\begin{figure}[p]
\centering
\includegraphics[width=0.9\textwidth]{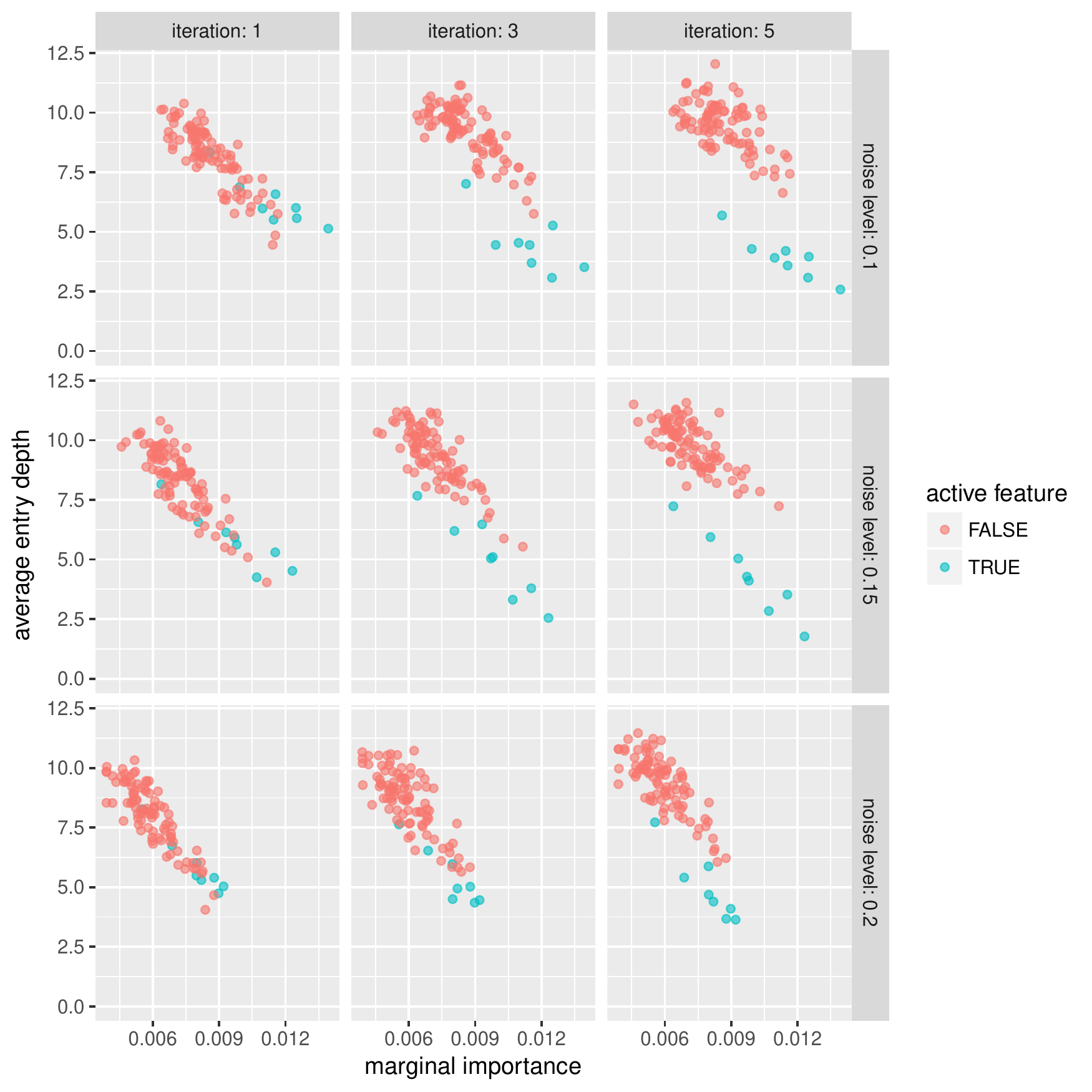}
\caption[Feature entry depth by marginal importance]{Average entry depth for active (blue) and inactive (red) features across the forest as a function of marginal importance, iteration, and noise level. Results are reported for a single replicate. In later iterations, the average depth at which active variables are selected is noticeably lower than inactive variables with comparable marginal importance, indicating that the active features appear earlier on decision paths.}
\label{fig:xor-depth}
\end{figure}

\begin{figure}[p]
  \hspace{1.3in} \textbf{A}
  \begin{center}
  \includegraphics[width=0.9\linewidth]{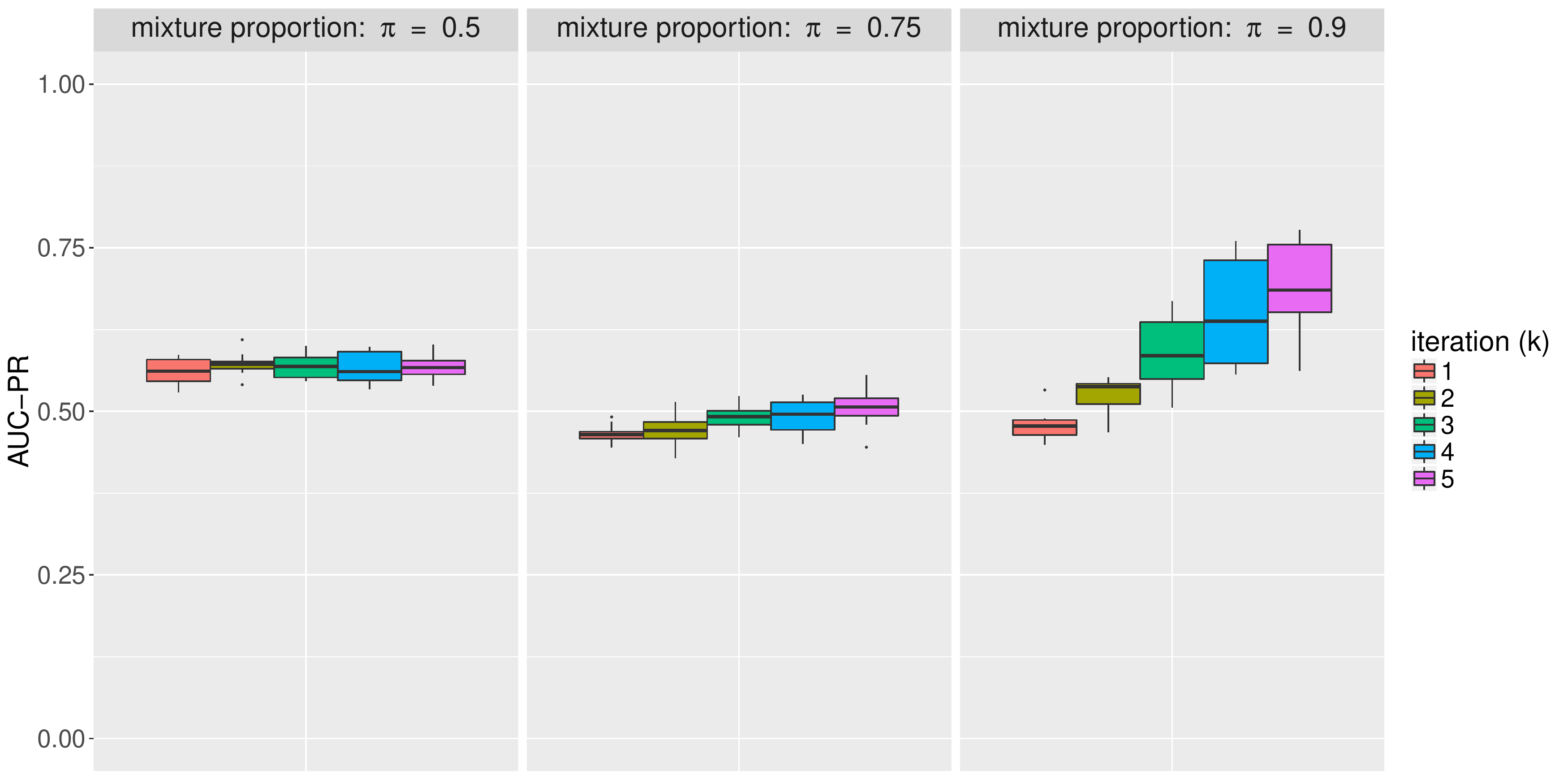}\\
  \end{center}
  \textbf{B}\hspace{0.49\textwidth}\textbf{C}\\
  
  \includegraphics[width=0.49\linewidth]{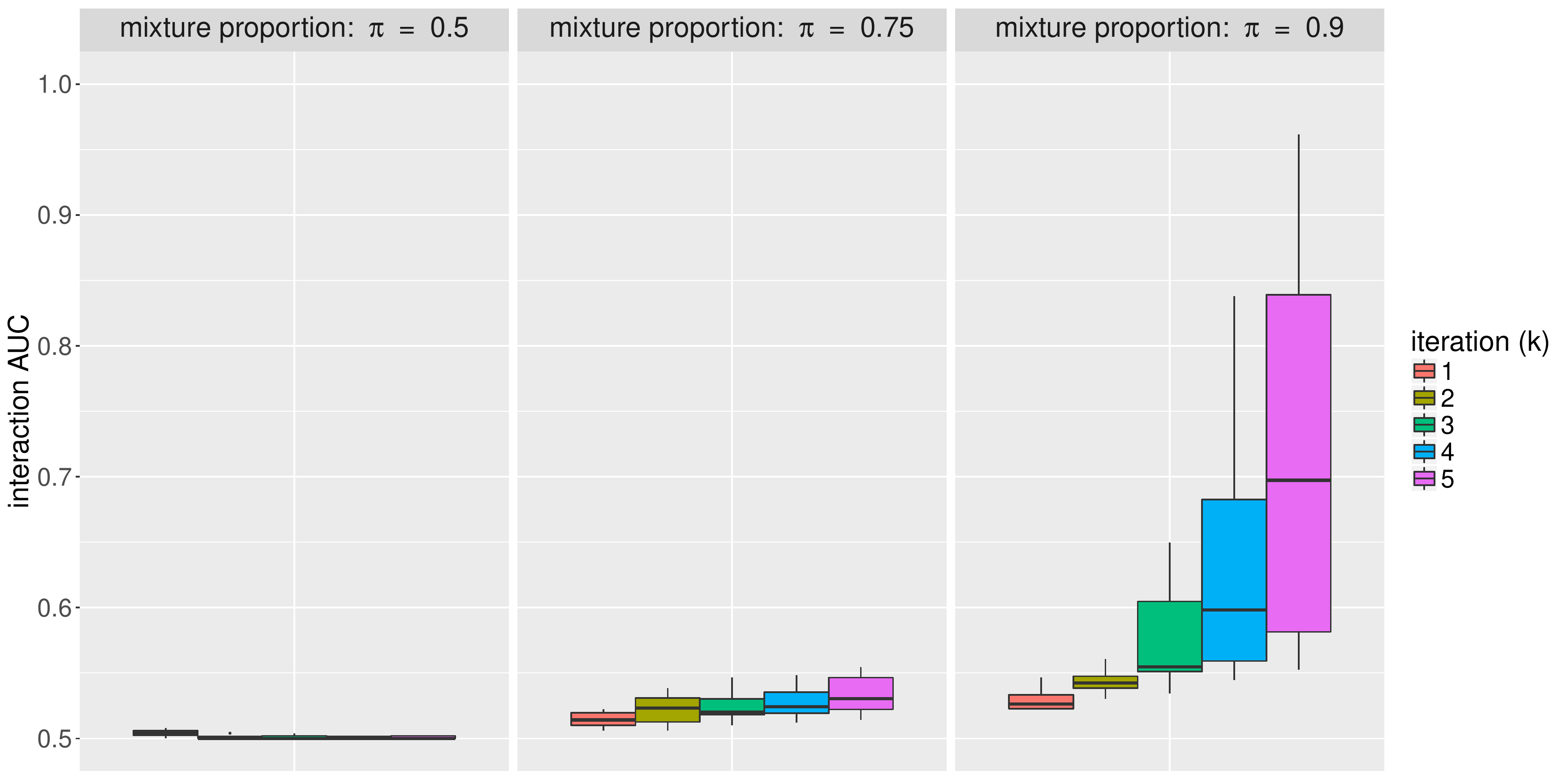}
  \includegraphics[width=0.49\linewidth]{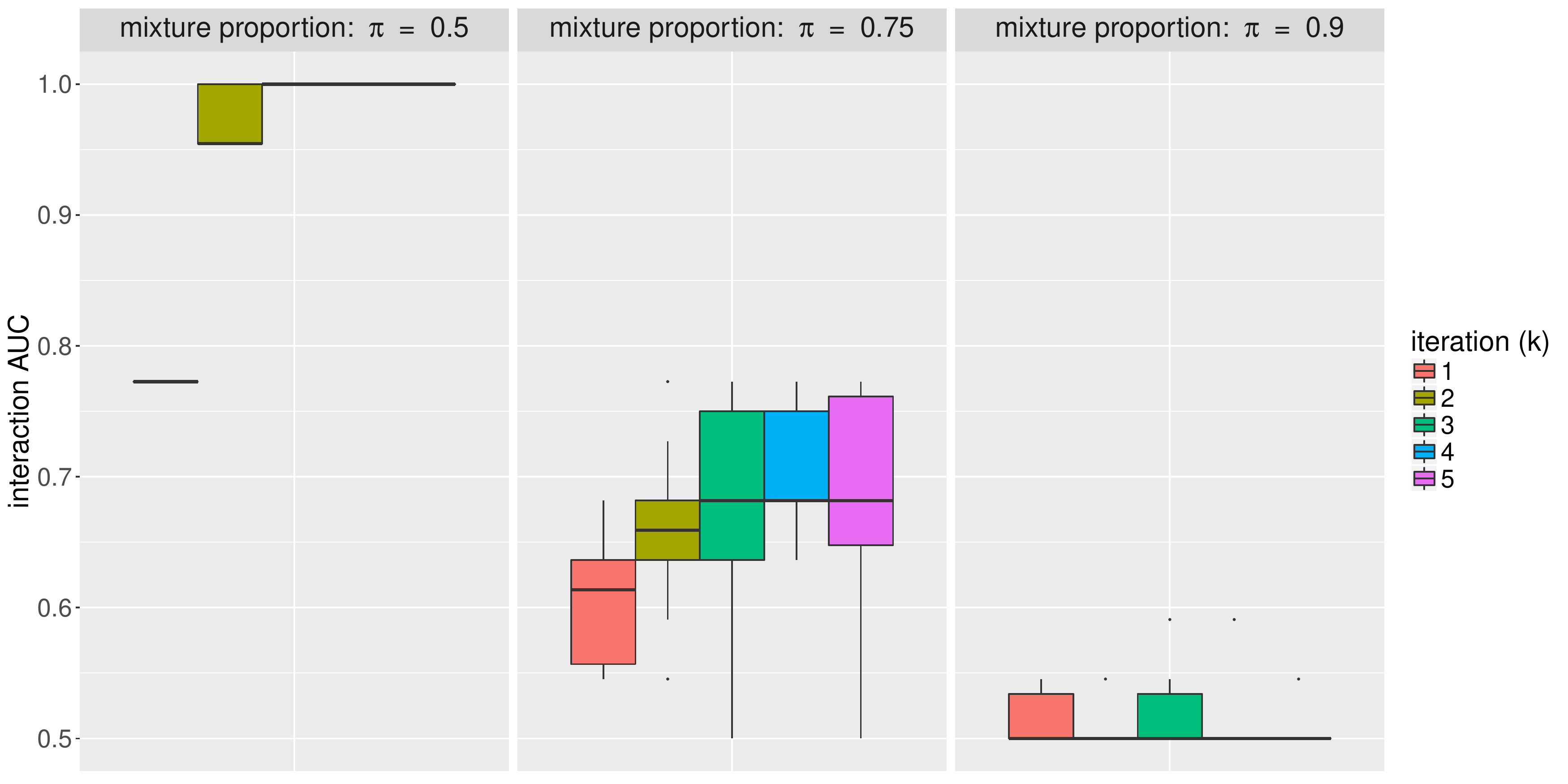}
\caption[iRF performance for mixture of order-$8$ XOR rule and order-$4$ AND rule]{iRF performance for mixture model as a function of mixture proportion ($\pi$) over 10 replicates. All models were trained using $5000$ observations. \textbf{[A]} Prediction accuracy (AUC-PR) is generally poor since iRF tends to learn rules that characterize only a subset of the data. \textbf{[B]} Interaction AUC for the XOR rule. iRF fails to recover this marginally less important rule unless it is represented in a large proportion of the data ($\pi=0.9$). \textbf{[C]} Interaction AUC for the AND rule. iRF recovers the full rule as the most stable interaction for $k\ge 3$ (AUC of $1$) for $\pi=0.5$ despite the fact that the AND interaction is only active in half of the observations. Perfect recovery of the AND rule in a setting where iRF fails to recover the XOR rule indicates that iterative re-weighting based on Gini importance encourages iRF identify rules with more marginally important features.}
\label{fig:mixxor-pred-rec}
\end{figure}

\begin{figure}[p]
    \textbf{A}\hspace{0.49\textwidth}\textbf{B}\\

    \includegraphics[width=0.49\linewidth]{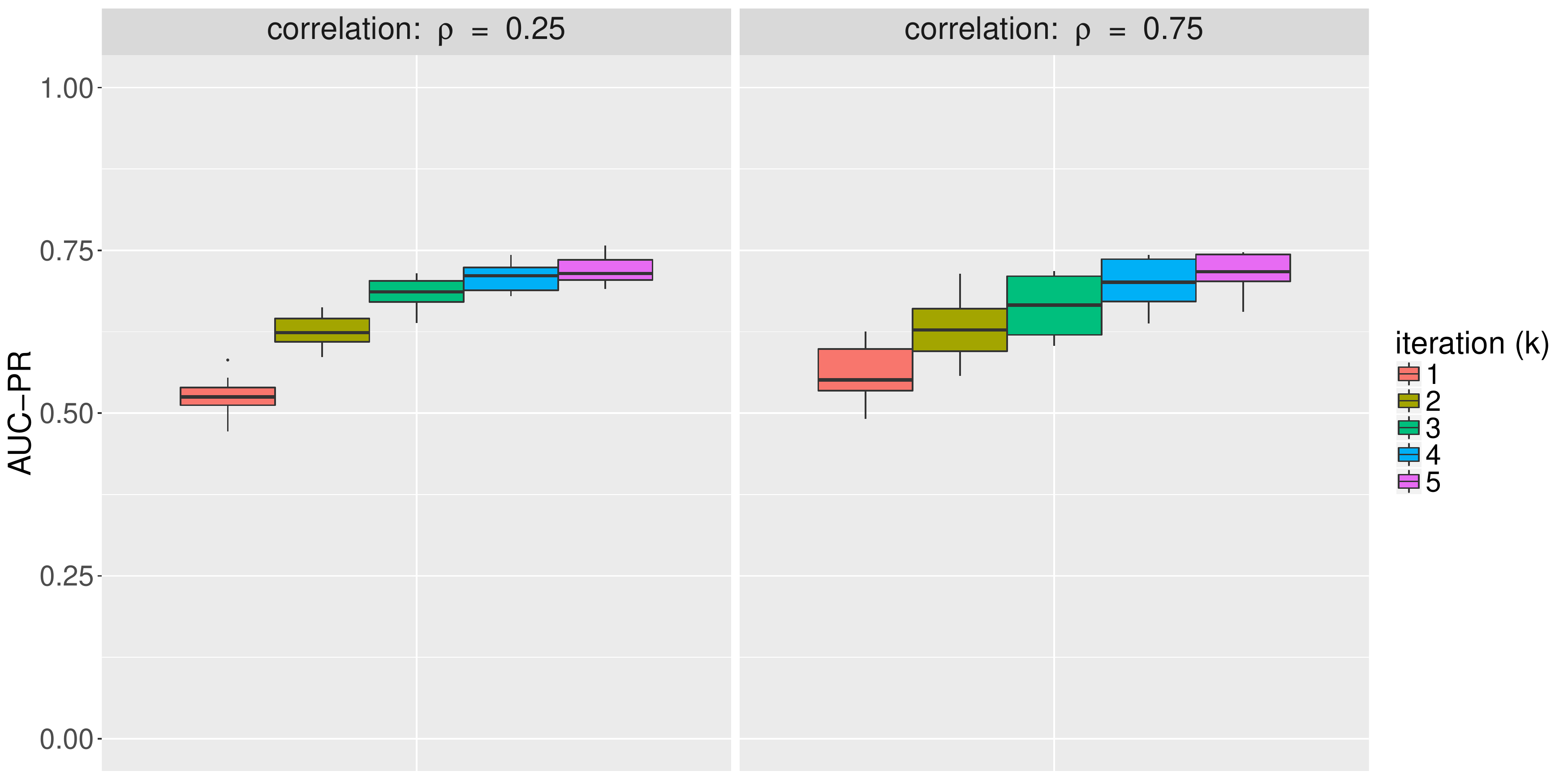}
    \includegraphics[width=0.49\linewidth]{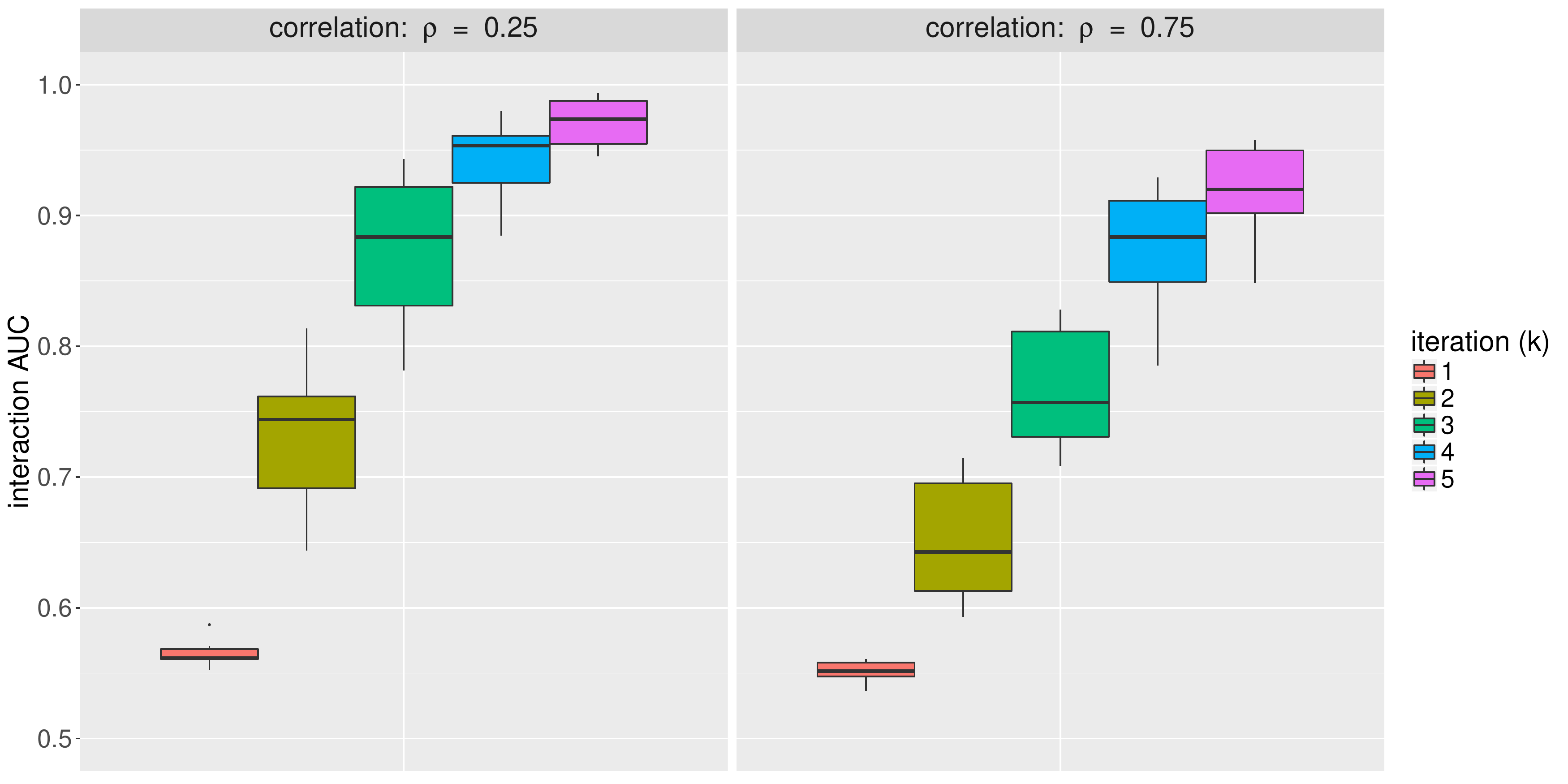}\\
    
    \textbf{C}\hspace{0.49\textwidth}\textbf{D}\\
    
    \includegraphics[width=0.49\linewidth]{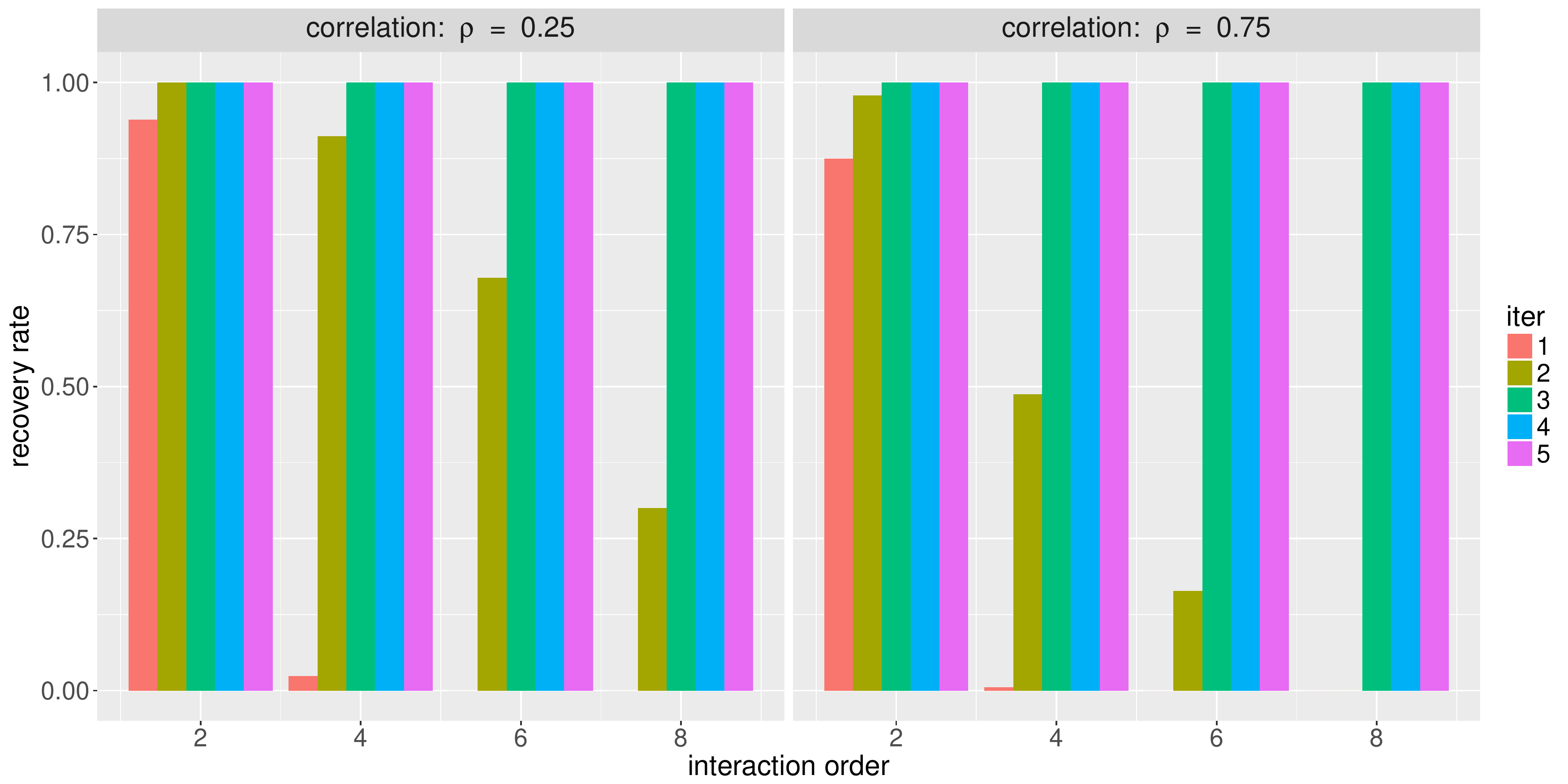}
    \includegraphics[width=0.49\linewidth]{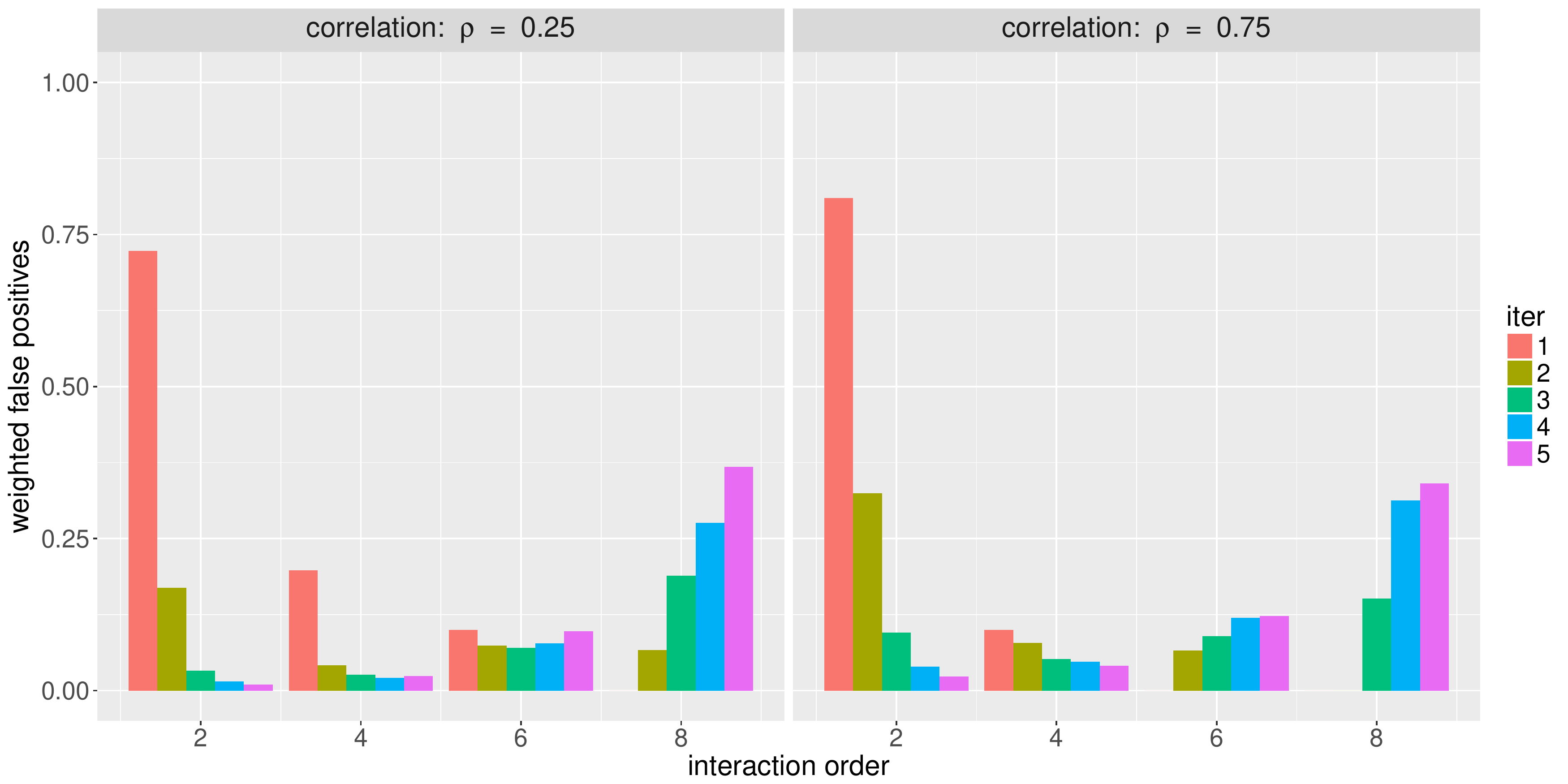} 
\caption[iRF performance for order-$8$ XOR rule, decaying covariance]{iRF performance for order-$8$ XOR rule over 10 replicates as a function of correlation level (decaying covariance structure).
All models were trained using $5000$ observations. \textbf{[A]} Prediction accuracy (AUC-PR) improves for increasing $k$. \textbf{[B]} Interaction AUC improves with increasing $k$, but is more variable than uncorrelated settings. \textbf{[C]} Recovery rate for interactions of all orders improves with increasing $k$. In particular, iRF with $k=1$ rarely recovers any interactions of order $ > 2$. \textbf{[D]} Weighted false positives increase in settings where iRF recovers high-order interactions as a result of many false positives with low stability scores. For order-$2$ interactions, later iterations of iRF filter out many of the false positives identified in earlier iterations.}
\label{fig:xortoeplitz}
\end{figure}

\begin{figure}[p]
    \textbf{A}\hspace{0.49\textwidth}\textbf{B}\\

    \includegraphics[width=0.49\linewidth]{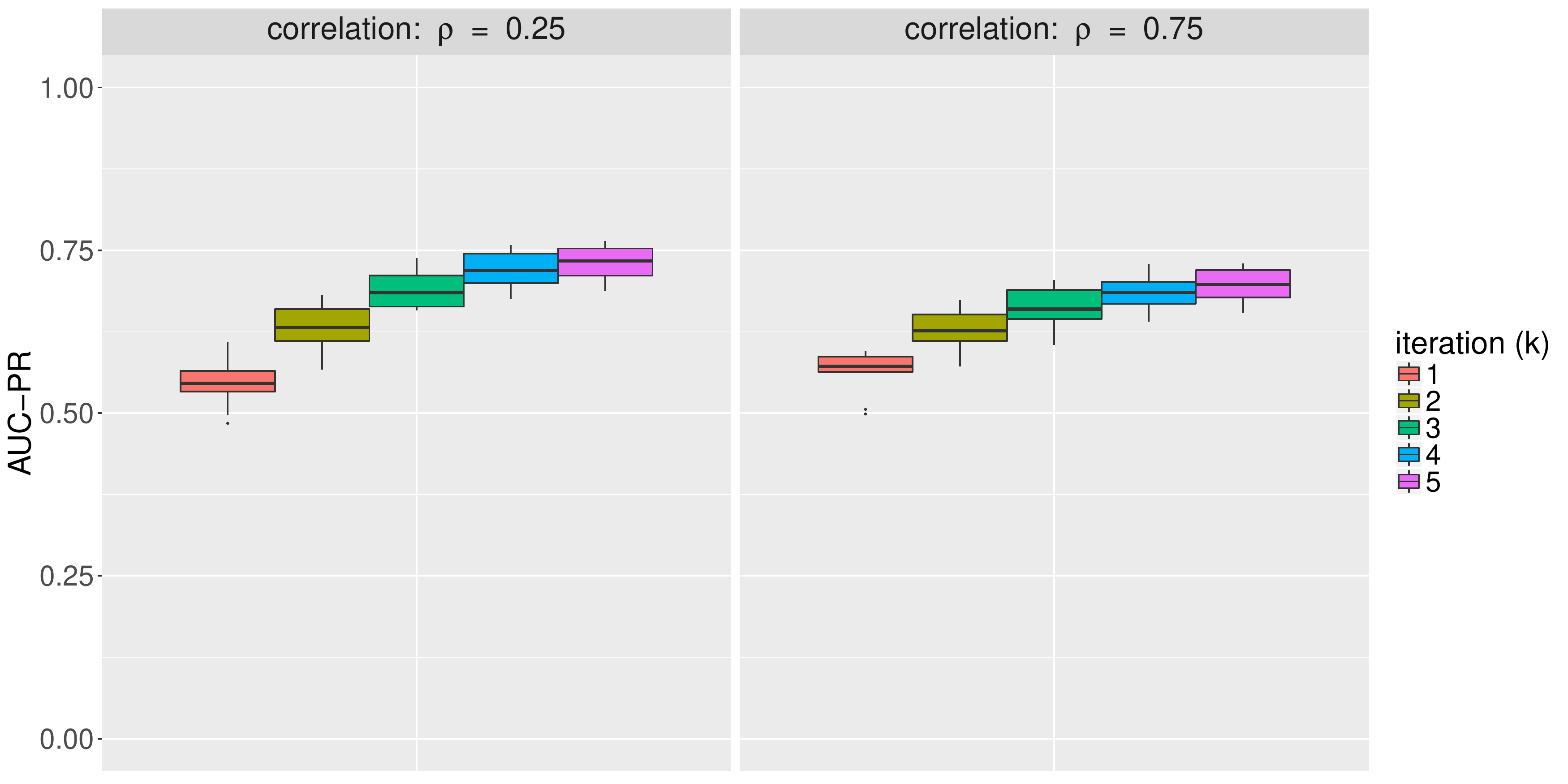}
    \includegraphics[width=0.49\linewidth]{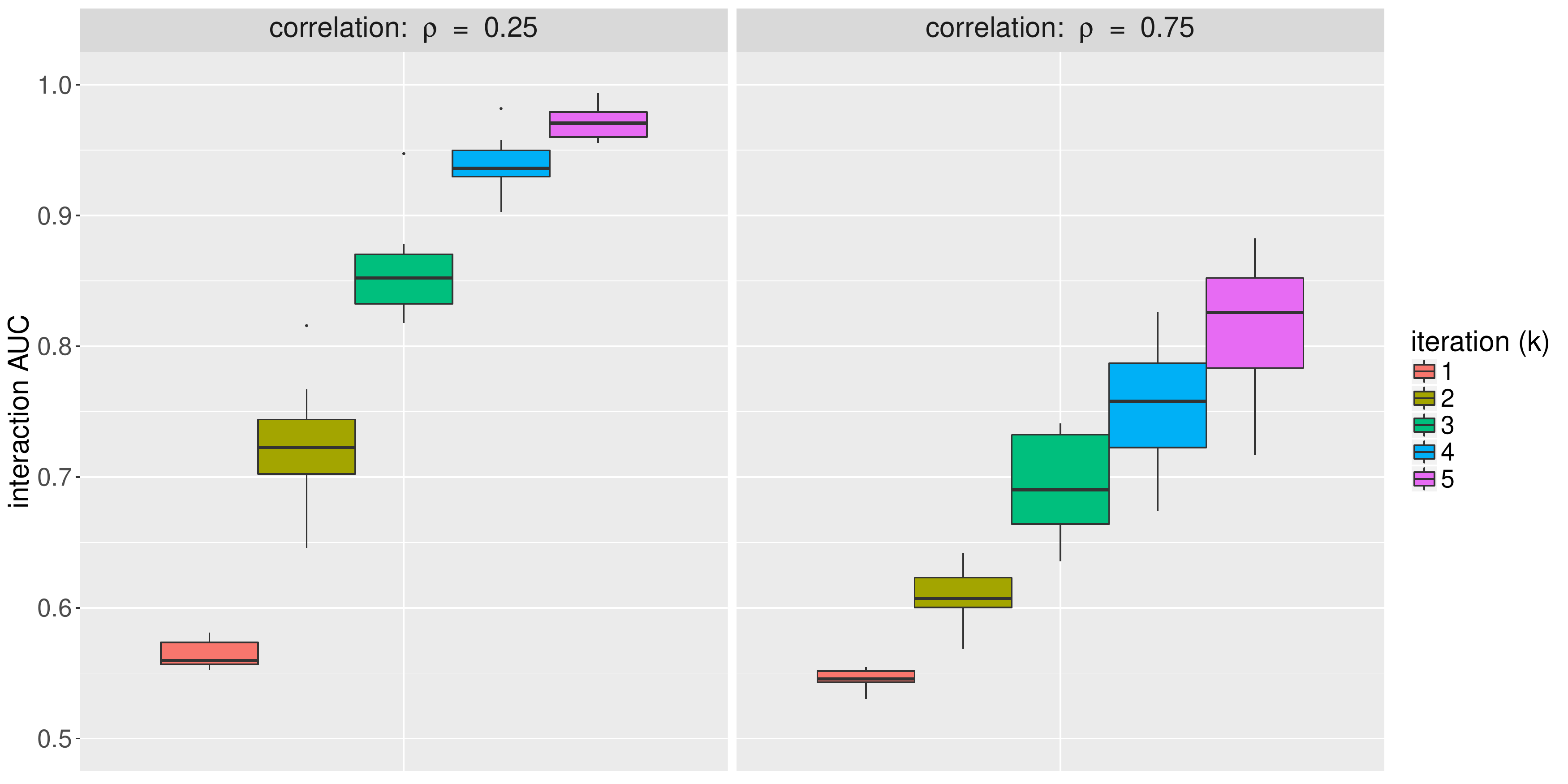}\\
    
    \textbf{C}\hspace{0.49\textwidth}\textbf{D}\\
    
    \includegraphics[width=0.49\linewidth]{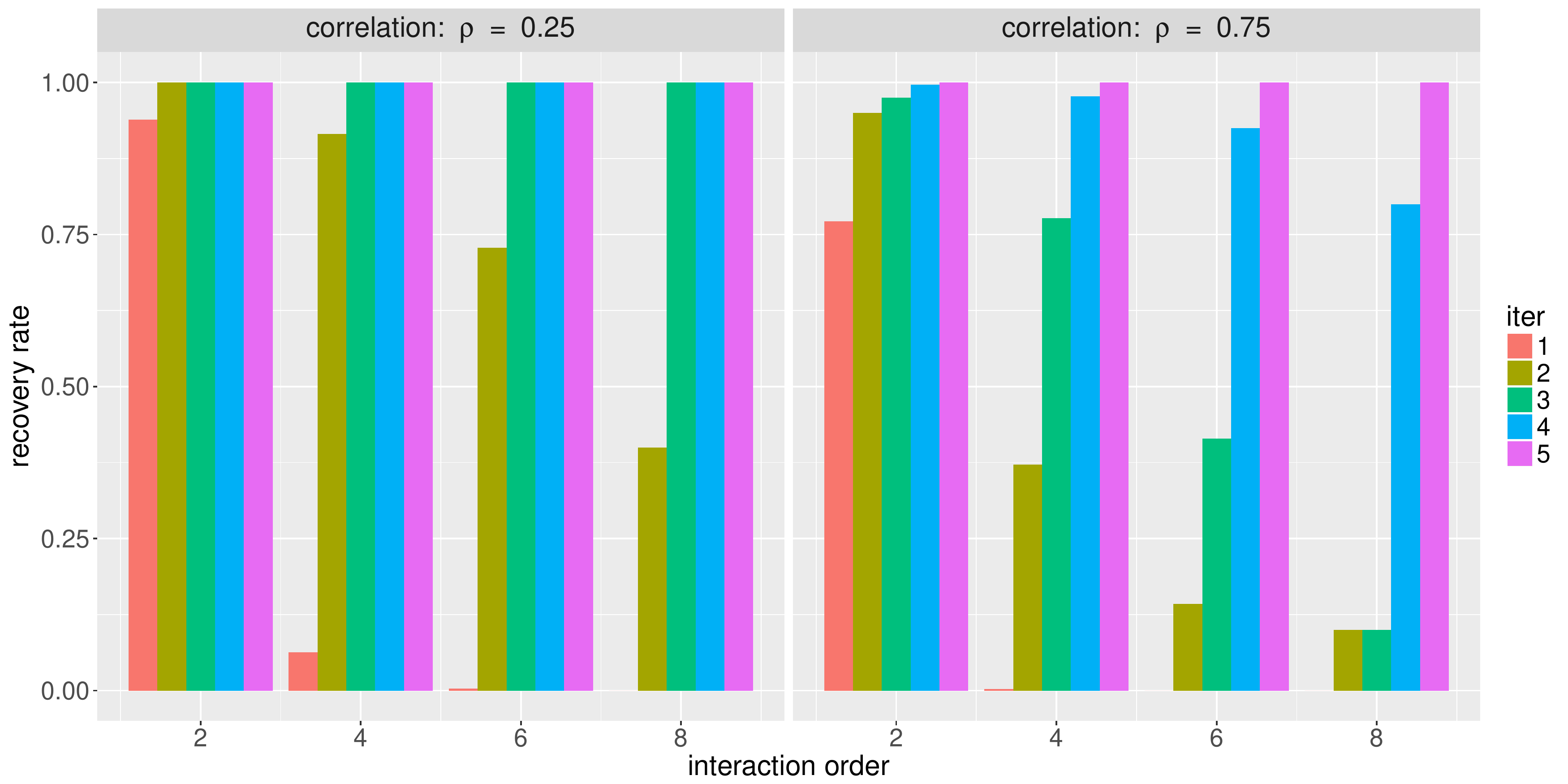}
    \includegraphics[width=0.49\linewidth]{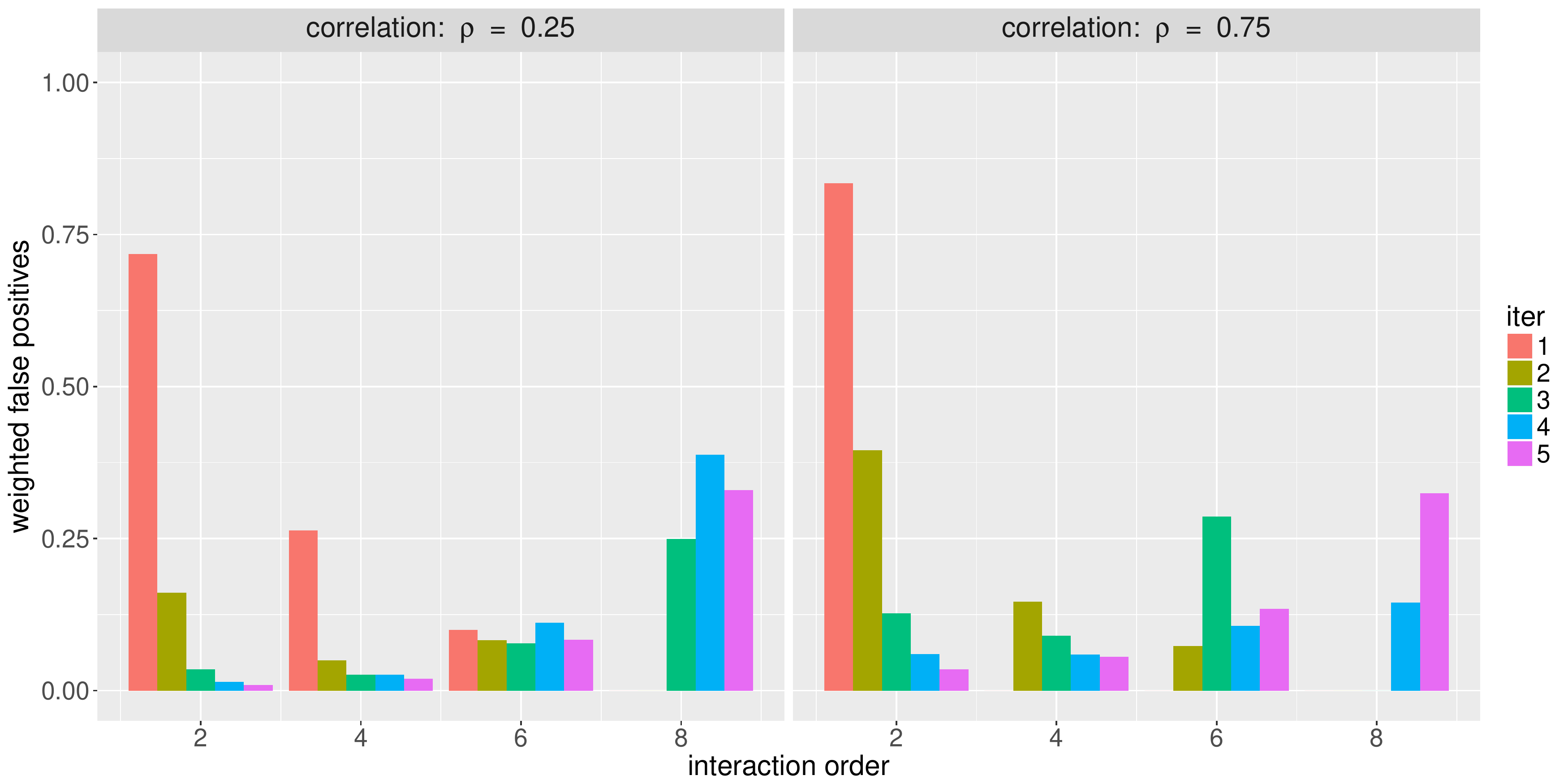}  
  \caption[iRF performance for order-$8$ XOR rule, block covariance]{iRF performance for order-$8$ XOR rule over 10 replicates as a function of correlation level (block covariance).  All models were trained on $5000$ observations. \textbf{[A]} Prediction accuracy (AUC-PR) improves with increasing $k$. \textbf{[B]} Interaction AUC improves with increasing $k$ and drops for large values of $\rho$. Variability is comparable to the decaying covariance case and greater than in uncorrelated settings. \textbf{[C]} Recovery rate for interactions of all orders improves with increasing $k$. In particular, iRF with $k=1$ rarely recovers any interactions of order $ > 2$. \textbf{[D]} Weighted false positives increase in settings where iRF recovers high-order interactions as a result of many false positives with low stability scores. For order-$2$ interactions, later iterations of iRF filter out many of the false positives identified in earlier iterations.} 
    \label{fig:xorblock}
\end{figure}

\begin{figure}[p]
    \textbf{A}\hspace{0.49\textwidth}\textbf{B}\\

    \includegraphics[width=0.49\linewidth]{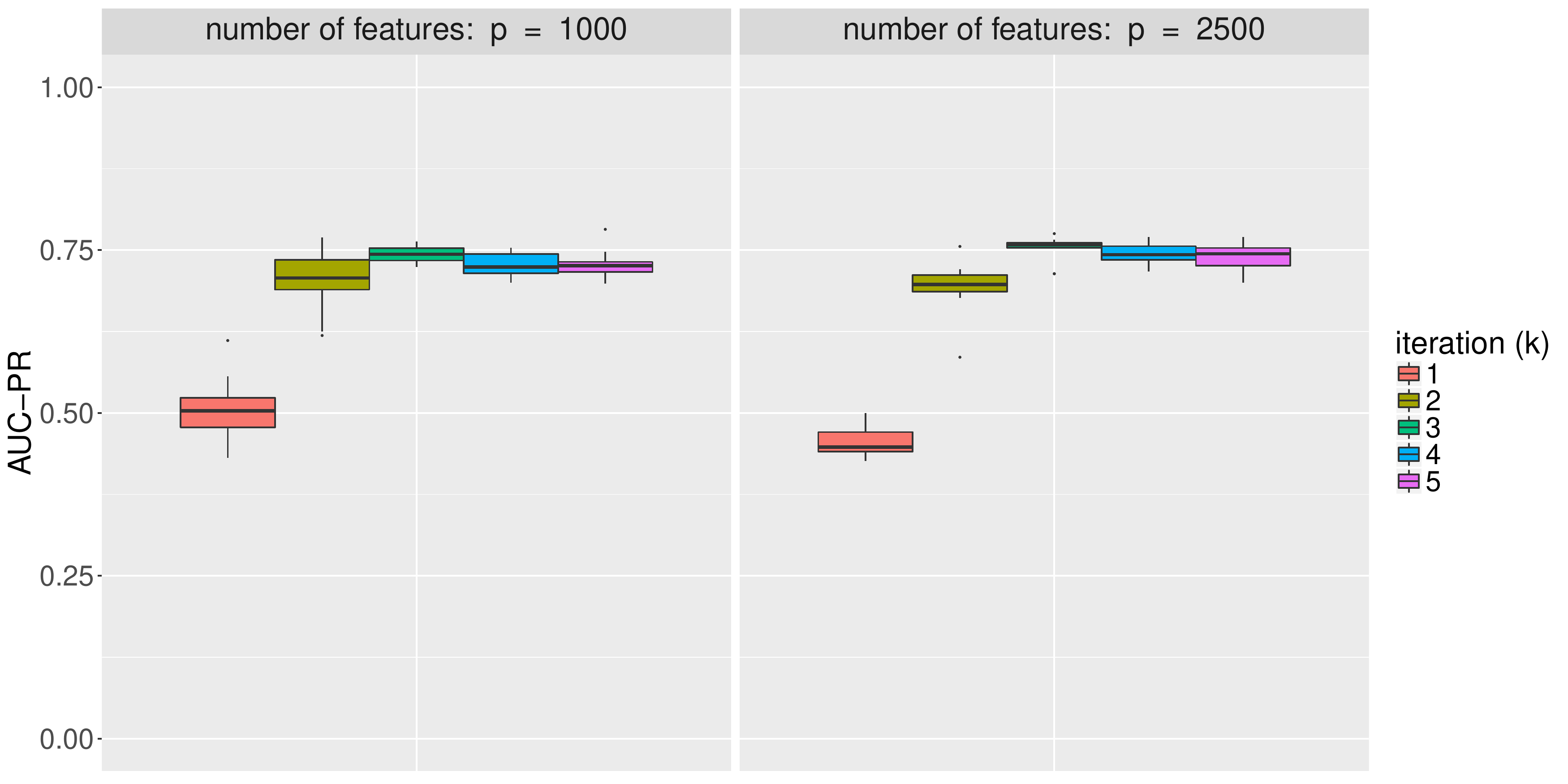}
    \includegraphics[width=0.49\linewidth]{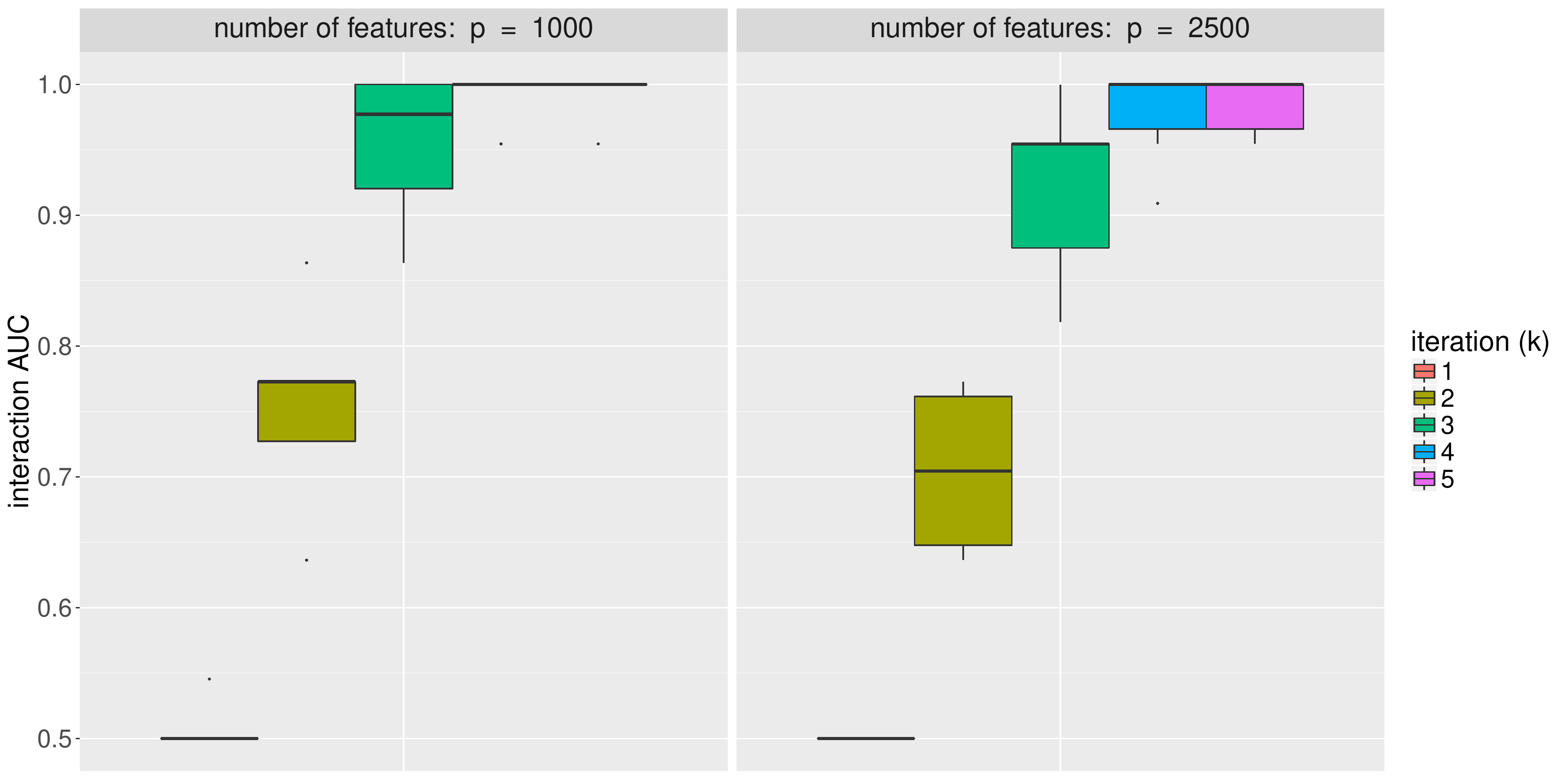}\\
    
    \textbf{C}\hspace{0.49\textwidth}\textbf{D}\\
    
    \includegraphics[width=0.49\linewidth]{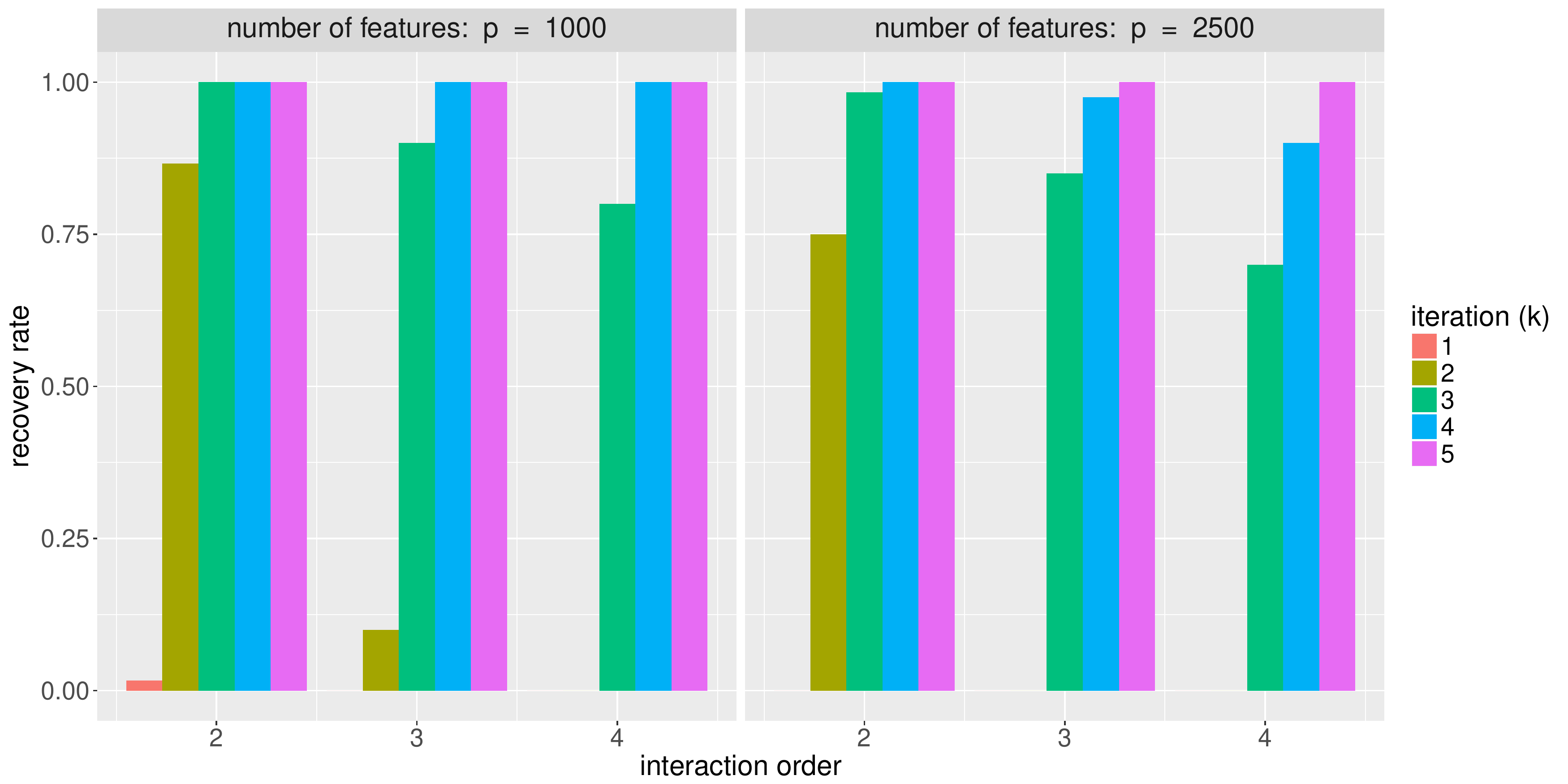}
    \includegraphics[width=0.49\linewidth]{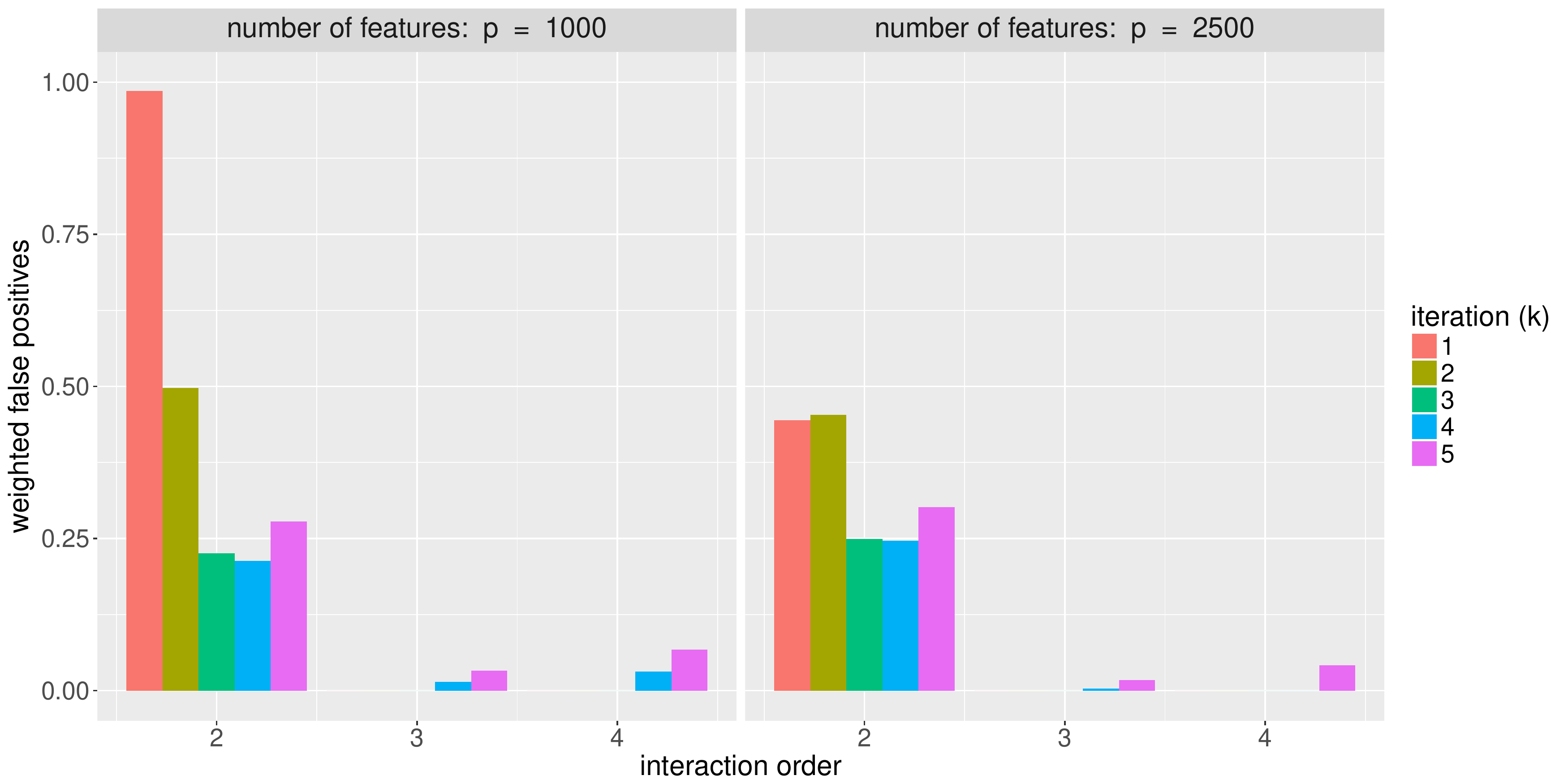} 

\caption[iRF performance for order-$4$ AND rule, noise level $0.2$]{iRF performance for order-$4$ AND rule over 10 replicates with class labels swapped for $20\%$ of observations selected at random. All models were trained using $500$ observations. \textbf{[A]} Prediction accuracy (AUC-PR) improves and stabilizes with increasing $k$. \textbf{[B]} Interaction AUC improves dramatically with increasing $k$. For $k>3$, iRF often recovers the full order-$4$ AND rule as the most stable interaction (AUC of $1$). \textbf{[C]} Recovery rate improves with increasing $k$. For $k=1$, iRF rarely recovers any portion of the data generating rule while for $k>3$ iRF often recovers the full data generating rule. \textbf{[D]} Weighted false positives are low for interactions of order $ > 2$ and drop with iteration for interactions of order-$2$, suggesting that iRF identifies active features through iterative re-weighting.}
\label{fig:p0.2}
\end{figure}

\begin{figure}[p]
    \textbf{A}\hspace{0.49\textwidth}\textbf{B}\\

    \includegraphics[width=0.49\linewidth]{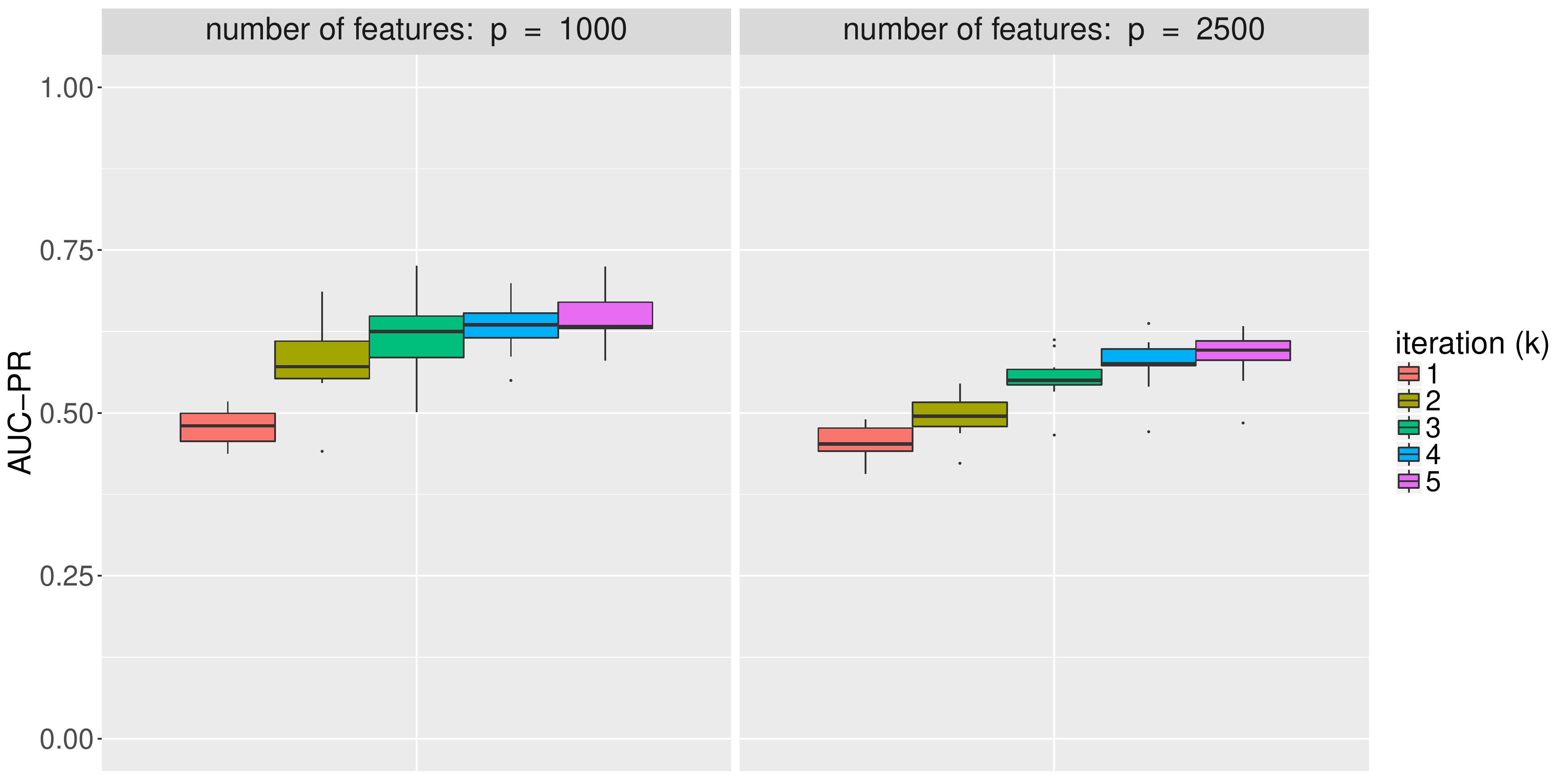}
    \includegraphics[width=0.49\linewidth]{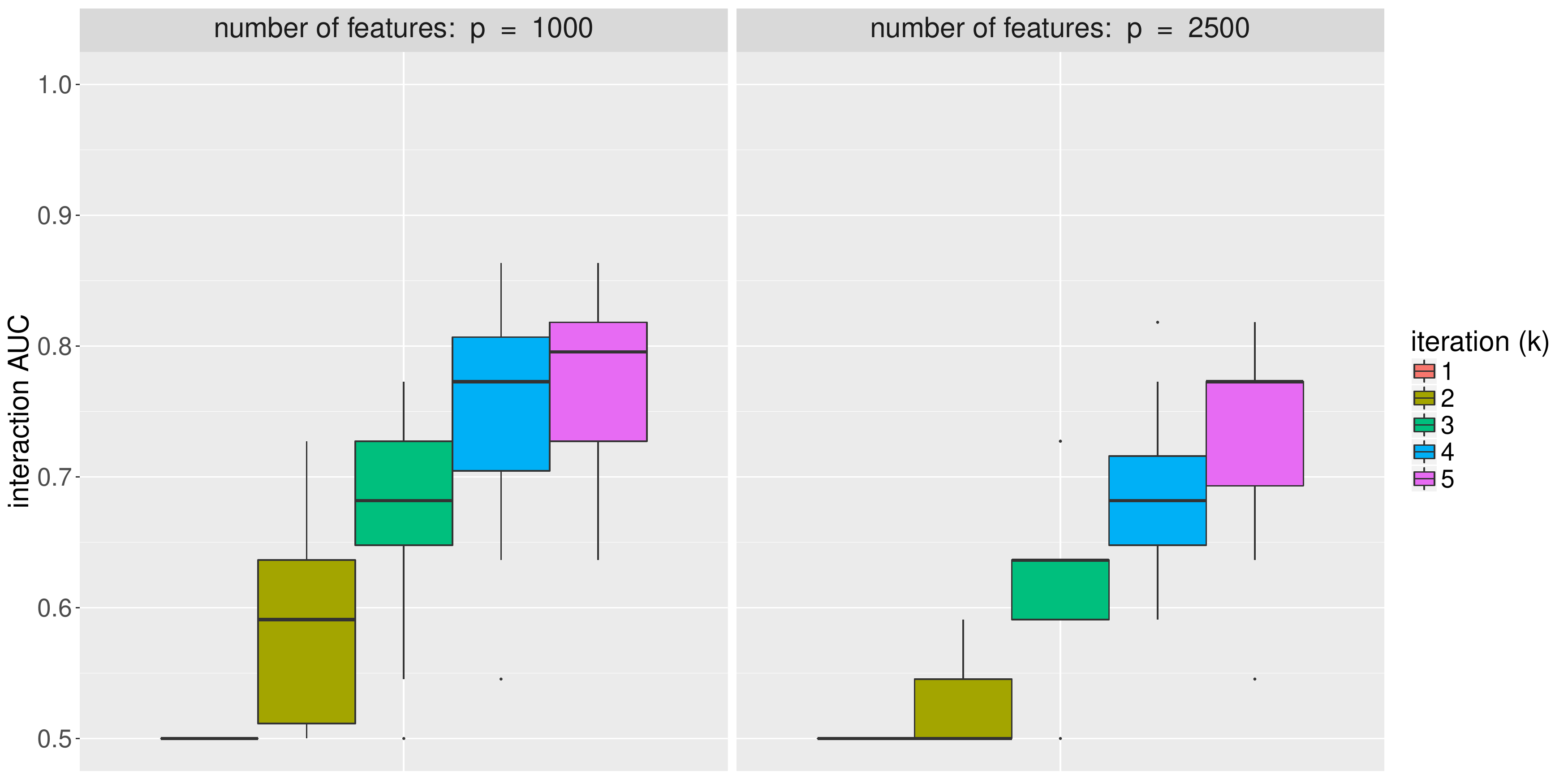}\\
    
    \textbf{C}\hspace{0.49\textwidth}\textbf{D}\\
    
    \includegraphics[width=0.49\linewidth]{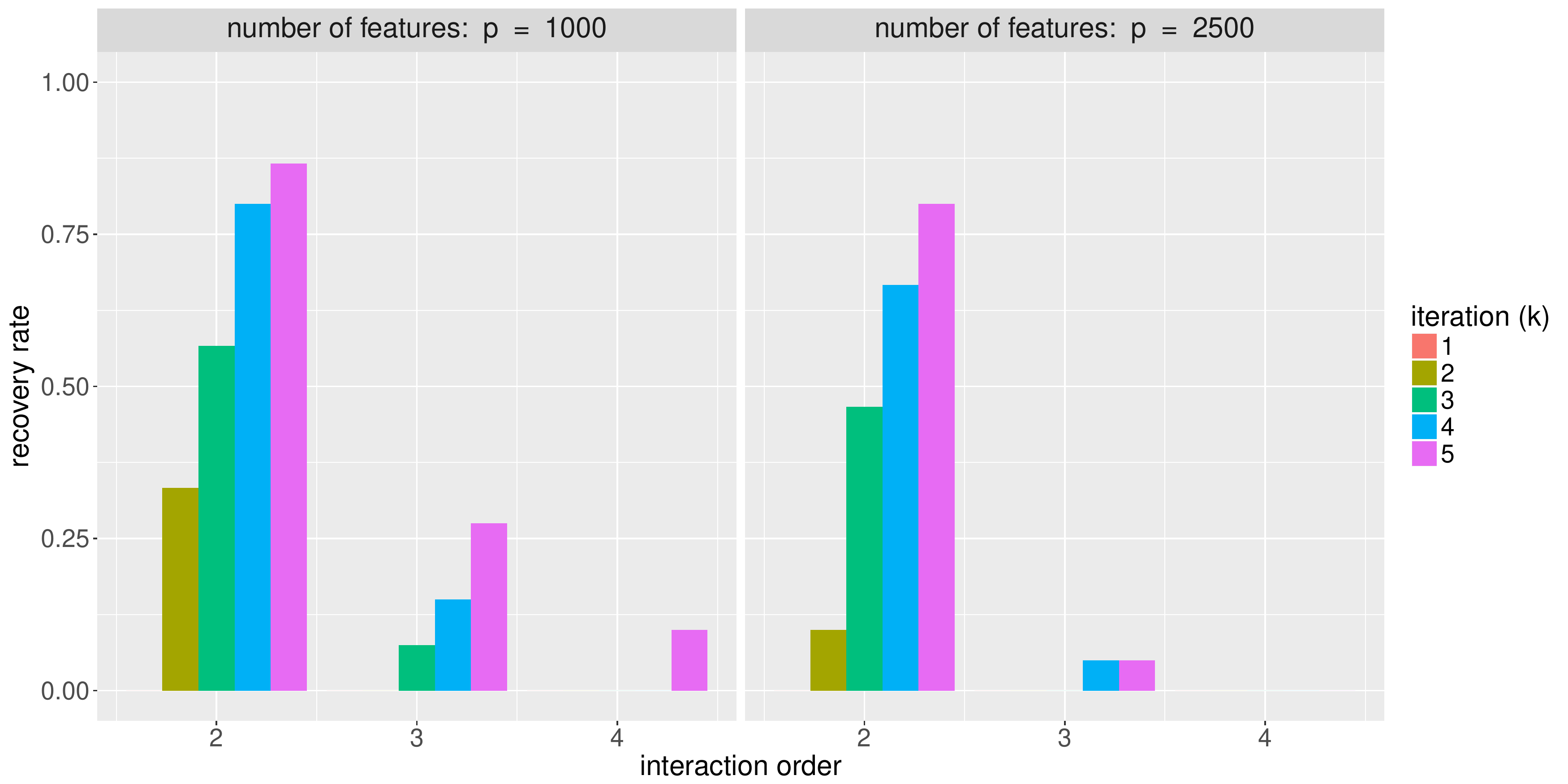}
    \includegraphics[width=0.49\linewidth]{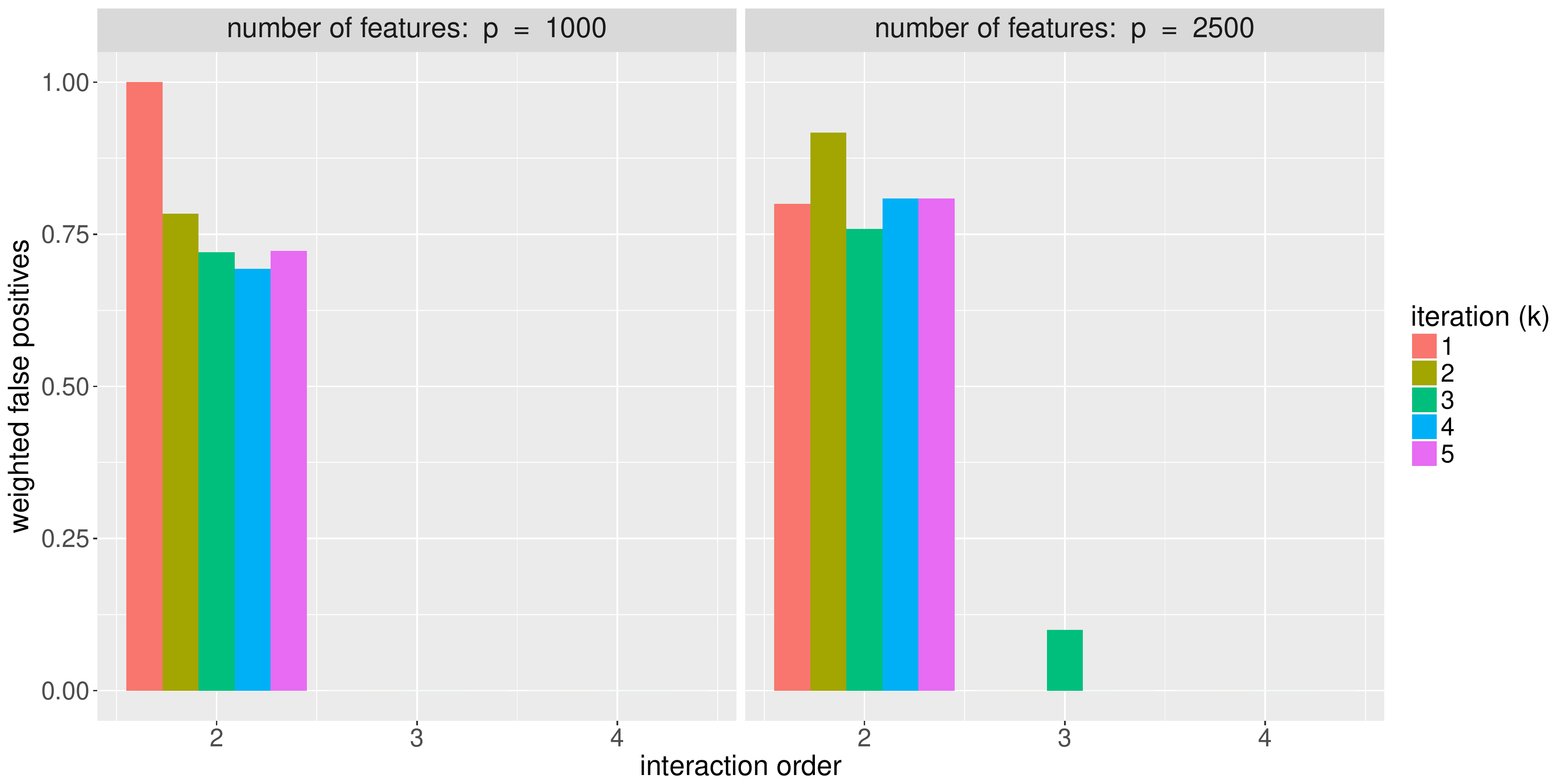} 

\caption[iRF performance for order-$4$ AND rule, noise level $0.3$]{iRF performance for order-$4$ AND rule over 10 replicates with class labels swapped for $30\%$ of observations selected at random. All models were trained using $500$ observations. \textbf{[A]} Prediction accuracy (AUC-PR) gradually improves with increasing $k$. \textbf{[B]} Interaction AUC gradually improves with increasing $k$ but does not achieve perfect recovery of the data generating rule. \textbf{[C]} Recovery rate improves with increasing $k$, but iRF recovers higher-order interactions less frequently than at lower noise levels. \textbf{[D]} Weighted false positives are comparable across $k$ and particularly high for order-$2$ interactions.}
\label{fig:p0.3}
\end{figure}

\begin{figure}[p]
\textbf{A}\hspace{0.49\textwidth}\textbf{B}\\
\includegraphics[width=0.49\textwidth]{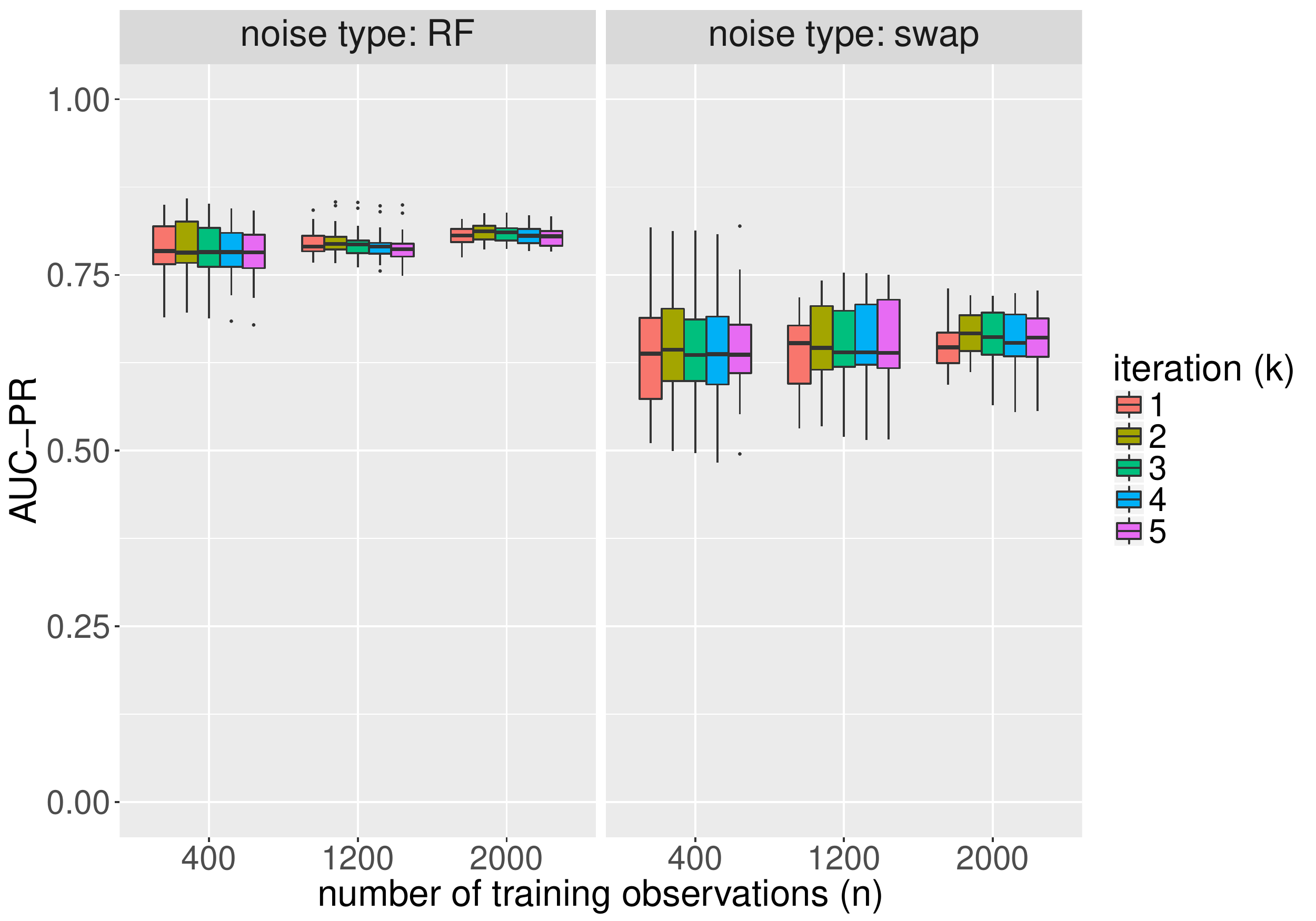}
\includegraphics[width=0.49\textwidth]{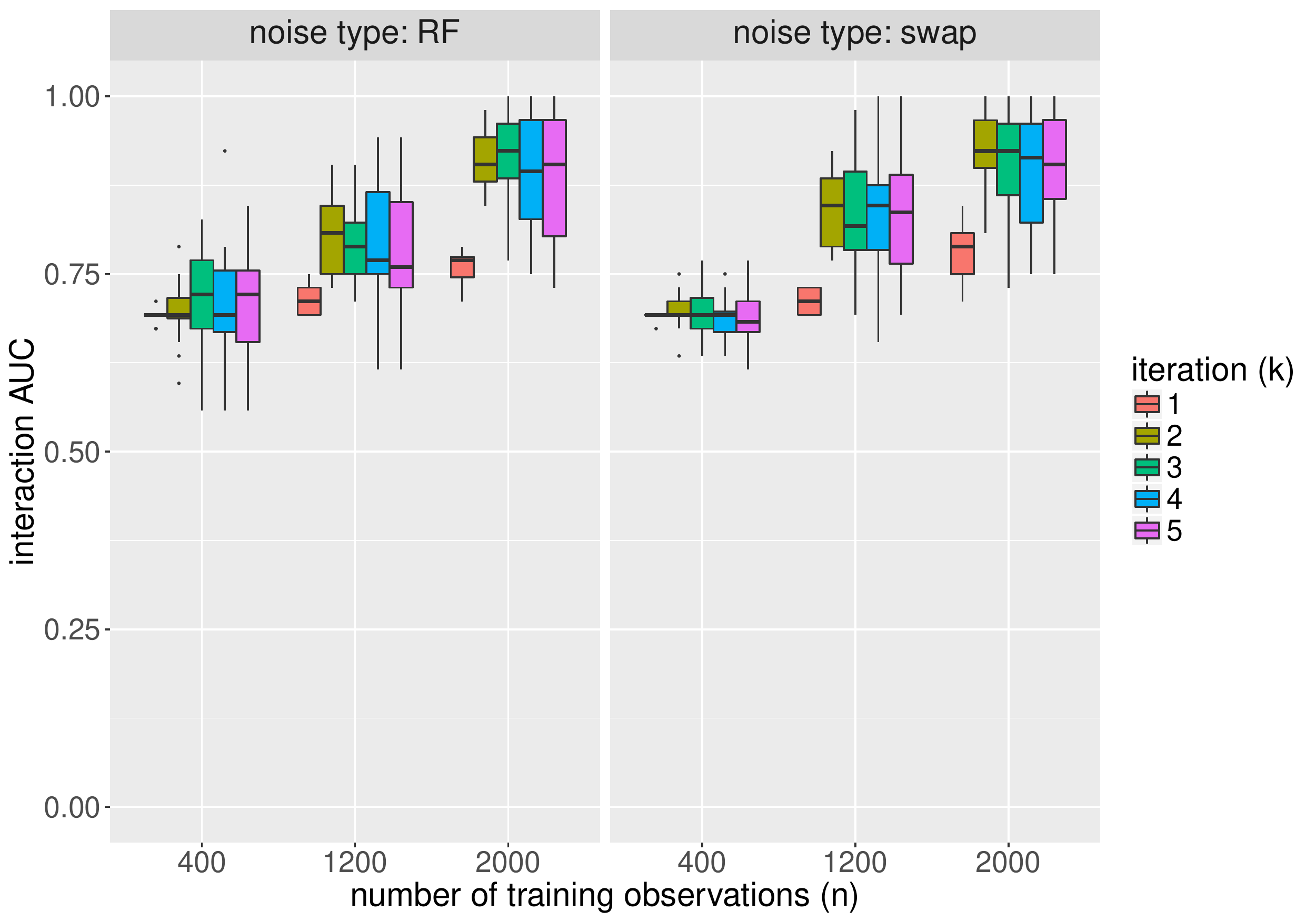}\\
\textbf{C}\hspace{0.49\textwidth}\textbf{D}\\
\includegraphics[width=0.49\textwidth]{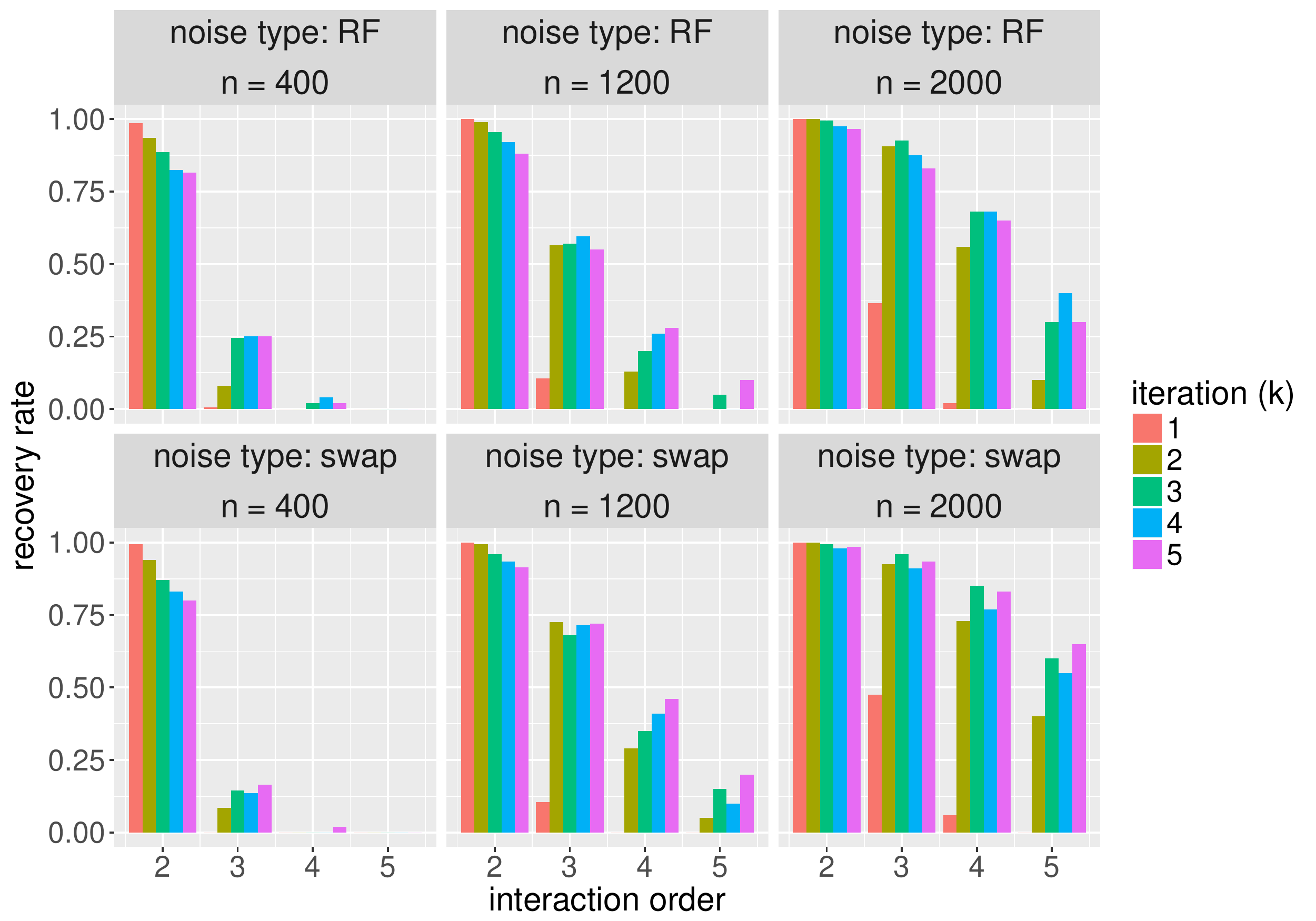}
\includegraphics[width=0.49\textwidth]{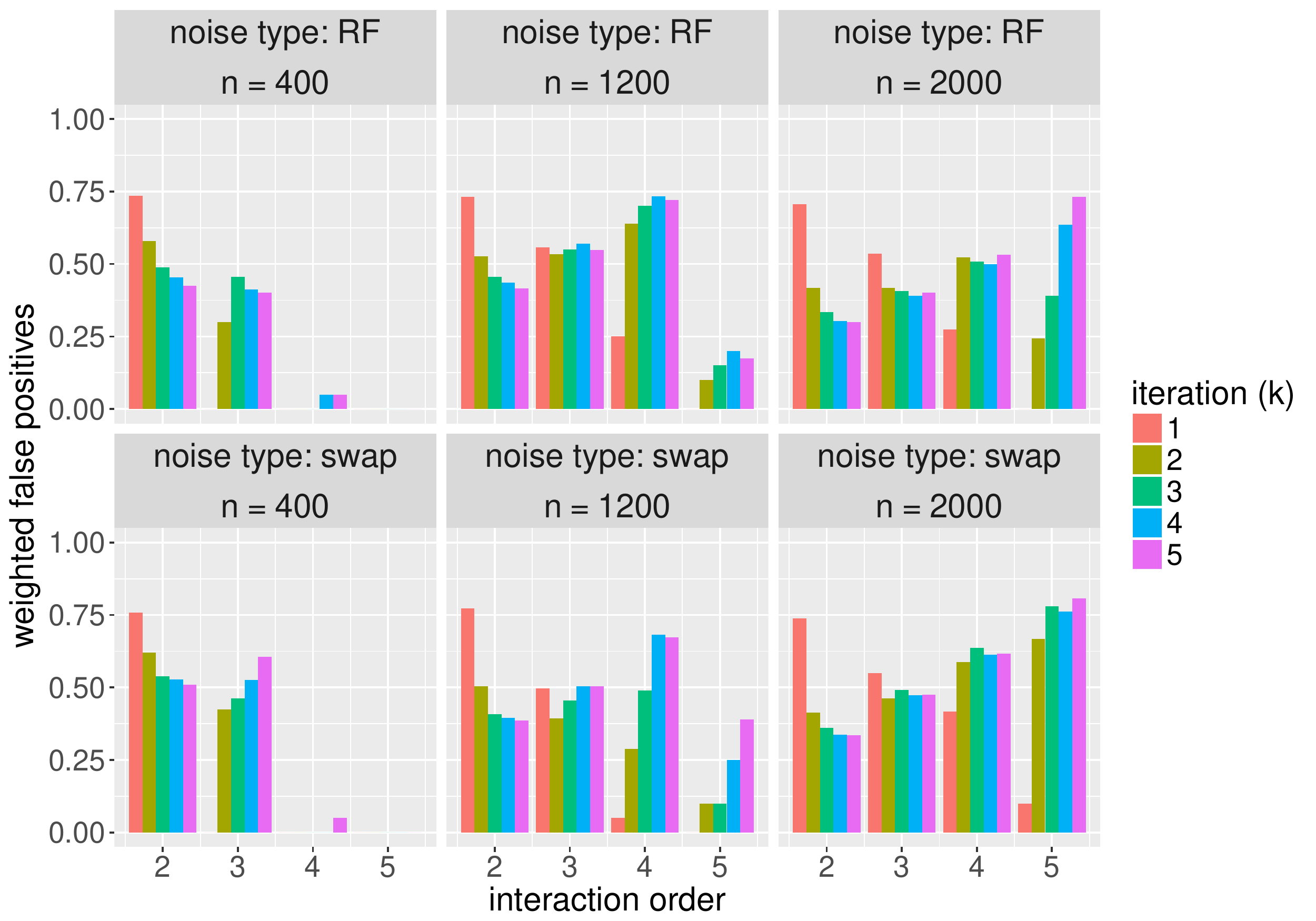}
\caption[iRF performance for enhancer data simulation]{iRF performance for the enhancer data simulations by noise type. Results are shown for models trained using $400, 1200, $ and $2000$ observations. \textbf{[A]} Prediction accuracy (AUC-PR) remains consistent with increasing $k$ in both noise models. \textbf{[B]} Interaction AUC improves after iteration $k=1$, especially for larger training samples where high-order interactions are recovered. Some settings show a drop in interaction AUC as $k$ increases from $2$ to $5$, emphasizing the importance of tuning $K$. \textbf{[C]} Recovery rate improves beyond $k=1$ for high-order interactions and is fairly consistent for $k=2, \dots, 5$. \textbf{[D]} Weighted false positives drop beyond $k=1$ for order-$2$ interactions as iterative re-weighting encourages the selection of active features. With larger training samples, iRF recovers many interactions among both active and inactive features. The stability scores of interactions among active features are consistently higher than interactions including inactive features.}\label{fig:sim3}
\end{figure}

\begin{figure}[p]
\begin{center}
\includegraphics[width=0.85\textwidth]{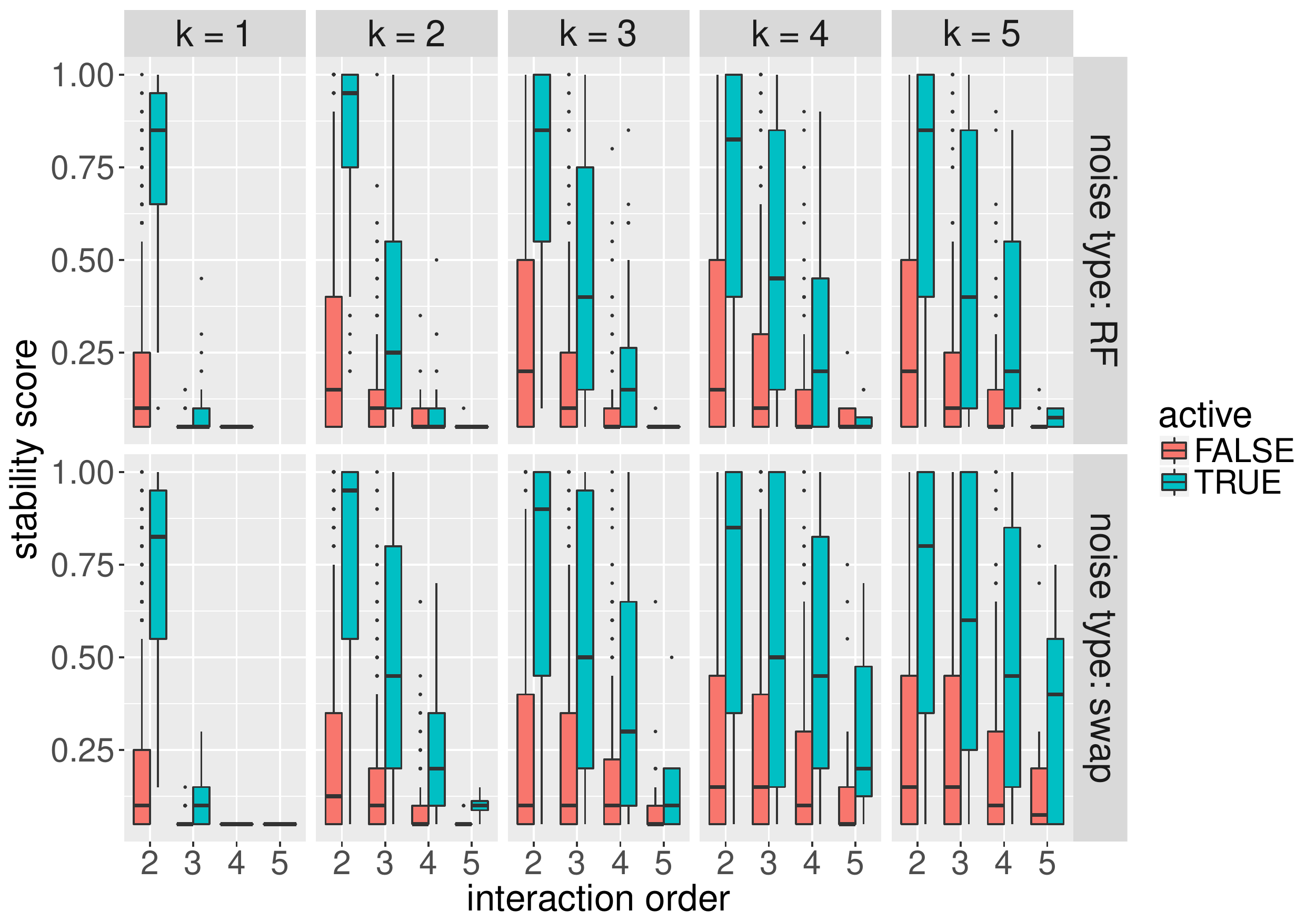}
\caption[iRF enhancer data simulation stability scores]{Distributions of iRF stability scores for active and inactive variables by iteration ($k$) and noise type. Both models were trained using $2000$ observations. Interactions among active features are consistently identified as more stable in both noise settings, and higher order interactions are only identified in later iterations.}\label{fig:enhancerStab}
\end{center}
\end{figure}

\begin{figure}[p]
\begin{center}
\includegraphics[width=0.9\textwidth]{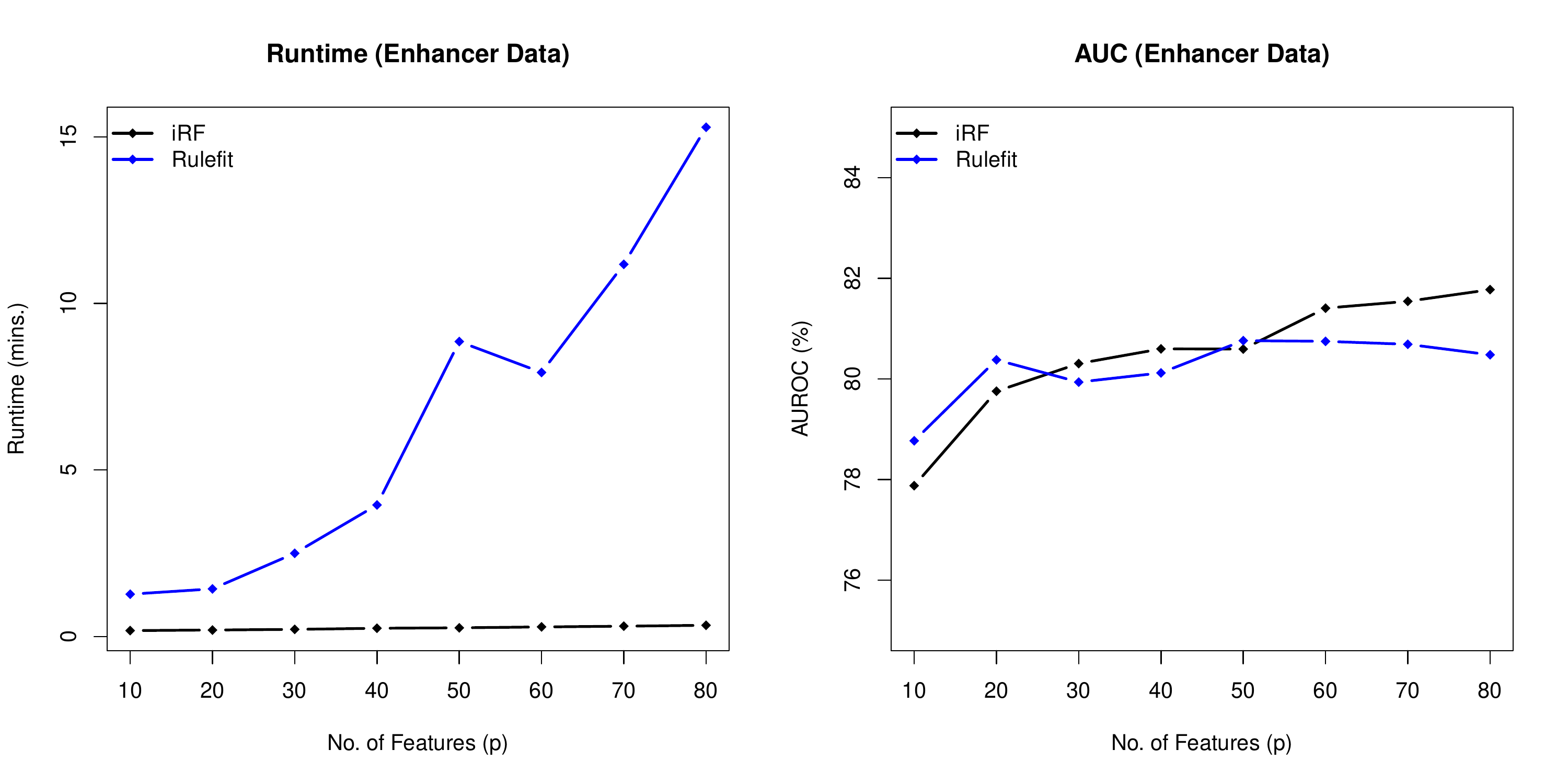}
\caption[Runtime comparison of iRF and Rulefit]{Runtime (left) of interaction detection and Area under ROC curve (right) of prediction by Rulefit and iRF on subsets of the enhancer data with $p \in \{10, 20, \ldots, 80 \}$ features and balanced training and test sets, each of size $n=731$. The results are averaged over $10$ different permutations of the original features in the enhancer dataset. The two algorithms provide similar classification accuracy in test data, although computational cost of iRF grows much slower with $p$, compared to the computational cost of Rulefit.}\label{fig:pic/runtime-taly-splicing}
\end{center}
\end{figure}